\PassOptionsToPackage{sort}{natbib}
\documentclass[nohyperref]{article}

\usepackage[T1]{fontenc}

\usepackage[tracking=true]{microtype}
\newcommand\spacedstyle[1]{\SetTracking{encoding=*}{#1}\lsstyle}
\usepackage{graphicx}
\usepackage{subfigure}
\usepackage{booktabs} 

\usepackage{hyperref}

\usepackage[accepted]{icml2022}

\usepackage{amsmath}
\usepackage{amssymb}
\usepackage{mathtools}
\usepackage{amsthm}

\usepackage[capitalize]{cleveref}

\theoremstyle{plain}

\theoremstyle{definition}

\theoremstyle{remark}

\usepackage{wrapfig}
\usepackage{enumitem}
\usepackage[small]{caption}
\usepackage{xcolor}
\usepackage[export]{adjustbox}
\usepackage{siunitx}
\usepackage{multirow}
\usepackage{bigdelim}

\newcommand{\code}{\texttt}
\usepackage{xspace}
\newcommand{\rsq}{$R^2$\xspace}
\newcommand{\up}{$\uparrow$\xspace}
\newcommand{\down}{$\downarrow$\xspace}

\newcommand{\xb}{\mathbf{x}}
\newcommand{\zb}{\mathbf{z}}
\newcommand{\yb}{\mathbf{y}}
\newcommand{\ybpred}{\hat{\mathbf{y}}}

\icmltitlerunning{Generalization and Robustness Implications in Object-Centric Learning}

\begin{document}

\twocolumn[
\icmltitle{Generalization and Robustness Implications in Object-Centric Learning}



\icmlsetsymbol{equal}{*}

\begin{icmlauthorlist}
\icmlauthor{Andrea Dittadi}{dtu,mpi}
\icmlauthor{Samuele Papa}{dtu}
\icmlauthor{Michele De Vita}{dtu}\\
\icmlauthor{Bernhard Schölkopf}{mpi}
\icmlauthor{Ole Winther}{dtu,ku,cph}
\icmlauthor{Francesco Locatello}{amazon}
\end{icmlauthorlist}

\icmlaffiliation{dtu}{Technical University of Denmark}
\icmlaffiliation{mpi}{Max Planck Institute for Intelligent Systems, Tübingen, Germany}
\icmlaffiliation{ku}{University of Copenhagen}
\icmlaffiliation{cph}{Rigshospitalet, Copenhagen University Hospital}
\icmlaffiliation{amazon}{Amazon}

\icmlcorrespondingauthor{Andrea Dittadi}{adit@dtu.dk}

\icmlkeywords{Machine Learning, Representation Learning, Deep Learning, Unsupervised, Object-Centric Learning, Object-Centric Representations, Generalization, Out-of-distribution, Robustness}

\vskip 0.3in
]



\printAffiliationsAndNotice{}  

\begin{abstract}
The idea behind object-centric representation learning is that natural scenes can better be modeled as compositions of objects and their relations as opposed to distributed representations. This inductive bias can be injected into neural networks to potentially improve systematic generalization and performance of downstream tasks in scenes with multiple objects.
In this paper, we train state-of-the-art unsupervised models on five common multi-object datasets and evaluate segmentation metrics and downstream object property prediction.
In addition, we study generalization and robustness by investigating the settings where either a single object is out of distribution---e.g., having an unseen color, texture, or shape---or global properties of the scene are altered---e.g., by occlusions, cropping, or increasing the number of objects.
From our experimental study, we find object-centric representations to be useful for downstream tasks and generally robust to most distribution shifts affecting objects. However, when the distribution shift affects the input in a less structured manner, robustness in terms of segmentation and downstream task performance may vary significantly across models and distribution shifts.\looseness=-1
\end{abstract}

\section{Introduction}

In object-centric representation learning, we make the assumption that visual scenes are composed of multiple entities or objects that interact with each other, and exploit this compositional property as inductive bias for neural networks. Informally, the goal is to find transformations $r$ of the data $\xb$ into a \emph{set} of vector representations $r(\xb) = \{\zb_k\}$ each corresponding to an individual object, without supervision \citep{eslami2016air, greff2017neural,kosiorek2018sequential,crawford2019spatially, greff2019multi, burgess2019monet, engelcke2020genesis, lin2020space, locatello2020object, chen2020learning,weis2020unmasking,mnih2014recurrent,gregor2015draw,yuan2019generative}.
Relying on this inductive bias, object-centric representations are conjectured to be more robust than distributed representations, and to enable the systematic generalization typical of symbolic systems while retaining the expressiveness of connectionist approaches \cite{bengio2013representation,lake2017building,greff2020binding,scholkopf2021toward}.
Grounding for these claims comes mostly from cognitive psychology and neuroscience \cite{spelke1990principles, teglas2011pure, wagemans2015oxford}. E.g., infants learn about the physical properties of objects as entities that behave consistently over time \cite{baillargeon1985object,spelke2007core} and are able to re-apply their knowledge to new scenarios involving previously unseen objects \cite{dehaene2020how}. Similarly, in complex machine learning tasks like physical modelling and reinforcement learning, it is common to train from the internal representation of a simulator \cite{battaglia2016interaction, sanchez-gonzalez2020learning} or of a game engine \cite{vinyals2019grandmaster, openai2019dota} rather than from raw pixels, as more abstract representations facilitate learning.
Finally, learning to represent objects separately is a crucial step towards learning causal models of the data from high-dimensional observations, as objects can be interpreted as causal variables that can be manipulated independently~\cite{scholkopf2021toward}. Such causal models are believed to be crucial for human-level generalization~\cite{pearl2009causality,peters2017elements}, but traditional causality research assumes causal variables to be given rather than learned~\cite{scholkopf2019causality}.

As object-centric learning developed recently as a subfield of representation learning, we identify three key hypotheses and design systematic experiments to test them. 
(1)~\emph{The unsupervised learning of objects as pretraining task is useful for downstream tasks.} Besides learning to separate objects without supervision, current approaches are expected to separately represent information about each object's properties, so that the representations can be useful for arbitrary downstream tasks. 
(2)~\emph{In object-centric models, distribution shifts affecting a single object do not affect the representations of other objects.}
If objects are to be represented independently of each other to act as compositional building blocks for higher-level cognition~\cite{greff2020binding}, changes to one object in the input should not affect the representation of the unchanged objects. This should hold even if the change leads to an object being out of distribution (OOD).
(3)~\emph{Object-centric models are generally robust to distribution shifts, even if they affect global properties of the scene.} 
Even if the whole scene is OOD---e.g., if it contains more objects than in the training set---object-centric approaches should be robust thanks to their inductive bias.\looseness=-1

In this paper, we systematically investigate these three concrete hypotheses by re-implementing popular unsupervised object discovery approaches and testing them on five multi-object datasets.\footnote{Training and evaluating all the models for the main study requires approximately 1.44 GPU years on NVIDIA V100.} 
We find that:
(1)~Object-centric models achieve good downstream performance on property prediction tasks. We also observe a strong correlation between segmentation metrics, reconstruction error, and downstream property prediction performance, suggesting potential model selection strategies. 
(2)~If a single object is out of distribution, the overall segmentation performance is not strongly impacted. Remarkably, the downstream prediction of in-distribution (ID) objects is mostly unaffected. 
(3)~Under more global distribution shifts, the ability to separate objects depends significantly on the model and shift at hand, and downstream performance may be severely affected.

As an additional contribution, we provide a library\footnote{\url{https://github.com/addtt/object-centric-library}}
for benchmarking object-centric representation learning, which can be extended with more datasets, methods, and evaluation tasks. We hope this will foster further progress in the learning and evaluation of object-centric representations.

\section{Study design and hypotheses}\label{sec:study_design}

\textbf{Problem definition:} 
Vanilla deep learning architectures learn distributed representations that do not capture the compositional properties of natural scenes---see, e.g., the ``superposition catastrophe''~\cite{von1986thinking,bowers2014neural,greff2020binding}. Even in disentangled representation learning~\cite{higgins2016beta, kim2018disentangling, chen2018isolating, ridgeway2018learning, kumar2017variational, eastwood2018framework}, factors of variations are encoded in a vector representation that is the output of a standard CNN encoder. This introduces an unnatural ordering of the objects in the scene and fails to capture its compositional structure in terms of objects.
Formally defining objects is challenging \citep{greff2020binding} and there is no consensus even outside of machine learning \citep{smith1996origin,green2019theory}. \citet{greff2020binding} put forth three properties for object-centric representations: \emph{separation}, i.e., object features in the set of vectors $r(\xb)$ do not interact with each other, and each object is individually captured in a single element of $r(\xb)$; \emph{common format}, i.e., each element of $r(\xb)$ shares the same representational format; and \emph{disentanglement}, i.e., each element of $r(\xb)$ is represented in a disentangled format that exposes the factors of variation. 
In this paper, we consider representations $r(\xb)$ that are sets of vectors with each element sharing the representational format. 
We take a pragmatic perspective and focus on two clear desiderata for object-centric approaches:

\textbf{Desideratum 1: } \emph{Object embodiment.}
The representation should contain information about the object's location and its embodiment in the scene.
As we focus on unsupervised object discovery, this translates to segmentation masks. This is related to separation and common format, as the decoder is applied to the elements of $r(\xb)$ with shared parameters.

\textbf{Desideratum 2:} \emph{Informativeness of the representation.} Instead of learning disentangled representations of objects, which is challenging even in single-object scenarios \citep{locatello2018challenging}, we want the representation to contain useful information for downstream tasks, not necessarily in a disentangled format. We define objects through their properties as annotated in the datasets we consider, and predict these properties from the representations. Note that this may not be the only way to define objects (e.g., defining faces and edges as objects and deducing shapes as composition of those). The fact that existing models learn informative representations is our first hypothesis (see below).

\textbf{Design principle:} These desiderata offer well-defined quantitative evaluations for object-centric approaches and we want to understand the implications of learning such representations. To this end, we train four different state-of-the-art methods on five datasets, taking hyperparameter configurations from the respective publications and adapting them to improve performance when necessary. Assuming these models succeeded in learning an object-centric representation, we investigate the following hypotheses.

\textbf{Hypothesis 1:} \emph{The unsupervised learning of objects as pretraining task is useful for downstream tasks.}  Existing empirical evaluations largely focus on \emph{Desideratum 1} and measure performance in terms of segmentation metrics. The hope, however, is that the learned representation would be useful for other downstream tasks besides segmentation (\emph{Desideratum 2}). We test this hypothesis by training small downstream models on the frozen object-representations 
to predict the object properties. We match the predictions to the ground-truth properties with the Hungarian algorithm \cite{kuhn1955hungarian} following \citet{locatello2020object}.

\textbf{Hypothesis 2:} 
\emph{In object-centric models, distribution shifts affecting a single object do not affect the representations of other objects.}
A change in the properties of one object in the input should not affect the representation of the other objects. Even OOD objects with previously unseen properties should be segmented correctly by a network that learned the notion of objects~\citep{greff2020binding,scholkopf2021toward}.
We test this hypothesis by (1)~evaluating the segmentation of the scene after the distribution shift, and (2)~training downstream models to predict object properties, and evaluating them on representations extracted from scenes with one OOD object. More specifically, we test changes in the shape, color, or texture of one object.

\textbf{Hypothesis 3:} \emph{Object-centric models are generally robust to distribution shifts, even if they affect global properties of the scene.} Early evidence~\citep{romijnders2021representation} points to the conjecture that learning object-centric representations biases the network towards learning more robust representations of the overall scene. Intuitively, the notion of objects is an additional inductive bias for the network to exploit to maintain accurate predictions if simple global properties of the scene are altered. We test this hypothesis by training downstream models to predict object properties, and evaluating them on representations of scenes with OOD global properties. In this case, we test robustness by cropping, introducing occlusions, and increasing the number of objects.\looseness=-1

\begin{figure}
    \centering
    \captionsetup{belowskip=-15pt}
    \includegraphics[width=\linewidth]{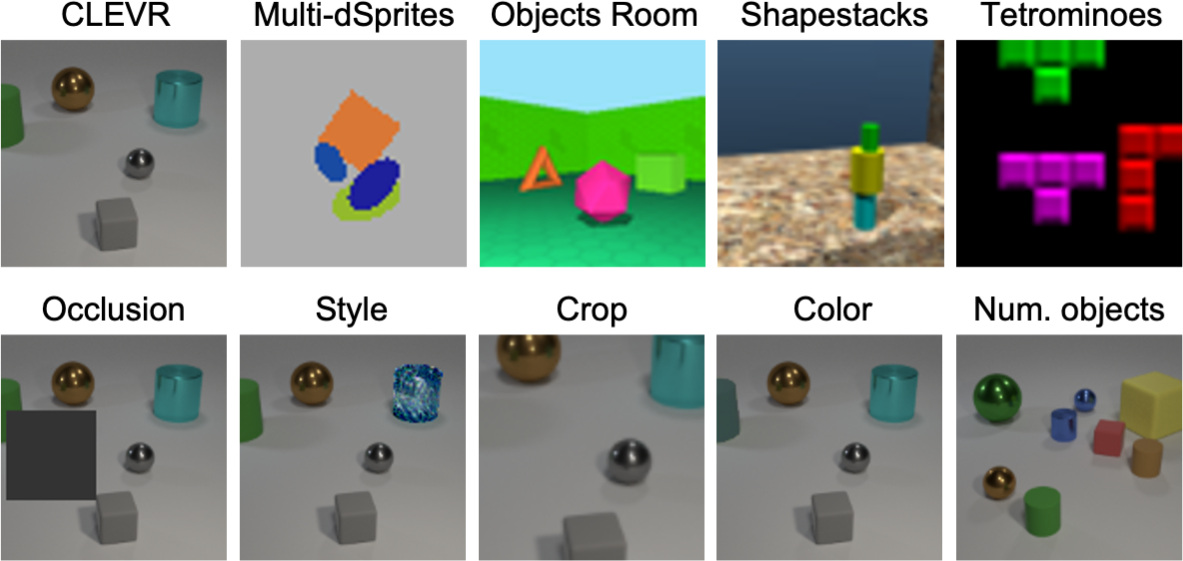}
    \caption{\textbf{Top}: examples from the five datasets in this study. \textbf{Bottom}: distribution shifts in CLEVR.}
    \label{fig:datasets}
\end{figure}

\section{Experimental setup}
\label{sec:experimental_setup}

Here we provide an overview of our experimental setup. After introducing the relevant models and datasets, we outline the evaluation protocols for segmentation accuracy (Desideratum~1) and downstream task performance (Desideratum~2). Then, we discuss the distribution shifts that we use to test robustness---the aforementioned evaluations are repeated once again under these distribution shifts.
We conclude with a discussion on the limitations of this study.

\textbf{Models and datasets.}
We implement four state-of-the-art object-centric models---MONet \cite{burgess2019monet}, GENESIS~\cite{engelcke2020genesis}, Slot Attention~\cite{locatello2020object}, and SPACE~\cite{lin2020space}---as well as vanilla variational autoencoders (VAEs) \citep{kingma2013auto,rezende2014stochastic} as baselines for distributed representations. 
We use one VAE variant with a broadcast decoder \cite{watters2019spatial} and one with a regular convolutional decoder.
See \cref{app:models} for an overview of the models with implementation details.
We then collect five popular multi-object datasets: \emph{Multi-dSprites}, \emph{Objects Room}, and \emph{Tetrominoes} from DeepMind's Multi-Object Datasets collection~\cite{multiobjectdatasets19}, \emph{CLEVR}~\cite{johnson2017clevr}, and \emph{Shapestacks}~\cite{groth2018shapestacks}. The datasets are shown in \cref{fig:datasets} (top row) and described in detail in \cref{app:datasets}.
For each dataset, we define train, validation, and test splits. The test splits, which always contain at least $\num{2000}$ images, are exclusively used for evaluation.
We train each model on all datasets, using 10 random seeds for object-centric models and 5 for each VAE variant, resulting in 250 models in total.

\textbf{Metrics.} We evaluate the segmentation accuracy of object-centric models with the Adjusted Rand Index (ARI)~\cite{hubertComparingPartitions1985}, Segmentation Covering (SC)~\cite{arbelaez2010contour}, and mean Segmentation Covering (mSC)~\cite{engelcke2020genesis}. For all models, we additionally evaluate reconstruction quality via the mean squared error (MSE).
\cref{app:evaluation_metrics} includes detailed definitions of these metrics.

\textbf{Downstream property prediction.}
We evaluate object-centric representations by training downstream models to predict ground-truth object properties from the representations. More specifically, exploiting the fact that object slots share a common representational format, a single downstream model $f$ can be used to predict the properties of each object independently: for each slot representation $\zb_k$ we predict a vector of object properties $\ybpred_k = f(\zb_k)$.
As in previous work on object property prediction~\cite{locatello2020object}, each model simultaneously predicts all properties of an object. For learning, we use the cross-entropy loss for categorical properties and MSE for numerical properties, and denote by $\ell(\ybpred_k, \yb_m)$ the overall loss for a single object, where $\yb_m$ are its ground-truth properties. Here $k \in \{1,\ldots,K\}$ and $m \in \{1,\ldots,M\}$ with $K$ the number of slots and $M$ the number of objects.
In order to optimize the downstream models, the vector $\ybpred_k$ (the properties predicted from the $k$th representational slot) needs to be matched to the ground-truth properties $\yb_m$ of the $m$th object. This is done by computing a $M \times K$ matrix of \emph{matching losses} for each slot--object pair, and then solving the assignment problem using the Hungarian algorithm~\cite{kuhn1955hungarian} to minimize the total matching loss, which is the sum of $\min(M, K)$ terms from the loss matrix.
As matching loss we use either the negative cosine similarity between predicted and ground-truth masks (as in \citet{greff2019multi}), or the downstream loss $\ell(\ybpred_k, \yb_m)$ itself (as in \citet{locatello2020object}). In the following, we will refer to these strategies as \emph{mask matching} and \emph{loss matching}, respectively.
For property prediction, we use 4 different downstream models: a linear model, and MLPs with up to 3 hidden layers of size 256 each.
Given a pretrained object-centric model, we train each downstream model on the representations of $\num{10000}$ images. The downstream models are then tested on $\num{2000}$ held-out images from the test set, which may exhibit distribution shifts as discussed below.
Further details on this evaluation are provided in \cref{app:evaluation_downstream}

\textbf{Evaluating distributed representations.}
Since in non-slot-based models, such as classical VAEs, the representations of the single objects are not readily available, matching representations to objects for downstream property prediction is not trivial.
Although this is an inherent limitation of distributed representations, we are nevertheless interested in evaluating their usefulness.
Using the matching framework presented above, we require the downstream model $f$ to output the predicted properties of \emph{all} objects, and then match these with the true object properties to evaluate prediction quality.
Our downstream model in this case will thus take as input the entire representation $\zb = r(\xb)$ (which is now a single vector rather than a set of vectors) and output the predictions for all objects together as a vector $f(\zb)$. Finally, we split $f(\zb)$ into $K$ vectors $\{\ybpred_k\}_{k=1}^K$, where $K$ loosely corresponds to the number of slots in object-centric models. At this point, we can compute the loss $\ell(\ybpred_k, \yb_m)$ for each pair, as usual.
We now consider two matching strategies: As before, \emph{loss matching} simply defines the matching loss of a slot--object pair as the prediction loss itself. In the \emph{deterministic matching} strategy, following \citet{greff2019multi}, we lexicographically sort objects according to a canonical order of object properties.
Calling $\pi$ the permutation that defines this sorting, the $k$th slot is deterministically matched with the $m$th object, where $m = \pi^{-1}(k)$.

\textbf{Baselines.}
To correctly assess performance on downstream tasks, it is fundamental to compare with sensible baselines. Here we consider as baseline the best performance that can be achieved by a downstream model that outputs a constant vector that does not depend on the image. When predicting properties independently for each object (in slot-based models), the optimal solution is to predict the mean of continuous properties and the majority class for categorical ones. When using deterministic matching in the distributed case, the downstream model can exploit the predefined total order to predict more accurately than random guessing even without using information from the input (this effect is non-negligible only for the properties that are most significant in the order). Finally, in a few cases, loss matching for distributed representations can be significantly better than deterministic matching.\footnote{Intuitively, a (constant) diverse set of uninformed predictions $\{\ybpred_k\}$ might be sufficient for the matching algorithm to find suitable enough objects for most predictions.} 
For simplicity, for both matching strategies in the distributed case, we directly learn a vector $\ybpred$ by gradient descent to minimize the prediction loss. As this depends on random initialization and optimization dynamics, we repeat this for 10 random seeds and report error bars in the plots.\looseness=-1

\textbf{Distribution shifts.}
We test the robustness of the learned representations under two classes of distribution shifts: one where one object goes OOD, and one where global properties of the scene are changed. All such distribution shifts occur at test time, i.e., the unsupervised models are always trained on the original datasets.
To evaluate generalization to distribution shifts affecting a \textit{single object}, we systematically induce changes in the color, shape, and texture of objects.
To change color, we apply a random color shift to one random object in the scene, using the available masks (we do not do this in Multi-dSprites, as the training distribution covers the entire RGB color space).
To test robustness to unseen textures, we apply neural style transfer~\cite{gatys_neural_2016} to one random object in each scene, using \emph{The Great Wave off Kanagawa} as style image.
When either a new color or a new texture is introduced, prediction of material (in CLEVR only) and color is not performed.
To introduce a new shape, we select images from Multi-dSprites that have at most 4 objects (in general, they have up to 5), and add a randomly colored triangle, in a random position, at a random depth in the object stack. In this case, shape prediction does not apply.
Finally, to test robustness to \textit{global} changes in the scene, we change the number of objects (in CLEVR only), introduce occlusions (a gray square at a random location), or crop images at the center and restore their original size via bilinear interpolation.
See \cref{fig:datasets} for examples, and \cref{app:evaluation_shifts} for further details.

\textbf{Limitations of this study.}
While we aim to conduct a sound and informative experimental study to answer the research questions from \cref{sec:experimental_setup}, inevitably there are limitations regarding datasets, models, and evaluations.
Although the datasets considered here vary significantly in complexity and visual properties, they all consist of synthetic images where object properties are independent of each other and independent between objects.
Regarding object-centric models, we only focus on autoencoder-based approaches that model a scene as a mixture of components.
As official implementations are not always available, and none of the methods in this work has been applied to all the datasets considered here, we re-implement these methods and choose hyperparameters following a best-effort approach.
Finally, we only consider the downstream task of object property prediction, and assess generalization using only a few representative single-object and global distribution shifts.

\section{Results}

In this section, we highlight our findings with plots that are representative of our main results. The full experimental results are presented in \cref{app:results}. In \cref{sec:results/metrics} we focus on the different evaluation metrics and the performance we obtained re-training the methods considered in this study. We then focus on our three hypotheses in \cref{sec:results/hyp1,sec:results/hyp2,sec:results/hyp3}.

\subsection{Learning and evaluating object discovery}
\label{sec:results/metrics}

    \begin{figure}
        \centering
        \captionsetup{belowskip=-14pt}
        \includegraphics[width=\linewidth,trim={10pt 10pt 5pt 10pt}]{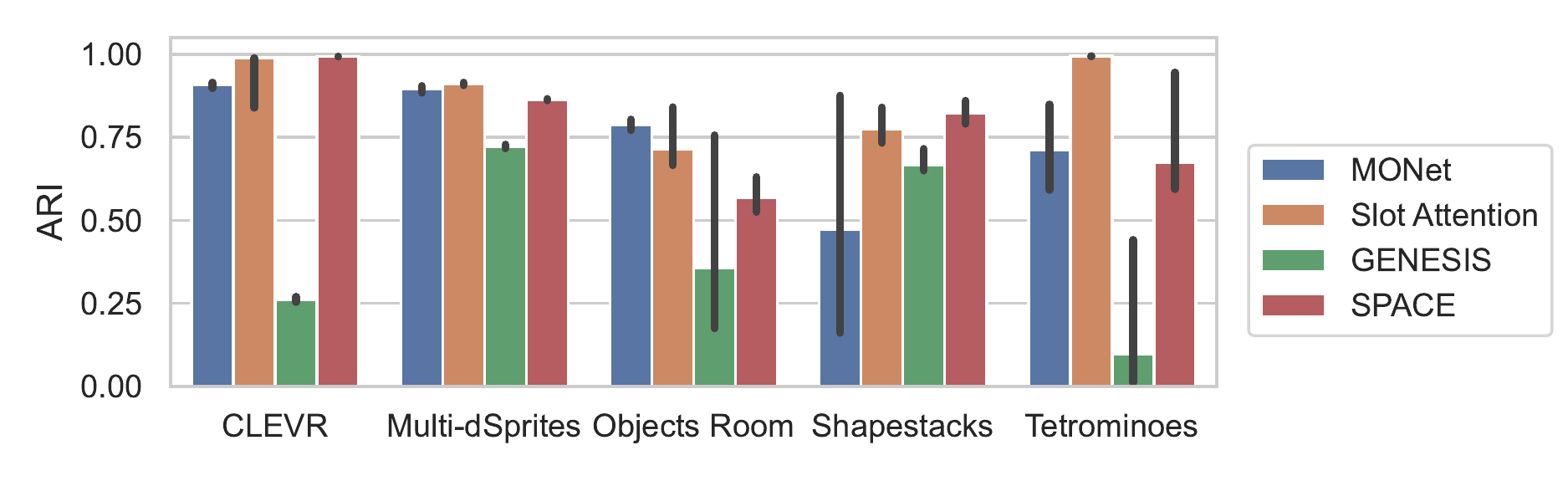}
        \caption{ARI of all models and datasets on $\num{2000}$ test images. Medians and 95\% confidence intervals with 10 seeds.}
        \label{fig:main_text/indistrib/metrics/box_plots}
    \end{figure}

Since all methods included in our study were originally evaluated only on a subset of the datasets and metrics considered here, we first test how well these models perform.

\cref{fig:main_text/indistrib/metrics/box_plots} shows the segmentation performance of the models in terms of ARI across models, datasets, and random seeds. \cref{fig:indistrib/metrics/box_plots} in \cref{app:results} provides an overview of the reconstruction MSE and all segmentation metrics. Although these results are in line with published work, we observe substantial differences in the ranking between models depending on the metric. This indicates that, in practice, these metrics are not equivalent for measuring object discovery.

\begin{figure}
        \centering
        \captionsetup{belowskip=-8pt}
        \includegraphics[width=\linewidth,trim={9pt 10pt 9pt 10pt}]{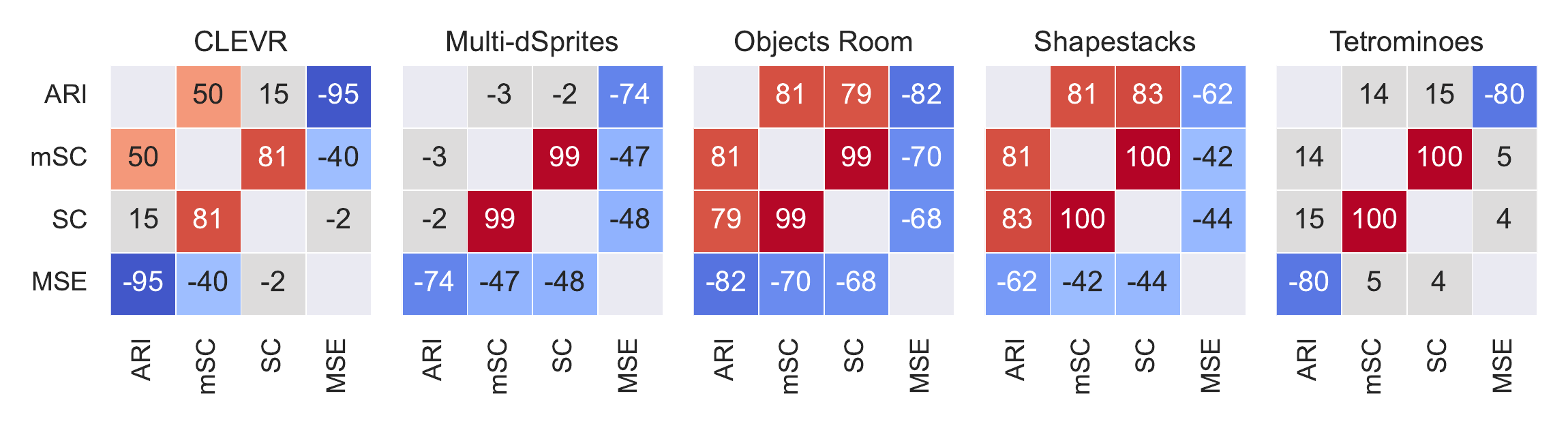}
        \caption{Spearman rank correlations between evaluation metrics across models and random seeds (color-coded only when p\textless 0.05).}
        \label{fig:indistrib_correlation/metrics_metrics}
    \end{figure}

This is confirmed in \cref{fig:indistrib_correlation/metrics_metrics}, which shows rank correlations between metrics on different datasets (aggregating over different models).
We also observe a strong negative correlation between ARI and MSE across models and datasets, suggesting that models that learn to more accurately reconstruct the input tend to better segment objects according to the ARI score.
This trend is less consistent for the other segmentation metrics, as MSE significantly correlates with mSC in only three datasets (Multi-dSprites, Objects Room, and CLEVR), and with SC in two (Multi-dSprites and Objects Room).
SC and mSC measure very similar segmentation notions and therefore are significantly correlated in all datasets, although to a varying extent. However, they correlate with ARI only on two and three datasets, respectively (the same datasets where they correlate with the MSE).

\textbf{Summary:} We observe strong differences in performance and ranking between the models depending on the evaluation metric. In the tested datasets, we find that the ARI, which requires ground-truth segmentation masks to compute, correlates particularly well with the MSE, which is unsupervised and provides training signal.

\subsection{Usefulness for downstream tasks (Hypothesis 1)}
\label{sec:results/hyp1}
To test Hypothesis 1, we first evaluate whether frozen object-centric representations can be used to train downstream models measuring {Desideratum 2} from \cref{sec:study_design}. As discussed in \cref{sec:experimental_setup}, this type of downstream task requires matching the true object properties with the predictions of the downstream model. In the following, we will only present results obtained with \emph{loss matching}, and show results for other matching strategies in \cref{app:results}.

    \begin{figure*}
        \centering
        \includegraphics[width=\linewidth]{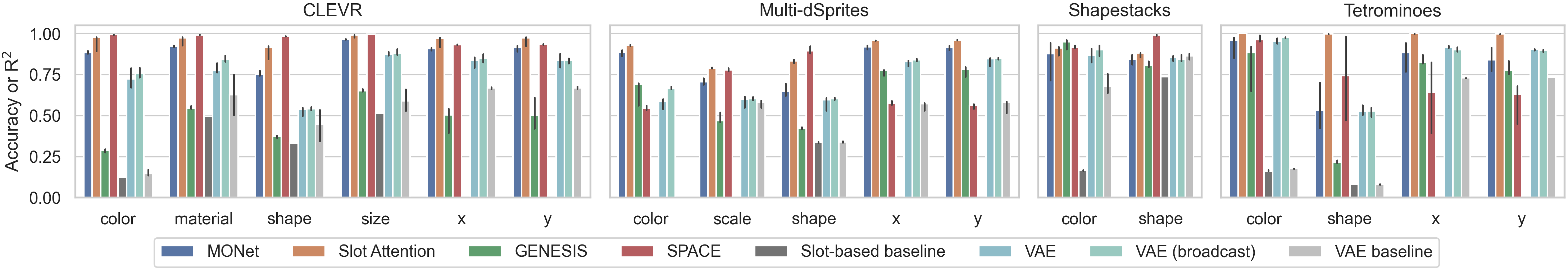}
        \caption{Comparison of downstream property prediction performance for object-centric (slot-based) and distributed (VAE) representations, using an MLP with one hidden layer as downstream model. The metric is accuracy for categorical properties or \rsq for numerical ones. The baselines in gray indicate the best performance that can be achieved by a model that outputs a constant vector that does not depend on the input. The bars show medians and 95\% confidence intervals with 10 random seeds.}
        \label{fig:indistrib/downstream/barplots_loss_linear_main}
    \end{figure*}

\cref{fig:indistrib/downstream/barplots_loss_linear_main} shows downstream prediction performance on all datasets and models, when the downstream model is a single-layer MLP.
Although results vary across datasets and models, accurate prediction of object properties seems to be possible in most of the scenarios considered here.
\cref{fig:indistrib/downstream/box_plots} in \cref{app:results} shows similar results when using a linear model or MLPs with up to 3 hidden layers.

In \cref{fig:indistrib/downstream/barplots_loss_linear_main}, we also compare the downstream prediction performance from object-centric and distributed representations. We observe that VAE representations tend to achieve lower scores in downstream prediction, although not always by a large margin. In particular, color and size in CLEVR and color in Tetrominoes are predicted relatively well, and significantly better than the baseline. On the other hand, in many cases where VAE representations perform well, they have in fact a considerable advantage if we take the baselines into account (scale in Multi-dSprites, color in Shapestacks, x~and~y in CLEVR, Multi-dSprites, and Tetrominoes). Moreover, performance from distributed representations often does not improve significantly when using a larger downstream model (see \cref{fig:indistrib/downstream/barplots_compare_downstream_models_loss}).
In conclusion, although the two classes of representations are difficult to compare on this task, these results suggest that the quantities of interest are present in the VAE representations, but they appear to be less explicit and less easily usable.

    \begin{figure}
        \centering
        \captionsetup{belowskip=-10pt}
        \includegraphics[width=\linewidth,trim={10pt 10pt 10pt 10pt}]{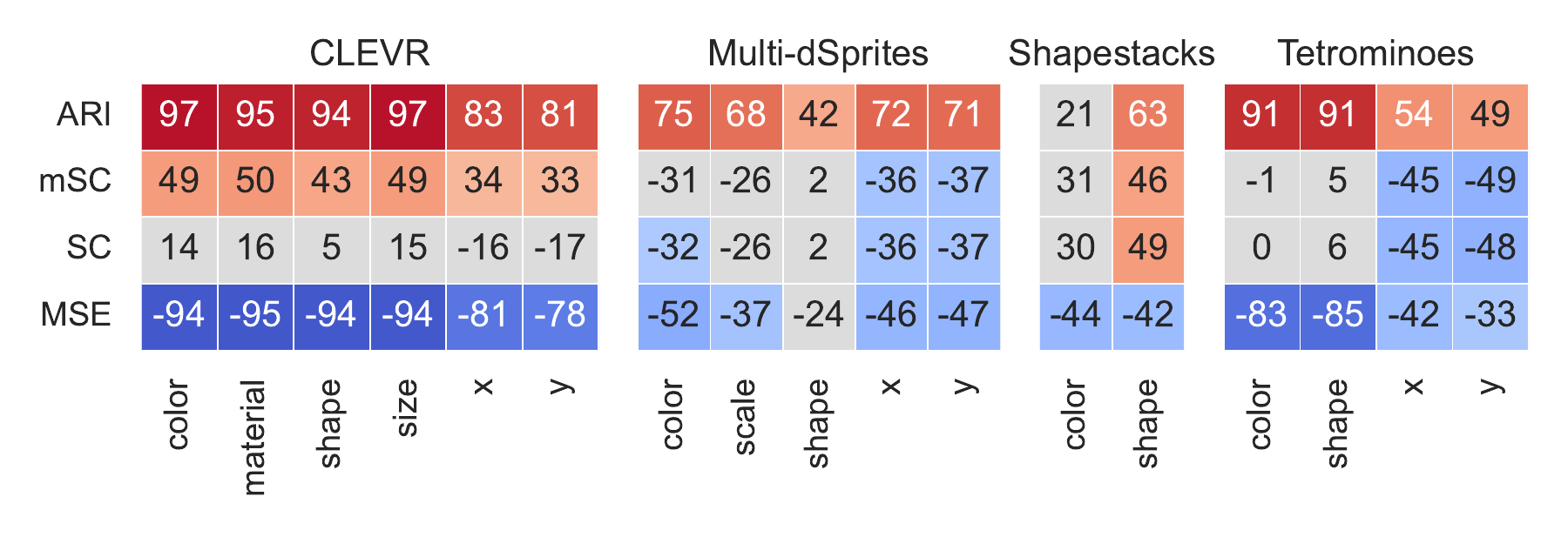}
        \caption{Spearman rank correlations between evaluation metrics and downstream performance with an MLP. The correlations are color-coded only when p\textless 0.05.}
        \label{fig:indistrib_correlations/metrics_downstream}
    \end{figure}

Finally, we investigate the relationship between downstream performance and evaluation metrics.
\cref{fig:indistrib_correlations/metrics_downstream} shows the Spearman rank correlation of the segmentation and reconstruction metrics with the test performance of downstream predictors.
For all datasets and 
object 
properties, downstream performance is strongly correlated with the ARI.
On the other hand, SC and mSC exhibit inconsistent trends across datasets.
Models that correctly separate objects according to the ARI are therefore useful for downstream object property prediction, confirming Hypothesis 1.
Downstream prediction performance is also significantly correlated with the reconstruction MSE in all datasets. This is not particularly surprising, since the representation of a model that cannot properly reconstruct the input might not contain the information necessary for property prediction. However, the correlation is generally stronger with the ARI than with the MSE, suggesting that having a notion of objects is more important for downstream tasks than reconstruction accuracy. This is consistent with the findings by \citet{papa2022inductive}, where the ARI still correlates strongly with downstream performance when objects have complex textures, while the MSE does not. When segmentation masks are available for validation, ARI should therefore be the preferred metric to select useful representations for downstream tasks.
\cref{fig:indistrib/downstream/corr_metrics_downstream_all} in \cref{app:results} shows analogous results for mask matching and for the three other downstream models---these results are broadly similar, except that correlations with ARI tend to be stronger when using mask matching (perhaps unsurprisingly) or larger downstream models.

\textbf{Summary:}
Models that accurately segment objects allow for good downstream prediction performance.
Despite often having an advantage, distributed representations generally perform worse,
but not always significantly: the information is present but less easily accessible. 
The ARI is consistently correlated with downstream performance, and is therefore useful for model selection when masks are available. The MSE can be a practical unsupervised alternative on these datasets, but it may be less robust on complex textures.

{
\spacedstyle{-10}
\subsection{Generalization with one OOD object (Hypothesis 2)}
\label{sec:results/hyp2}
}
To test Hypothesis 2, we construct settings where a single object is OOD and the others are ID. We change the object style with neural style transfer, change the color of one object at random (only in CLEVR, Tetrominoes, and Shapestacks), or introduce a new shape (only in Multi-dSprites).
The unsupervised models are always trained on the original datasets. Then we train downstream models to predict the object properties from the learned representations. 
We consider two scenarios for this task: (1)~train the predictors on the original datasets and test them on the variants with a modified object, (2)~train and test the predictors on each variant.
In both cases, we test the predictors on representations that might be inaccurate, because the representation function (encoder) is OOD. However, since in case (2) the downstream model is \emph{trained} under distribution shift, this experiment quantifies the extent to which the representation can still be used by a downstream task that is allowed to adapt to the shift---although the representation might no longer represent objects faithfully, it could still contain useful information.\looseness=-1

    \begin{figure*}
        \centering
        \captionsetup{belowskip=0pt}
        \includegraphics[width=\linewidth,trim={4pt 8pt 4pt 8pt}]{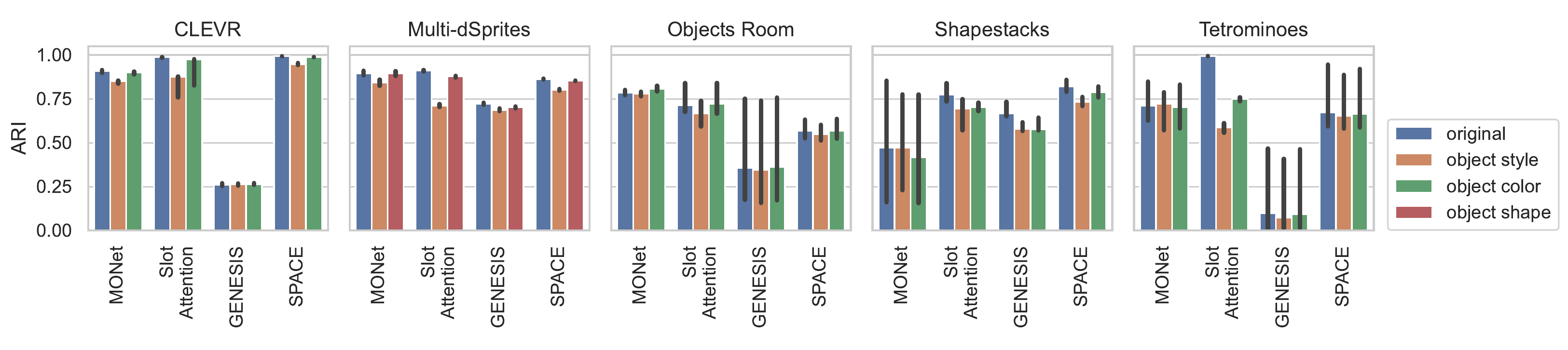}
        \caption{Effect of single-object distribution shifts on the ARI. Medians and 95\% confidence intervals with 10 random seeds.}
        \label{fig:main_text/ood/metrics/generalization_compositional/box_plots}
    \end{figure*}

For {Desideratum 1}, we observe in \cref{fig:main_text/ood/metrics/generalization_compositional/box_plots} that the models are generally robust to distribution shifts affecting a single object.
Introducing a new color or a new shape typically does not affect segmentation quality (but note Slot Attention on Tetrominoes), while changing the texture of an object via neural style transfer leads to a moderate drop in ARI in some cases.
In \cref{fig:ood/metrics/generalization_compositional/box_plots} (\cref{app:results}) we observe that SC and mSC show a compatible but less pronounced trend, while the MSE more closely mirrors the ARI. 
We conclude that the encoder is still partially able to separate objects when one object undergoes a distribution shift at test time.

    \begin{figure}
        \centering
        \captionsetup{belowskip=-10pt}
        \includegraphics[width=\linewidth,trim={0pt 5pt 0pt 8pt}]{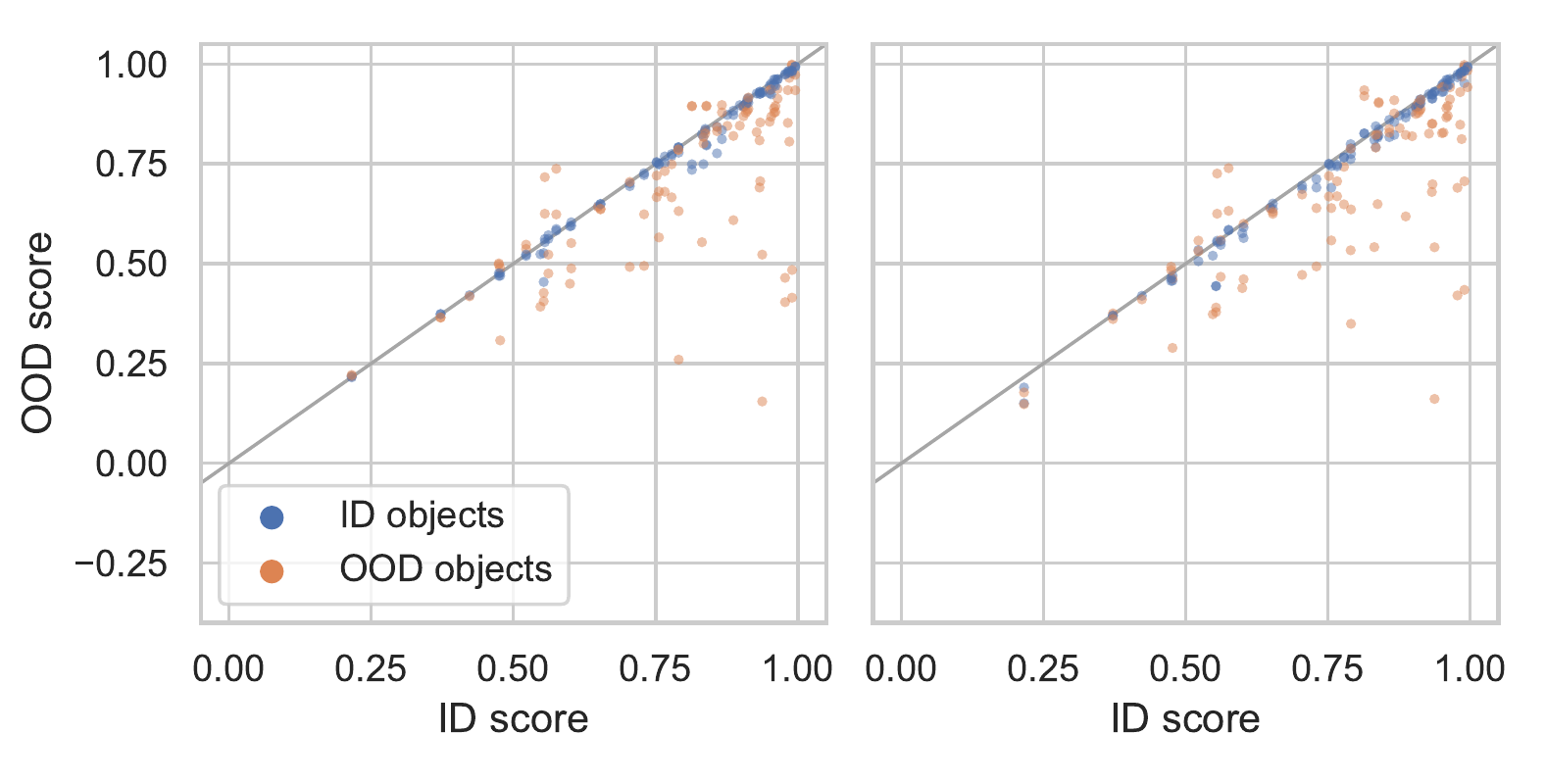}
        \caption{\textbf{ID vs OOD downstream performance} with single-object distribution shifts. All datasets, models, and object properties are shown. Metrics: accuracy for categorical attributes, \rsq for numerical attributes. The downstream model (an MLP with one hidden layer) is tested zero-shot out-of-distribution (left) or retrained after the distribution shift has occurred (right).}
        \label{fig:main_text/ood/downstream/generalization_compositional/scatter}
    \end{figure}

For {Desideratum 2}, we observe in \cref{fig:main_text/ood/downstream/generalization_compositional/scatter} (left) that property prediction performance for objects that underwent distribution shifts (color, shape, or texture) is often significantly worse than in the original dataset, whereas the prediction of ID objects is largely unaffected.
This is in agreement with Hypothesis 2: changes to one object do not affect the representation of other objects, even when these objects are OOD.
Extensive results, including further splits and all downstream models, are shown in \cref{fig:ood/downstream/slotted/match_loss/object/retrain_False} in \cref{app:results}.
On the right plot in \cref{fig:main_text/ood/downstream/generalization_compositional/scatter}, we observe that retraining the downstream models after the distribution shifts does not lead to significant improvements. This suggests that the shifts introduced here negatively affect not only the downstream model, but also the representation itself.
This result also holds with different downstream models and with mask matching (see \cref{fig:ood/downstream/slotted/match_loss/object/retrain_True,fig:ood/downstream/slotted/match_mask/object/retrain_True}). 
While in principle we observe a similar trend for VAEs (see e.g. \cref{fig:ood/downstream/vae/match_loss/object/retrain_False,fig:ood/downstream/vae/match_loss/object/retrain_True} in \cref{app:results}), their performance is often too close to the respective baseline (\cref{fig:indistrib/downstream/barplots_loss_linear_main}) for a definitive conclusion to be drawn.\looseness=-1

\textbf{Summary:}
The models are generally robust to distribution shifts affecting a single object.
Downstream prediction is largely unaffected for ID objects, but may be severely affected for OOD objects.
Finally, there seems to be no clear benefit in retraining downstream models after the shifts, indicating that the deteriorated representations cannot easily be adjusted \emph{post hoc}.

\subsection{Robustness to global shifts (Hypothesis 3)}
\label{sec:results/hyp3}
Finally, we investigate the robustness of object-centric models to transformations changing the global properties of a scene at test time.
Here, we consider variants of the datasets with occlusions, cropping, or more objects (only on CLEVR).
We train downstream predictors on the original datasets and report their test performance on the dataset variants with global shifts. As before, we also report results of downstream models retrained on the OOD datasets.

    \begin{figure*}
        \centering
        \captionsetup{belowskip=-4pt}
        \includegraphics[width=\linewidth,trim={4pt 5pt 4pt 8pt}]{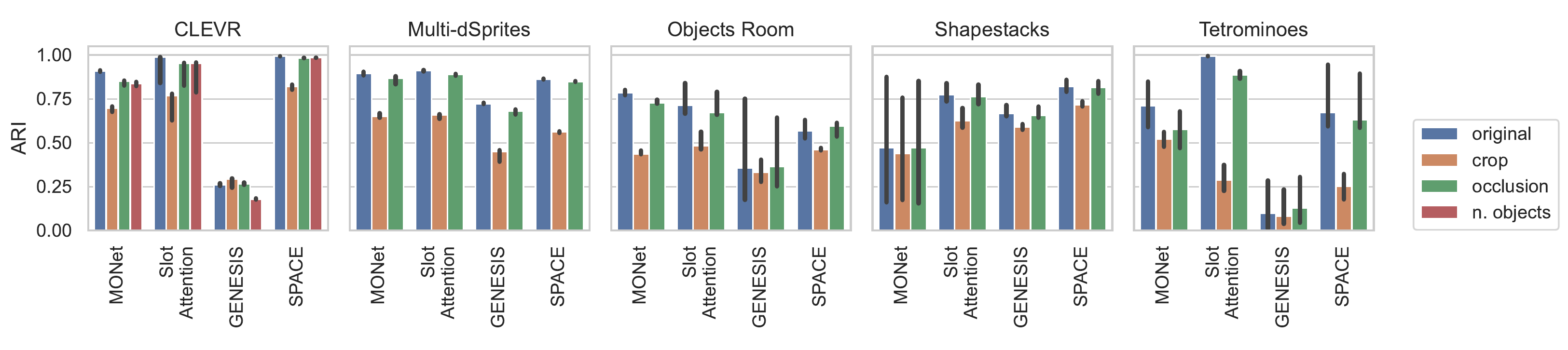}
        \caption{Effect of distribution shifts on global scene properties on the ARI. Medians and 95\% confidence intervals with 10 seeds.}
        \label{fig:figures/ood/metrics/only_ARI_ood_global/barplots_ARI}
    \end{figure*}

For {Desideratum 1}, \cref{fig:figures/ood/metrics/only_ARI_ood_global/barplots_ARI} shows that segmentation quality is generally only marginally affected by occlusion, but cropping often leads to a significant degradation. In CLEVR, the effect on the ARI of increasing the number of objects is comparable to the effect of occlusions, which suggests that learning about objects is useful for this type of systematic generalization.
These trends persist when considering SC and mSC, but appear less pronounced and less consistent across datasets (see \cref{fig:ood/metrics/generalization_global/box_plots} in \cref{app:results} for detailed results).
As might be expected, when the number of objects is increased in CLEVR, the MSE increases more conspicuously for VAEs than for object-centric models (\cref{fig:ood/metrics/generalization_global/box_plots}, bottom left), likely due to their explicit modeling of objects. However, \cref{fig:ood_visualizations_clevr} shows that VAEs may, in fact, generalize relatively well to an unseen number of objects, although not nearly as well as some object-centric models. 

    \begin{figure}
        \centering
        \captionsetup{belowskip=-10pt}
        \includegraphics[width=\linewidth]{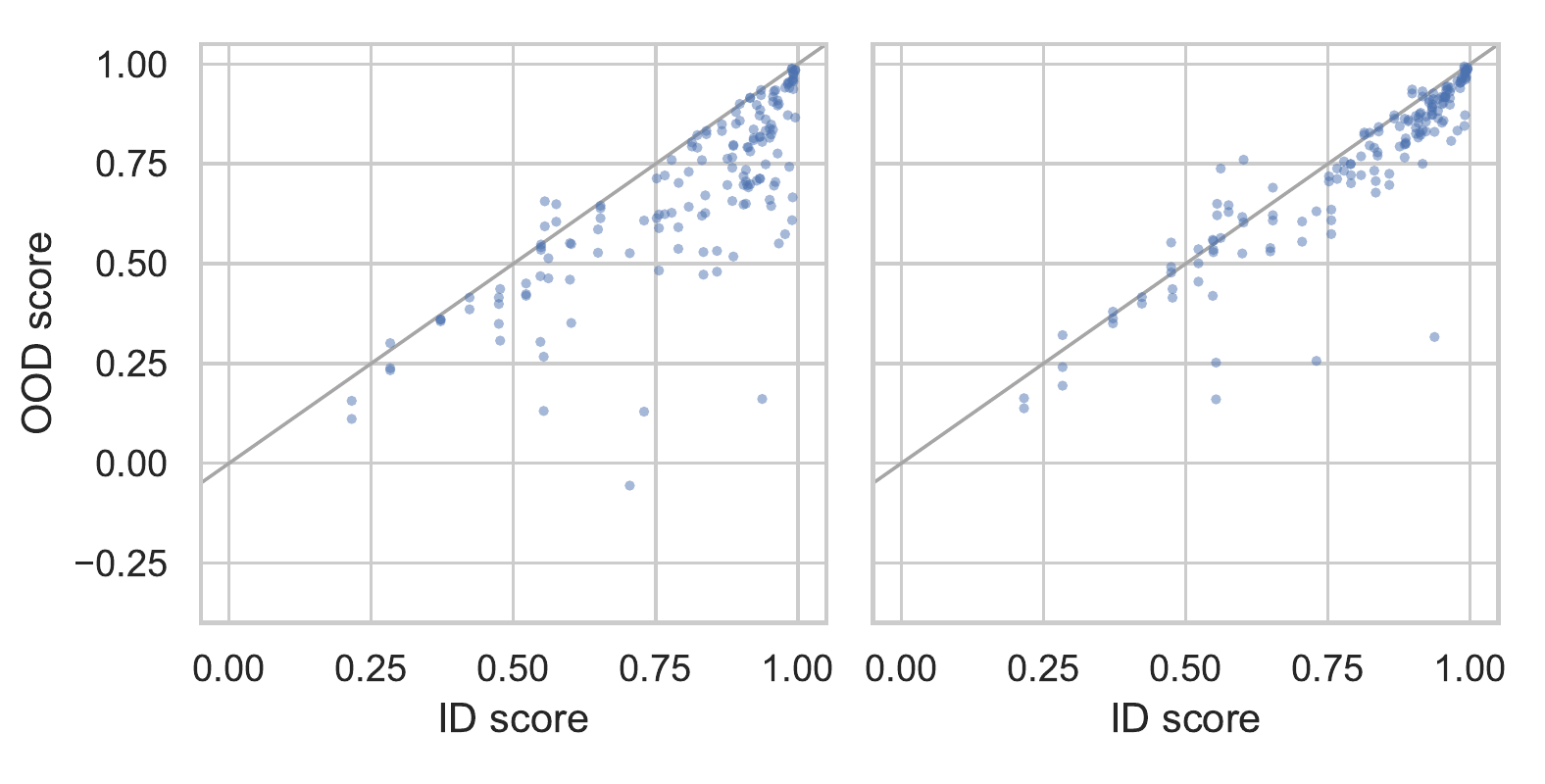}
        \caption{\textbf{ID vs OOD downstream performance} with global distribution shifts. All datasets, models, and object properties are shown. Metrics: accuracy for categorical attributes, \rsq for numerical attributes. The downstream model (an MLP with one hidden layer) is tested zero-shot out-of-distribution (left) or retrained after the distribution shift has occurred (right).}
        \label{fig:main_text/ood/downstream/generalization_global/scatter}
    \end{figure}
    
For {Desideratum 2}, we train a downstream model on the original dataset and test it under global distribution shifts. These shifts generally have a negative effect on downstream property prediction (\cref{fig:main_text/ood/downstream/generalization_global/scatter}, left), although this is comparable to the effect on OOD objects when only one object is OOD. This is in agreement with the observation made in \cref{sec:results/hyp2} that these shifts negatively affect the representation, which is no longer accurate because the encoder is OOD (cf. the ``OOD2'' scenario in \citet{dittadi2020transfer}).
When retraining the downstream models on the OOD datasets while keeping the representation frozen, the performance improves slightly but does not reach the corresponding results on the training distribution (\cref{fig:main_text/ood/downstream/generalization_global/scatter}, right), as in \cref{sec:results/hyp2}. 
These observations also hold for different downstream models and with mask matching (\cref{fig:ood/downstream/slotted/match_loss/global/retrain_False,fig:ood/downstream/slotted/match_mask/global/retrain_False,fig:ood/downstream/slotted/match_loss/global/retrain_True,fig:ood/downstream/slotted/match_mask/global/retrain_True}), as well as for distributed representations (see, e.g., \cref{fig:ood/downstream/vae/match_loss/global/retrain_False/scatter}) although with similar caveats as in \cref{sec:results/hyp2}.

\textbf{Summary:}
The impact of global distribution shifts on the segmentation capability of object-centric models depends on the chosen shift; e.g., cropping consistently has a significant effect. Moreover, the usefulness for downstream tasks decreases substantially in many cases, and the performance of downstream prediction models cannot be satisfactorily recovered by retraining them.

\section{Other related work}

Recent years have seen a number of systematic studies on disentangled representations~\citep{locatello2018challenging,locatello2019fairness,van2019disentangled,trauble2020independence}, some of which focusing on their effect on generalization~\citep{gondal2019transfer,dittadi2020transfer,montero2021the,pmlr-v89-esmaeili19a,traeuble2022role}.
In the context of object-centric learning, \citet{engelcke2020reconstruction} investigate their reconstruction bottlenecks to understand how these models can separate objects from the input in an unsupervised manner. In contrast, we specifically test some key implications of learning object-centric representations.

Slot-based object-centric models can be classified according to their approach to separating the objects at a representational level~\cite{greff2020binding}.
In models that use \emph{instance slots}~\cite{greff_tagger_2016,greff2017neural,van2018relational,greff2019multi,locatello2020object,lowe2020learning, huang2020better,le2011learning,goyal2019recurrent,van2020investigating,NEURIPS2019_32bbf7b2,yang2020learning,kipf2019contrastive,racah2020slot,kipf2021conditional}, each slot is used to represent a different part of the input. This introduces a routing problem, because all slots are identical but they cannot all represent the same object, so a mechanism needs to be introduced to allow slots to communicate with each other.
In models based on \emph{sequential slots}~\cite{eslami2016air,stelzner2019supair,kosiorek2018sequential,stove,burgess2019monet,engelcke2020genesis,engelcke2021genesisv2,vonkugelgen2020econ}, the representational slots are computed in a sequential fashion, which solves the routing problem and allows to dynamically change the number of slots, but introduces dependencies between slots.
In models based on \emph{spatial slots}~\cite{nash2017multi,crawford2019spatially,jiang2020scalor,crawford2020exploiting,lin2020space,deng2021generative,lin2020improving,dittadi2019lavae}, a spatial coordinate is associated with each slot, introducing a dependency between slot and spatial location. 
In this work, we focus on four scene-mixture models as representative examples of approaches based on instance slots (Slot Attention), sequential slots (MONet and GENESIS), and spatial slots (SPACE). 

\section{Conclusions}

In this paper, we identify three key hypotheses in object-centric representation learning: learning about objects is useful for downstream tasks, it facilitates strong generalization, and it improves overall robustness to distribution shifts.
To investigate these hypotheses,
we re-implement and systematically evaluate four state-of-the-art unsupervised object-centric learners on a suite of five common multi-object datasets.
We find that object-centric representations are generally useful for downstream object property prediction, and downstream performance is strongly correlated with segmentation quality and reconstruction error.
Regarding generalization, we observe that when a single object undergoes distribution shifts the overall segmentation quality and downstream performance for in-distribution objects is largely unaffected.
Finally, we find that object-centric models can still relatively robustly separate objects even under global distribution shifts. However, this may depend on the specific shift, and downstream performance appears to be more severely affected.

An interesting avenue for future work is to continue our systematic investigation of object-centric learning on more complex data with diverse textures, as well as a wide range of more challenging downstream tasks. 
Furthermore, it would be interesting to compare object-centric and non-object-centric models more fairly: while learning about objects offers clear advantages, the full potential of distributed representations in this context is still not entirely clear, particularly when scaling up datasets and models.
Finally, while we limit our study to unsupervised object discovery, it would be relevant to consider methods that leverage some form of supervision when learning about objects.
We believe our benchmarking library will facilitate progress along these and related lines of research.

\section*{Acknowledgements}
We would like to thank Thomas Brox, Dominik Janzing, Sergio Hernan Garrido Mejia, Thomas Kipf, and Frederik Träuble for useful comments and discussions, and the anonymous reviewers for valuable feedback.

\bibliography{references}
\bibliographystyle{icml2022}

\newpage
\appendix
\onecolumn


\clearpage
\section{Models}
\label{app:models}

In this section, we give an informal overview of the models included in this study and provide details on the implementation and hyperparameter choices.

\subsection{Overview of the models}

\paragraph{MONet.} In MONet~\cite{burgess2019monet}, attention masks are computed by a recurrent segmentation network that takes as input the image and the current \emph{scope}, which is the still unexplained portion of the image. For each slot, a variational autoencoder (the \emph{component VAE}) encodes the full image and the current attention mask, and then decodes the latent representation to an image reconstruction and mask. The reconstructed images are combined using the \emph{attention} masks (\emph{not} the masks decoded by the component VAE) into the final reconstructed image. 
The reconstruction loss is the negative log-likelihood of a spatial Gaussian mixture model (GMM) with one component per slot, where each pixel is modeled independently. The overall training loss is a (weighted) sum of the reconstruction loss, the KL divergence of the component VAEs, and an additional mask reconstruction loss for the component VAEs.

\paragraph{GENESIS.} Similarly to MONet, GENESIS~\cite{engelcke2020genesis} models each image as a spatial GMM. The spatial dependencies between components are modeled by an autoregressive prior distribution over the latent variables that encode the mixing probabilities.
From the image, an encoder and a recurrent network are used to compute the latent variables that are then decoded into the mixing probabilities. The mixing probabilities are pixel-wise and can be seen as attention masks for the image. Each of these is concatenated with the original image and used as input to the component VAE, which finds latent representations and reconstructs each scene component. These are combined using the mixing probabilities to obtain the reconstruction of the image.
While in MONet the attention masks are computed by a deterministic segmentation network, GENESIS defines an autoregressive prior on latent codes that are decoded into attention masks. GENESIS is therefore a proper probabilistic generative model, and it is trained by maximizing a modification of the ELBO introduced by \citet{rezende2018taming}, which adaptively trades off the likelihood and KL terms in the ELBO.

\paragraph{Slot Attention.} As our focus is on the object discovery task, we use the autoencoder model proposed in the Slot Attention paper~\cite{locatello2020object}.
The encoder consists of a CNN followed by the Slot Attention module, which maps the feature map to a set of {slots} through an iterative refinement process. 
At each iteration, dot-product attention is computed with the input vectors as keys and the current slot vectors as queries. The attention weights are then normalized over the slots, introducing competition between the slots to explain the input. Each slot is then updated using a GRU that takes as inputs the current slot vectors and the normalized attention vectors. After the refinement steps, the slot vectors are decoded into the appearance and mask of each object, which are then combined to reconstruct the entire image. The model is optimized by minimizing the MSE reconstruction loss.
While MONet and GENESIS use sequential slots to represent objects, Slot Attention employs instance slots.

\paragraph{SPACE.}
Spatially Parallel Attention and Component Extraction (SPACE) \cite{lin2020space} combines the approaches of scene-mixture models and spatial attention models. The foreground objects are segregated using bounding boxes computed through a parallel spatial attention process. The parallelism allows for a larger number of bounding boxes to be processed compared to previous related approaches. The background elements are instead modeled by a mixture of components. The use of bounding boxes for the foreground objects could lead to under- or over-segmentation if the size of the bounding box is not tuned appropriately. An additional boundary loss tries to address the over-segmentation issue by penalizing splitting objects across bounding boxes.

\paragraph{VAE baselines.}
We train variational autoencoders (VAEs)~\cite{kingma2013auto,rezende2014stochastic} as baselines that learn distributed representations. Following \citet{greff2019multi}, we use two different decoder architectures: one consisting of an MLP followed by transposed convolutions, and one where the MLP is replaced by a broadcast decoder~\cite{watters2019spatial}.
The VAEs are trained by maximizing the usual variational lower bound (ELBO).

\subsection{Implementation details}

\begin{table}
    \centering
    \caption{Datasets used for quantitative and/or qualitative evaluation in the publications corresponding to the four object-centric models considered in this study. Here we train and evaluate all models on all datasets.}
    \label{tab:app_models_x_datasets}
    \begin{tabular}{lccccc}
    \toprule
                   & CLEVR      & Multi-dSprites & Objects Room & Shapestacks & Tetrominoes\\
   \midrule
    MONet          & \checkmark & \checkmark$^*$ & \checkmark   &             &          \\
    Slot Attention & \checkmark & \checkmark\phantom{$^*$} &              &             &\checkmark\\
    GENESIS        &            & \checkmark$^*$ & \checkmark   & \checkmark  &          \\
    SPACE          &            &                &              &             &          \\
    \midrule 
    \multicolumn{6}{c}{\small $^*$These publications use a variant of Multi-dSprites with colored background as opposed to grayscale.}
    \\ \bottomrule
    \end{tabular}
\end{table}

We implement our library in PyTorch \cite{paszke2019pytorch}. All models are either re-implemented or adapted from available code, and quantitative results from the literature are reproduced, when available.
As shown in \cref{tab:app_models_x_datasets}, all methods included in our study were originally evaluated only on a subset of the datasets considered in our study. Thus, the recommended hyperparameters for a given model are likely to be suboptimal in the datasets on which such model was not evaluated. When a model performed particularly bad on a dataset, we attempted to find better hyperparameter values for the sake of the soundness of our study.
We provide implementation and training details for each model below.

\paragraph{MONet.}

We re-implement MONet following the implementation details in \citet{burgess2019monet}. In order to make this model work satisfactorily on Shapestacks and Tetrominoes---the two datasets where MONet was not originally tested---we ran a grid search over hyperparameters on both datasets, as follows:
\begin{itemize}[topsep=0pt,itemsep=0pt,partopsep=0pt,parsep=\parskip]
    \item Optimizer: Adam or RMSprop, both with default PyTorch parameters.
    \item $\beta \in \{0.1, 0.5\}$.
    \item Learning rate in \{3e-5, 1e-4\}.
    \item $(\sigma_{\mathrm{bg}}, \sigma_{\mathrm{fg}}) \in \{(0.06, 0.1), (0.12, 0.18), (0.2, 0.24), (0.25, 0.3), (0.3, 0.36)\}$.
\end{itemize}
A summary of the final hyperparameter choices is shown in \cref{tab:app_models_monet_parameters}.

\begin{table}
\centering
\caption{Overview of the main hyperparameter values for MONet. When dataset-specific values are not given, the defaults are used.}
\label{tab:app_models_monet_parameters}
\begin{tabular}{lcccc}
\toprule
\multicolumn{1}{c}{\multirow{2}{*}{\textbf{Hyperparameter}}} & \multirow{2}{*}{\textbf{Default value}} & \multicolumn{3}{c}{\textbf{Dataset-specific values}}                         \\ \cmidrule(lr){3-5} 
\multicolumn{1}{c}{}                                &                                & \textbf{CLEVR}                & \textbf{Shapestacks}          & \textbf{Tetrominoes}          \\ \midrule
Optimizer                                           & Adam                           & RMSprop              & RMSprop              & ---                  \\
Learning rate                                       & 1e-4                   & 3e-5         & ---                  & ---                  \\
Batch size                                          & 64                             & 32                   & ---                  & ---                  \\
Training steps                                      & 500k                    & ---                  & ---                  & ---                  \\
$\sigma_\mathrm{bg}$                                & 0.06                           & ---                  & 0.2                  & 0.3                  \\
$\sigma_\mathrm{fg}$                                & 0.1                            & ---                  & 0.24                 & 0.36                 \\
$\beta$                                             & 0.5                            & ---                  & 0.1                  & ---                  \\
$\gamma$                                            & 0.5                            & ---                  & ---                  & ---                  \\
Latent space size                                   & 16                             & ---                  & ---                  & ---                  \\
U-Net blocks                                        & 5                              & 6                    & ---                  & 4                    \\ \bottomrule
\end{tabular}
\end{table}

\paragraph{Slot Attention.}
We re-implement the Slot Attention autoencoder based on the official TensorFlow implementation and the corresponding publication \cite{locatello2020object}. 
We mostly use the recommended hyperparameter values and learning rate schedule.
On Objects Room and Shapestacks, we use the same parameters as for Multi-dSprites, which has the same resolution.
On CLEVR, we make a few changes to accommodate the larger image size. For the decoder, we follow the approach in \citet{locatello2020object} and use the broadcast decoder from a broadcasted shape of $8\times 8$ rather than $128\times 128$, and use four times a stride of 2 in the decoder.
For the encoder, we follow the set prediction architecture in \citet{locatello2020object} and use two strides of 2 in the encoder.
Finally, we use a batch size of 32 rather than 64.

\paragraph{GENESIS.}

We re-implement GENESIS based on the official implementation and the corresponding publication \cite{engelcke2020genesis}, and use the recommended hyperparameter values.
On Objects Room, we use the same hyperparameters as described in the paper for Multi-dSprites and Shapestacks, which have the same resolution. 
On CLEVR, which has $128 \times 128$ images, we use an additional stride of $2$ in the convolutional layer at the middle of both encoder and decoder (the output padding in the decoder is adjusted accordingly). In this case we also reduce the batch size from 64 to 32.
On Tetrominoes ($32\times 32$ images), we change the first stride in the encoder and the last stride in the decoder from 2 to 1.

\paragraph{SPACE.}

We adapt the official PyTorch implementation of SPACE to integrate it in our library. While in \citet{lin2020space} the authors train SPACE for 160k steps, here we train it for 200k. Since SPACE was not tested on any of the five datasets considered here (see \cref{tab:app_models_x_datasets}), we perform a hyperparameter sweep for all datasets. For each of the five datasets, we run a random search over hyperparameters by training 100 models for 100k steps. \cref{tab:app_models_space_parameters} shows the random search definition, the hyperparameter values used for each dataset, and how they differ from those used in the original publication for the \emph{3D-Rooms} dataset (although we omit some hyperparameters that we leave unchanged).

\begin{table}
\centering
\caption{Hyperparameters for SPACE experiments. Here we show: the hyperparameters recommended by \citet{lin2020space} for the 3D-Rooms dataset on the official code repository; the hyperparameter space considered for our random search; the chosen default values across datasets; the dataset-specific values for CLEVR and Tetrominoes, which override the defaults. We omit some of the hyperparameters that we left unchanged from \citet{lin2020space}.}
\label{tab:app_models_space_parameters}
\begin{tabular}{lccccc}
\toprule
\multirow{2}{*}{\textbf{Hyperparameter}} &
  \multirow{2}{*}{\begin{tabular}[c]{@{}c@{}}\textbf{Original}\\ \textbf{(3D-Rooms)}\end{tabular}} &
  \multirow{2}{*}{\textbf{Sweep values}} &
  \multirow{2}{*}{\textbf{Default value}} &
  \multicolumn{2}{c}{\textbf{Dataset-specific values}} \\ \cmidrule(lr){5-6} 
\multicolumn{1}{c}{}   &         &                              &         & \textbf{CLEVR} & \textbf{Tetrominoes} \\ \midrule
FG optimizer           & RMSprop & RMSprop                      & RMSprop & ---   & ---         \\
FG learning rate       & 1e-5    & \{3e-6, 1e-5, 3e-5, 1e-4\}   & 3e-5    & 1e-4  & 1e-4        \\
BG optimizer           & Adam    & Adam                         & Adam    & ---   & ---         \\
BG learning rate       & 1e-3    & 1e-3                         & 1e-3    & ---   & ---         \\
Batch size             & 12      & $\{16, 32\}$                 & 32      & ---   & ---         \\
$\sigma_\mathrm{bg}$   & 0.15    & $\{0.05, 0.15, 0.35\}$       & 0.15    & 0.05  & ---         \\
$\sigma_\mathrm{fg}$   & 0.15    & $\{0.02, 0.05, 0.15, 0.35\}$ & 0.15    & 0.05  & ---         \\
$G$ (FG grid size)     & 8       & $\{4, 8\}$                   & 8       & ---   & 4           \\
$K$ (BG n. of slots)   & 5       & $\{1, 5\}$                   & 5       & ---   & ---         \\
Boundary loss off step & 100k    & \{20k, 100k\}                & 20k     & ---   & 100k        \\
$\tau$ anneal end step & 20k     & \{20k, 50k\}                 & 50k     & 20k   & ---         \\
\begin{tabular}[c]{@{}l@{}}Mean of $p(\zb_\mathrm{pres})$ \\ \ \ (start/end values)\end{tabular} &
  (0.1, 0.01) &
  \{(0.1, 0.01), (0.5, 0.05)\} &
  (0.5, 0.05) &
  (0.1, 0.01) &
  (0.1, 0.01) \\
\begin{tabular}[c]{@{}l@{}}Mean of $p(\zb_\mathrm{scale})$ \\ \ \ (start/end values)\end{tabular} &
  $(-1, -2)$ &
  $\{(-1, -2), (0, -1)\}$ &
  $(0, -1)$ &
  --- &
  --- \\ \bottomrule
\end{tabular}
\end{table}

\paragraph{VAEs.}

The architecture details for the VAEs are presented in \cref{tab:bsaeline_VAE_encoder_structure,tab:baseline_VAE_decoder_MLP,tab:baseline_VAE_decoder_broadcast}. These are used for Shapestacks, Multi-dSprites, and Objects Room. For CLEVR, an additional ResidualBlock with 64 channels and a AvgPool2D layer is added at the end of the stack of ResidualBlocks, to downsample the image one more time. This is mirrored in the decoder, where a  ResidualBlock with 256 channels and a (bilinear) Interpolation layer is added at the beginning of the stack of ResidualBlocks. The same happens in the broadcast decoder case. For Tetrominoes, the number of layers is the same, but the last AvgPool2D layer is removed from the encoder and the first Interpolation layer is removed from the decoder, to have one less downsampling and upsampling, respectively. 
The latent space size is chosen to be 64 times the number of slots that would be used when training an object-centric model on the same dataset. Note that the default number of slots varies depending on the dataset, as shown in \cref{tab:params_datasets}.\footnote{Here we consider the default for MONet, Slot Attention, and GENESIS, and we disregard SPACE. Although SPACE has a much larger number of slots, this is not comparable with the other models because of the grid-based spatial attention mechanism.}

\begin{table}
\centering
\caption{Structure of the encoder for both the vanilla and broadcast VAE, excluding the final linear layer that parameterizes $\mu$ and $\log \sigma^2$ of the approximate posterior.}
\begin{tabular}{lcc}
\toprule
\textbf{Encoder}           &          &                             \\ \toprule
\emph{Type}              & \emph{Size/Ch.} & \emph{Notes}                     \\ \midrule
Input: $\xb$   & $3$        &                             \\
Conv $5 \times 5$ &          & Stride $2$, Padding $2$     \\
LeakyReLU        &          &                             \\
          &          &                             \\
Residual Block    & $64$       & $2$ Conv layers               \\
Residual Block    & $64$       & $2$ Conv layers               \\
Conv $1 \times 1$ & $128$      &                             \\
AvgPool2D       &          & Kernel size $2$, Stride $2$ \\
          &          &                             \\
Residual Block    & $128$      & $2$ Conv layers               \\
Residual Block    & $128$      & $2$ Conv layers               \\
AvgPool2D       &          & Kernel size $2$, Stride $2$ \\
          &          &                             \\
Residual Block    & $128$      & $2$ Conv layers               \\
Residual Block    & $128$      & $2$ Conv layers               \\
Conv $1 \times 1$ & $256$      &                             \\
          &          &                             \\
Residual Block    & $256$      & $2$ Conv layers               \\
Residual Block    & $256$      & $2$ Conv layers               \\
          &          &                             \\
Flatten           &          &                             \\
LeakyReLU        &          &                             \\
Linear            & $512$      &                             \\
LeakyReLU        &          &                             \\
LayerNorm        &          &                             \\ \bottomrule
\end{tabular}
\label{tab:bsaeline_VAE_encoder_structure}
\end{table}

\begin{table}
\centering
\caption{Structure of the decoder for the vanilla VAE.}
\begin{tabular}{lcc}
\toprule
\textbf{Vanilla Decoder}   &                        &                         \\ \toprule
\emph{Type}            & \emph{Size/Ch.}               & \emph{Notes}                 \\ \midrule
Input: $\zb$   & $64 \times$ num. slots &                         \\
LeakyReLU        &                        &                         \\
Linear            & $512$                  &                         \\
LeakyReLU        &                        &                         \\
Unflatten         &                        &                         \\
          &          &                             \\
Residual Block    & $256$                  & $2$ Conv layers         \\
Residual Block    & $256$                  & $2$ Conv layers         \\
Conv $1 \times 1$ & $128$                  &                         \\
Interpolation     &                        & Scale $2$               \\
          &          &                             \\
Residual Block    & $128$                  & $2$ Conv layers         \\
Residual Block    & $128$                  & $2$ Conv layers         \\
Interpolation     &                        & Scale $2$               \\
          &          &                             \\
Residual Block    & $128$                  & $2$ Conv layers         \\
Residual Block    & $128$                  & $2$ Conv layers         \\
Conv $1 \times 1$ & $64$                   &                         \\
Interpolation     &                        & Scale $2$               \\
          &          &                             \\
Residual Block    & $64$                   & $2$ Conv layers         \\
Residual Block    & $64$                   & $2$ Conv layers         \\
Interpolation     &                        & Scale $2$               \\
          &          &                             \\
LeakyReLU        &                        &                         \\
Conv $5 \times 5$ & Image channels         & Stride $1$, Padding $2$\\\bottomrule
\end{tabular}
\label{tab:baseline_VAE_decoder_MLP}
\end{table}

\begin{table}
\centering
\caption{Structure of the decoder for the broadcast VAE. One less Interpolation is required, because the final image size for this architecture is $64$ and the broadcasting is to a feature map of size~$8$. }
\begin{tabular}{lll}
\toprule
\textbf{Broadcast Decoder} &                               &                         \\ \toprule
\emph{Type}              & \emph{Size/Ch.}             & \emph{Notes}        \\ \midrule
Input: $\zb$            & $64 \times$ num. slots        &                         \\
Broadcast                  & $64 \times$ num. slots $ + 2$ & Broadcast dim. $8$      \\
          &          &                             \\
Residual Block             & $256$                         & $2$ Conv layers         \\
Residual Block             & $256$                         & $2$ Conv layers         \\
Conv $1 \times 1$          & $128$                         &                         \\
          &          &                             \\
Residual Block             & $128$                         & $2$ Conv layers         \\
Residual Block             & $128$                         & $2$ Conv layers         \\
Interpolation              &                               & Scale $2$               \\
          &          &                             \\
Residual Block             & $128$                         & $2$ Conv layers         \\
Residual Block             & $128$                         & $2$ Conv layers         \\
Conv $1 \times 1$          & $64$                          &                         \\
Interpolation              &                               & Scale $2$               \\
          &          &                             \\
Residual Block             & $64$                          & $2$ Conv layers         \\
Residual Block             & $64$                          & $2$ Conv layers         \\
          &          &                             \\
LeakyReLU                 &                               &                         \\
Conv $5 \times 5$          & Image channels                & Stride $1$, Padding $2$ \\ \bottomrule
\end{tabular}
\label{tab:baseline_VAE_decoder_broadcast}
\end{table}


\clearpage

\section{Datasets}
\label{app:datasets}

\begin{figure}
    \def\datasetleftspacing{\qquad\qquad\qquad\qquad}
    \def\datasetvspace{\vspace{11pt}}
    
    \newlength{\datasetheight}
    \setlength{\datasetheight}{3.75cm}
    
    \datasetleftspacing \includegraphics[height=\datasetheight]{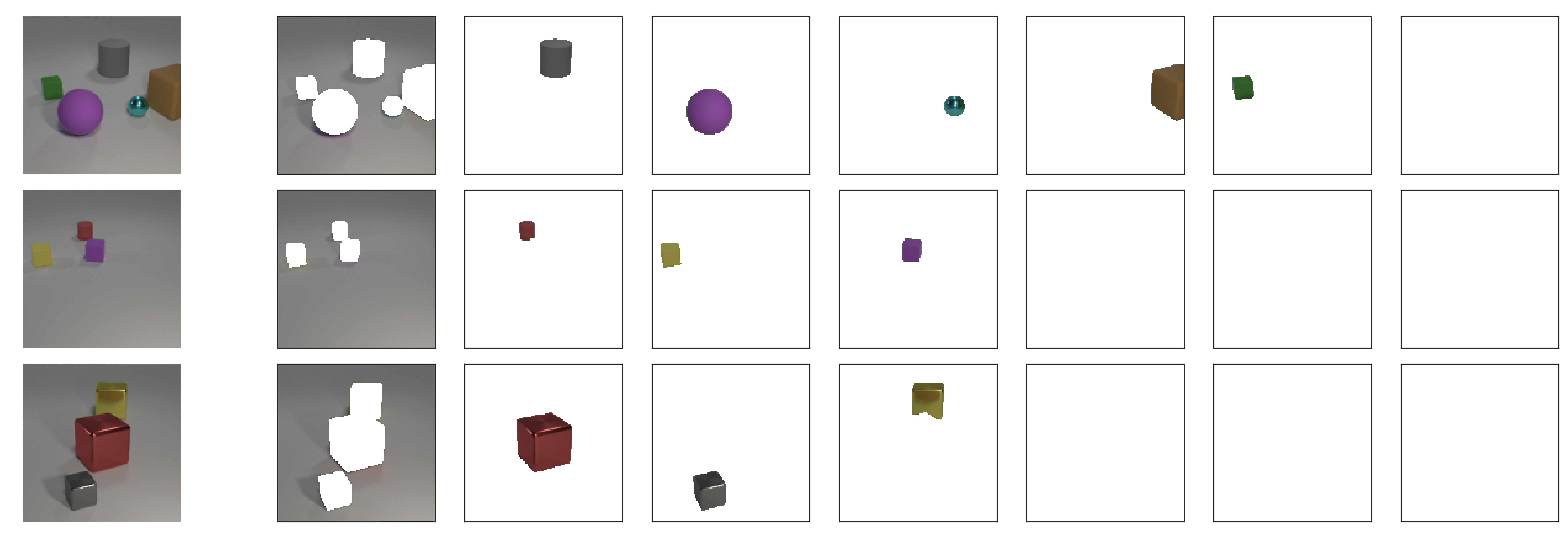}
    
    \datasetvspace
    
    \datasetleftspacing
    \includegraphics[height=\datasetheight]{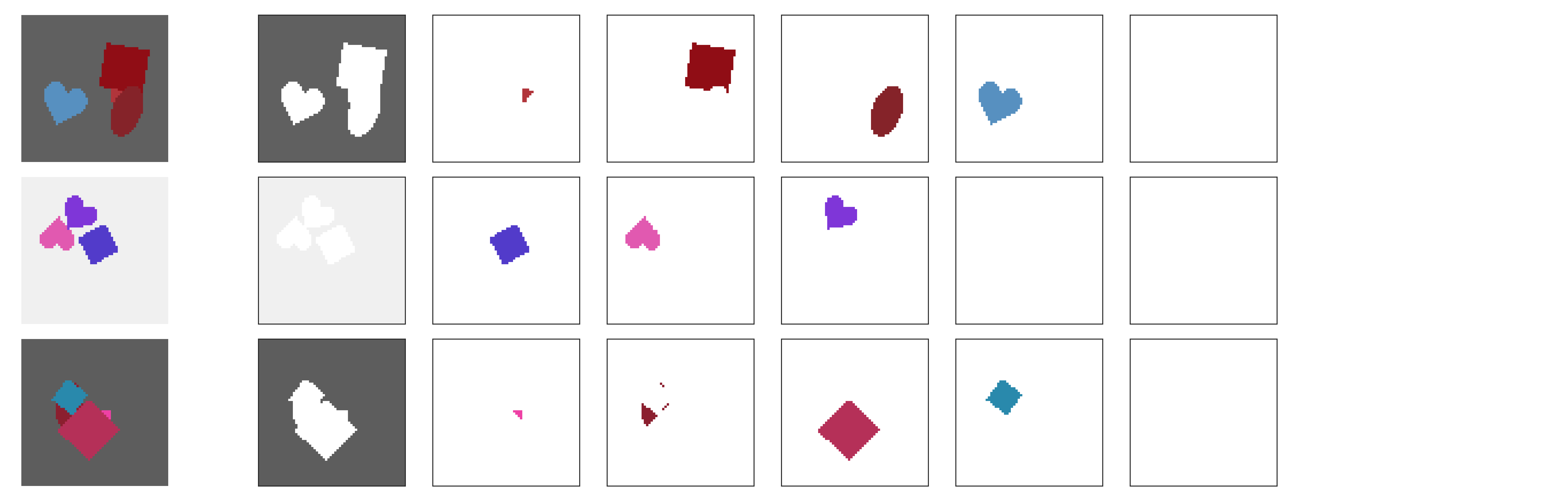}
    
    \datasetvspace
    
    \datasetleftspacing \includegraphics[height=\datasetheight]{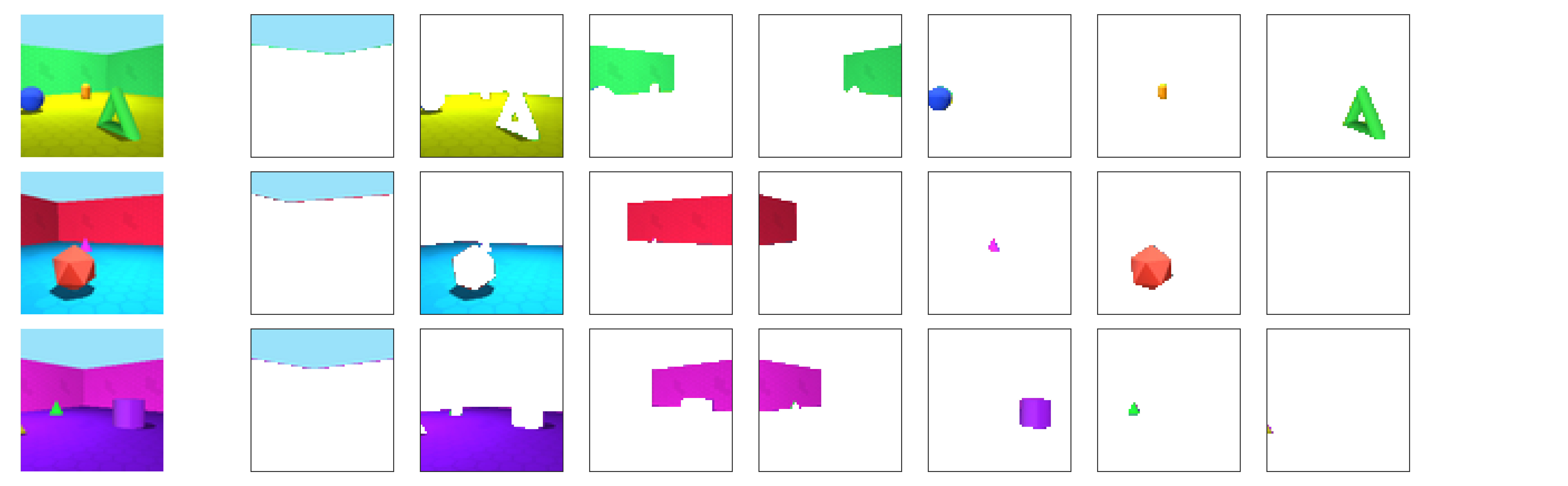}
    
    \datasetvspace
    
    \datasetleftspacing \includegraphics[height=\datasetheight]{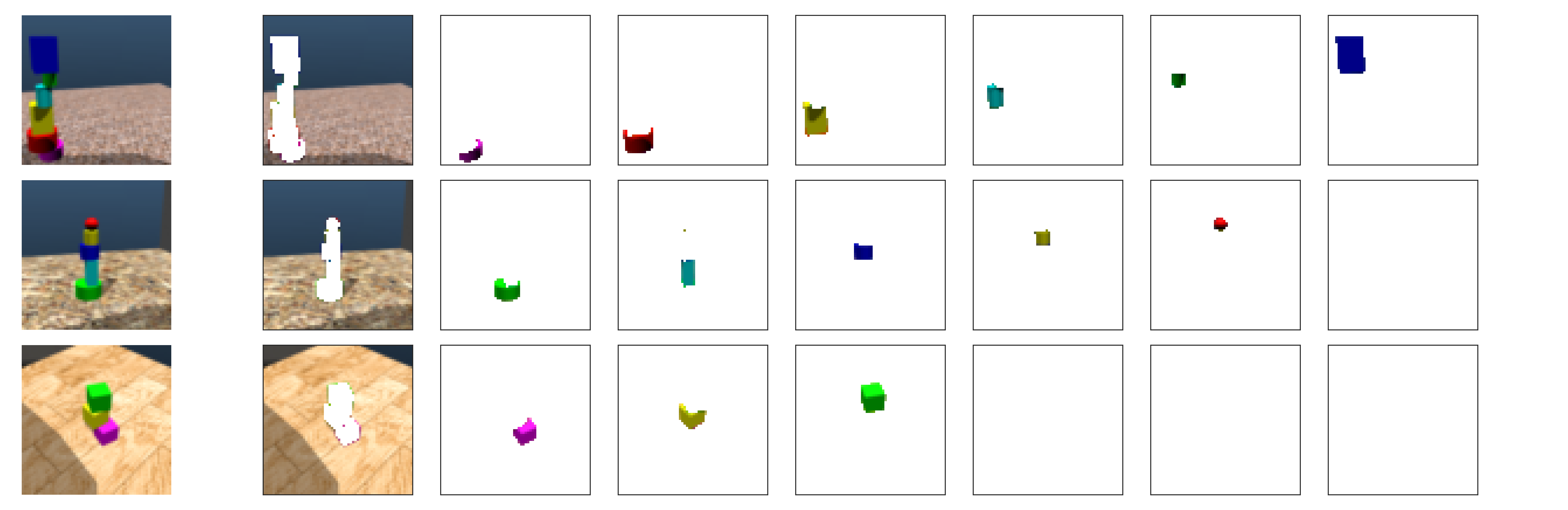}
    
    \datasetvspace
    
    \datasetleftspacing \includegraphics[height=\datasetheight]{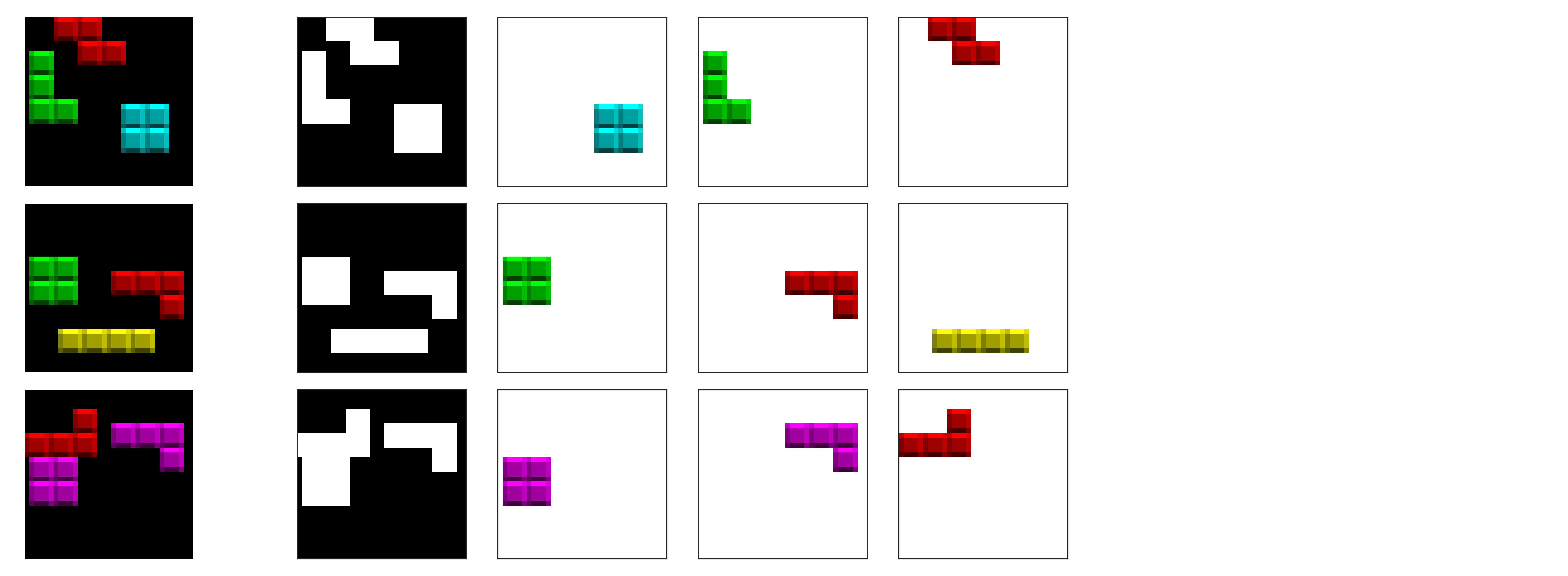}
    
    \datasetvspace

	\caption{Examples of images from the datasets considered in this work. The leftmost column represents the original image, the other columns show all the objects in the scene according to the ground-truth segmentation masks. Top to bottom: CLEVR6, Multi-dSprites, Objects Room, Shapestacks, Tetrominoes.}
	\label{fig:dataset_overview}
\end{figure}

We collected 5 existing multi-object datasets and converted them into a common format. 
\emph{Multi-dSprites}, \emph{Objects Room} and \emph{Tetrominoes} are from DeepMind's Multi-Object Datasets collection, under the Apache 2.0 license~\cite{multiobjectdatasets19}. \emph{CLEVR} was originally proposed by \citet{johnson2017clevr}, with segmentation masks introduced by \citet{multiobjectdatasets19}. \emph{Shapestacks} was proposed by \citet{groth2018shapestacks} under the GPL 3.0 license.
Details on these datasets are provided in the following subsections. See \cref{fig:dataset_overview} for sample images and ground-truth segmentation masks for these datasets.
In \cref{tab:params_datasets}, we report dataset splits, number of foreground and background objects, and number of slots used when training object-centric models.\looseness=-1

\subsection{CLEVR}

This dataset consists of $128 \times 128$ images of 3D scenes with up to 10 objects, possibly occluding each other.
Objects can have different colors (8 in total), materials (rubber or metal), shapes (sphere, cylinder, cube), sizes (small or large), x and y positions, and rotations.
Objects can be occluded by others. On average, 6.2 objects are visible.
As in previous work~\cite{greff2019multi,locatello2020object}, we learn object-centric representations on the CLEVR6 variant, which contains at most 6 objects.
There are $\num{100000}$ samples in the full dataset, and $\num{53483}$ in the CLEVR6 variant (at most 6 objects).
The CLEVR dataset has been cropped and resized according to the procedure detailed originally by~\citet{burgess2019monet}.

Each object is annotated with the following properties:
\begin{itemize}
 \item \code{color} (categorical): 8 colors:
        \begin{itemize}
            \item Red. RGB:\code{[173, 35, 35]}
            \item Cyan. RGB:\code{[41, 208, 208]}
            \item Green. RGB:\code{[29, 105, 20]}
            \item Blue. RGB:\code{[42, 75, 215]}
            \item Brown. RGB:\code{[129, 74, 25]}
            \item Gray. RGB:\code{[87, 87, 87]}
            \item Purple. RGB:\code{[129, 38, 192]}
            \item Yellow. RGB:\code{[255, 238, 51]}
        \end{itemize}
 \item \code{material} (categorical): The material of the object: rubber or metal.
 \item \code{shape} (categorical): The shape of the object: sphere, cylinder or cube.
 \item \code{size} (categorical): The size of the object: small or large.
 \item \code{x} (numerical): The x coordinate in 3D space.
 \item \code{y} (numerical): The y coordinate in 3D space.
\end{itemize} 

\subsection{Multi-dSprites}
This dataset is based on the \emph{dSprites} dataset~\cite{dsprites17}. Following previous work~\cite{greff2019multi,locatello2020object}, we use the Multi-dSprites variant with colored sprites on a grayscale background. Each scene has 2--5 objects with random shapes (ellipse, square, heart), sizes (6 discrete values in $[0.5, 1]$), x and y position, orientation, and color (randomly sampled in HSV space). Objects can occlude each other. The intensity of the uniform grayscale background is randomly sampled in each image. Images have size $64 \times 64$.

Each object is annotated with the following properties:
\begin{itemize}
    \item \code{color} (numerical): 3-dimensional RGB color vector.
    \item \code{scale} (numerical): Scaling of the object, 6 uniformly spaced values between 0.5 and 1.
    \item \code{shape} (categorical): The shape type of the object (ellipse, heart, square).
    \item \code{x} (numerical): Horizontal position between 0 and 1.
    \item \code{y} (numerical): Vertical position between 0 and 1.
\end{itemize}

\begin{table}
\centering
\caption{Dataset splits, number of foreground and background objects, and number of slots used when training object-centric models.}
\label{tab:params_datasets}
\begin{tabular}{lcccccc}
\toprule
\textbf{Dataset Name} & \textbf{Train} & \textbf{Validation} & \textbf{Test} & \textbf{Background} &  \textbf{Foreground} & \textbf{Slots}  \\
                      &  \textbf{Size} & \textbf{Size}       & \textbf{Size} & \textbf{Objects}       & \textbf{Objects}        &    \\ 
\midrule
{CLEVR6}         & $\num{49483}$ & $\num{2000}$ & $\num{2000}$ & 1 & 3--6 & 7$^*$ \\
{Multi-dSprites} & $\num{90000}$ & $\num{5000}$ & $\num{5000}$ & 1 & 2--5  & 6$^*$ \\
{Objects Room}   & $\num{90000}$ & $\num{5000}$ & $\num{5000}$ & 4 & 1--3 & 7$^*$\\ 
{Shapestacks}    & $\num{90000}$ & $\num{5000}$ & $\num{5000}$ & 1 & 2--6 & 7$^*$\\
{Tetrominoes}    & $\num{90000}$ & $\num{5000}$ & $\num{5000}$ & 1 & 3 & 4$^\dagger$\\
\midrule
\multicolumn{7}{l}{\small
$^*$In SPACE we use 69 slots: 5 background slots, and a grid of $8\times 8$ foreground slots.
}\\
\multicolumn{7}{l}{\small
$^\dagger$In SPACE we use 21 slots: 5 background slots, and a grid of $4\times 4$ foreground slots.
}\\
\bottomrule
\end{tabular}
\end{table}

\subsection{Objects Room}

This dataset was originally introduced by \citet{eslami2018neural} and consists of $64\times 64$ images of 3D scenes with up to three objects.
Since this dataset includes masks but no labels for the object properties, we can use it only to evaluate segmentation performance.

\subsection{Shapestacks}

This dataset consists of $64\times 64$ images of 3D scenes where objects are stacked to form a tower. Each scene is available under different camera views. Object properties are shape (cube, cylinder, sphere), color (6 possible values), size (numerical) and ordinal position in the stack.

%
%
Each object is annotated with the following properties:
\begin{itemize}
 \item \code{shape} (categorical): shape of the object: cylinder, sphere or cuboid.
 \item \code{color} (categorical): 6 colors:
            \begin{itemize}
                 \item Blue. RGB:\code{[0, 0, 255]}
                 \item Green. RGB:\code{[0, 255, 0]}
                 \item Cyan. RGB:\code{[0, 255, 255]}
                 \item Red. RGB:\code{[255, 0, 0]}
                 \item Purple. RGB:\code{[255, 0, 255]}
                 \item Yellow. RGB:\code{[255, 255, 0]}
            \end{itemize}
\end{itemize} 

\subsection{Tetrominoes}
This dataset consists of $32 \times 32$ images (cropped from the original $35 \times 35$ for simplicity) of 3D-textured tetris pieces placed on a black background. There are always 3 objects in a scene, and no occlusions. Objects have different shapes (19 in total), colors (6 fully saturated colors), x and y position.

Each object is annotated with the following properties:
\begin{itemize}
 \item \code{shape} (categorical): 19 shapes:
        \begin{itemize}
         \item Horizontal I piece.
         \item Vertical I piece.
         \item L piece pointing downward.
         \item J piece pointing upward.
         \item L piece pointing upward.
         \item J piece pointing downward.
         \item L piece pointing left.
         \item J piece pointing left.
         \item J piece pointing right.
         \item L piece pointing right.
         \item Horizontal Z piece.
         \item Horizontal S piece.
         \item Vertical Z piece.
         \item Vertical S piece.
         \item T piece pointing upward.
         \item T piece pointing downward.
         \item T piece pointing left.
         \item T piece pointing right.
         \item O piece.
        \end{itemize}
 
 \item \code{color} (categorical): 6 colors:
        \begin{itemize}
         \item Blue. RGB:\code{[0, 0, 255]}
         \item Green. RGB:\code{[0, 255, 0]}
         \item Cyan. RGB:\code{[0, 255, 255]}
         \item Red. RGB:\code{[255, 0, 0]}
         \item Purple. RGB:\code{[255, 0, 255]}
         \item Yellow. RGB:\code{[255, 255, 0]}
        \end{itemize}
 \item \code{x} (numerical): Horizontal position.
 \item \code{y} (numerical): Vertical position.
\end{itemize}

\clearpage

\section{Evaluations}
\label{app:evaluation}

In this section, we discuss in more detail the chosen reconstruction and segmentation metrics (\cref{app:evaluation_metrics}), provide implementation details on the downstream property prediction task (\cref{app:evaluation_downstream}), and more closely examine the distribution shifts considered in this study (\cref{app:evaluation_shifts}).

\subsection{Reconstruction and segmentation metrics}\label{app:evaluation_metrics}

\paragraph{Mean reconstruction error.}
Since all models in this study are autoencoders, we can use the reconstruction error to 
This is potentially an informative metric as it should roughly indicate the amount and accuracy of information captured by the models and present in the representations.
All models include some form of reconstruction term in their losses, but they may take different forms. We then choose to evaluate the reconstruction error with the mean squared error (MSE), defined for an image $\xb$ and its reconstruction $\hat{\xb}$ as follows:
\begin{equation}
    \mathrm{MSE} \left( \xb, \hat{\xb} \right) = \| \xb - \hat{\xb} \|_2^2 = \frac{1}{D} \sum_{i=1}^D (x_i - \hat{x}_i)^2
\end{equation}
where for simplicity we assume a vector representation of $\xb$ and $\hat{\xb}$, both with dimension $D$ equal to the number of pixels times the number of color channels.

\paragraph{Adjusted Rand Index (ARI).}

The Adjusted Rand Index (ARI) \cite{hubertComparingPartitions1985} measures the similarity between two partitions of a set (or clusterings). Interpreting segmentation as clustering of pixels, the ARI can be used to measure the degree of similarity between two sets of segmentation masks. Segmentation accuracy is then assessed by comparing ground-truth and predicted masks.
The expected value of the ARI on random clustering is 0, and the maximum value is 1 (identical clusterings up to label permutation).
As in prior work \cite{burgess2019monet,engelcke2020genesis,locatello2020object}, we only consider the ground-truth masks of foreground objects when computing the ARI.
Below, we define the Rand Index and the Adjusted Rand Index in more detail.

The Rand Index is a symmetric measure of the similarity between two partitions of a set~\citep{rand1971objective,hubertComparingPartitions1985,wagner2007comparing}. It is inspired by traditional classification metrics that compare the number of correctly and incorrectly classified elements. 
The Rand Index is defined as follows:
Let $S$ be a set of $n$ elements, and let $A = \{A_1, \dots, A_{n_A}\}$ and $B = \{B_1, \dots, B_{n_B}\}$ be partitions of $S$. Furthermore, let us introduce the following quantities:%
\begin{itemize}[topsep=0pt,itemsep=0pt,partopsep=0pt,parsep=\parskip]
	\item $m_{11}$: number of pairs of elements that are in the same subset in both $A$ and $B$,
	\item $m_{00}$: number of pairs of elements that are in different subsets in both $A$ and $B$,
	\item $m_{10}$: number of pairs of elements that are in the same subset in $A$ and in different subsets in $B$,
	\item $m_{01}$: number of pairs of elements that are in different subsets in $A$ and in the same subset in $B$.
\end{itemize}
The Rand Index is then given by:
\begin{equation}
	\mathrm{RI} (A,B) = \cfrac{m_{11}+m_{00}}{m_{11}+m_{00}+m_{10}+m_{01}} = \cfrac{2(m_{11}+m_{00})}{n(n-1)}
\end{equation}
and quantifies the number of elements that have been correctly classified over the total number of elements. 

The Rand Index ranges from 0 (no pair classified in the same way under $A$ and $B$) to 1 ($A$ and $B$ are identical up to a permutation). However, the result is strongly dependent on the number of clusters and on the number of elements in each cluster. If we fix $n_A$, $n_B$, and the proportion of elements in each subset of the two partitions, then the Rand Index will increase as $n$ increases, and even converge to 1 in some cases \cite{fowlkes1983method}. The expected value of a random clustering also depends on the number of clusters and on the number of elements $n$.

The Adjusted Rand Index (ARI) \cite{hubertComparingPartitions1985} addresses this issue by normalizing the Rand Index such that, with a random clustering, the metric will be 0 in expectation.
Given the same conditions as above, let $n_{i,j} = |A_i \cap B_j|$, $a_i = |A_i|$, and $b_i = |B_i|$, with $i=1,\ldots,n_A$ and $i=1,\ldots,n_B$. The ARI is then defined as:
\begin{equation}
	\mathrm{ARI}(A,B)  = \cfrac{\sum_{i,j} \binom{n_{i,j}}{2} - \cfrac{\sum_{i} \binom{a_i}{2}\sum_{j} \binom{b_j}{2}}{\binom{n}{2}}}{\frac{1}{2}\left[\sum_i\binom{a_i}{2}+\sum_j\binom{b_j}{2}\right]-\cfrac{\sum_i\binom{a_i}{2}\sum_j\binom{b_j}{2}}{\binom{n}{2}}}
\end{equation}
which is 0 in expectation for random clusterings, and 1 for perfectly matching partitions (up to a permutation). Note that the ARI can be negative.

\paragraph{Segmentation covering metrics.}
Segmentation Covering (SC) \cite{arbelaez2010contour} uses the intersection over union (IOU) between pairs of segmentation masks from the sets $A$ and $B$. How the segmentation masks are matched depends on whether we are considering the covering of $B$ by $A$ (denoted by $A \rightarrow B$) or vice versa ($B \rightarrow A$).
We use the slightly modified definition by \citet{engelcke2020genesis}:
\begin{equation}
    \mathrm{SC}(A\rightarrow B) = \frac{1}{\sum_{R_B \in B} \left|R_B\right|} \sum_{R_B \in B} \left|R_B\right| \max_{R_A \in A} \operatorname{\textsc{iou}}(R_A, R_B)\ ,
    \label{eq:sc}
\end{equation}
where $|R|$ denotes the number of pixels belonging to mask $R$, and the intersection over union is defined as:
\begin{equation}
    \operatorname{\textsc{iou}}(R_A, R_B) = \frac{\left|R_A \cap R_B\right|}{\left|R_A \cup R_B\right|}\ .
    \label{eq:iou}
\end{equation}

While standard (weighted) segmentation covering weights the IOU by the size of the ground truth mask, mean (or unweighted) segmentation covering (mSC) \cite{engelcke2020genesis} gives the same importance to masks of different size:
\begin{equation}
    \mathrm{mSC}(A\rightarrow B) = \frac{1}{|B|} \sum_{R_B \in B}  \max_{R_A \in A} \operatorname{\textsc{iou}}(R_A, R_B)\ ,
    \label{eq:msc}
\end{equation}
where $|B|$ denotes the number of non-empty masks in $B$.
Since a high SC score can still be attained when small objects are not segmented correctly, mSC is considered to be a more meaningful and robust metric across different datasets~\cite{engelcke2020genesis}.

Note that neither SC nor mSC are symmetric: Following \citet{engelcke2020genesis}, we consider $A$ to be the predicted segmentation masks and $B$ the ground-truth masks of the foreground objects.
As observed by \citet{engelcke2020genesis}, both SC and mSC penalize over-segmentation (segmenting one object into separate slots), unlike the ARI.
Both SC and mSC take values in $[0,1]$.

\subsection{Downstream property prediction}\label{app:evaluation_downstream}

Here we start by briefly summarizing the downstream property prediction task presented in the main text, and then provide additional details on the models and evaluation protocol.

\paragraph{Overview of the property prediction task.}
As outlined in \cref{sec:experimental_setup}, 
we evaluate scene representations by training downstream models to predict ground-truth object properties from the representations. 
Exploiting the fact that object slots share a common representational format, a single downstream model $f$ can be used to predict the properties of each object independently: for each slot representation $\zb_k$ we predict a vector of object properties $\ybpred_k = f(\zb_k)$. This vector represents predictions for \emph{all} properties of an object.
We then match each slot's prediction to a corresponding ground-truth object using \emph{mask matching} or \emph{loss matching} (see main text).
In non-slotted models such as the VAE baselines considered in this study, we do not have access to separate object representations $\{\zb_k\}_{k=1}^K$. Therefore, the downstream model $f$ in this case takes as input the overall distributed representation $\zb$, which is a flat vector, and outputs a prediction of \emph{all objects at once}: $\ybpred = f(\zb)$. This is then split into $K$ vectors, which are matched to ground-truth objects with either \emph{loss matching} or \emph{deterministic matching} (see main text).

\paragraph{Implementation details.}
We use 4 different downstream models: a linear model, and MLPs with up to 3 hidden layers of size 256 each.
Let $P$ be the size of the ground-truth property vector, which includes all numerical and categorical\footnote{Here we use the one-hot representation of categorical properties.} properties according to an order specified by the dataset.
We denote by $K$ be the number of slots and $d$ the dimensionality of a slot representation $\zb_k$ in object-centric models. Note that we must include in $\zb_k$ \emph{all} representations related to a slot, possibly including different latent variables that are explicitly responsible for modeling, e.g., the location, appearance, or presence of an object.
The downstream model $f$ has input size $d$ and output size $P$, and is applied in parallel (with shared weights) to all slots.
In non-slotted models, we always define the dimensionality of the distributed representation $\zb$ in terms of $K$ for fair comparison with slot-based models, hence we can write the latent dimensionality of such models as $d \cdot K$. In this case, the input and output sizes of the downstream model ($d$ and $P$, respectively) are multiplied by $K$, and we apply this model only once, to the entire scene representation.
The linear downstream model is implemented as a linear layer. MLP models (with at least one hidden layer) have hidden size 256 and LeakyReLU nonlinearities, as shown in \cref{tab:app_downstream_models}.

\begin{table}
\centering
\caption{Architecture of the downstream MLP models for property prediction. The third and fourth items are repeated 0 or more times, depending on the required number of hidden layers.}
\label{tab:app_downstream_models}
\phantom{repeated 0 or----}
\begin{tabular}{lccl}
\cmidrule[\heavyrulewidth]{1-3}
\textbf{Layer type} & \textbf{Input size} & \textbf{Output size} &                                                                    \\ 
\cmidrule{1-3}
Linear              & $d$ or $d \cdot K$  & 256                  &                                                                    \\
LeakyReLU($0.01$)   & 256                 & 256                  &                                                                    \\
Linear              & 256                 & 256                  & \rdelim\}{2}{30mm}[\ \ \parbox{24.5mm}{repeated 0 or\\more times}] \\
LeakyReLU($0.01$)   & 256                 & 256                  &                                                                    \\
Linear              & 256                 & $P$ or $P \cdot K$   &                                                                    \\ 
\cmidrule[\heavyrulewidth]{1-3}
\end{tabular}
\end{table}

\paragraph{Data splits.}
Let $\mathcal{D}_s$ be a source dataset and $\mathcal{D}_t$ a target dataset. When doing in-distribution evaluation, we train and test the downstream model without distribution shifts, so we simply have $\mathcal{D}_s = \mathcal{D}_t$.
Given a representation function $r$, and a matching strategy to match the slots with ground-truth objects, we consider:
\begin{itemize}[topsep=0pt,itemsep=0pt,partopsep=0pt,parsep=\parskip]
    \item a train split of $\num{10000}$ images from $\mathcal{D}_s$\ ,
    \item a validation split of $\num{1000}$ images from $\mathcal{D}_s$\ ,
    \item a test split of $\num{2000}$ images from $\mathcal{D}_t$\ .
\end{itemize}
The test split only contains images that were not used when training the upstream unsupervised models.

\paragraph{Training.}
We then train the downstream model to predict $\ybpred$ from $\zb=r(\xb)$ using the Adam optimizer with an initial learning rate of 1e-3 and a batch size of 64, for a maximum of $\num{6000}$ steps. The learning rate is halved every $\num{2000}$ steps. We perform early stopping as follows: We use the validation set to compute the (in-distribution) validation loss every $\num{250}$ training steps---if the loss does not decrease by more than $0.01$ for 3 evaluations (750 steps), training is interrupted.
In this stage, the representation for each image is fixed, i.e. the representation function $r$ is never updated.
The loss is computed independently for each object property, and is a sum of MSE and cross-entropy terms, depending on whether an object property is numerical or categorical.

\paragraph{Downstream training and evaluation under distribution shifts.}
As mentioned earlier, when doing in-distribution evaluation we simply have $\mathcal{D}_s = \mathcal{D}_t$. In the general case, we may for example train on the original Multi-dSprites dataset, and test on the Multi-dSprites variant that has an unseen shape or an occlusion. In the special case in which we allow retraining of the downstream model (see \cref{sec:results/hyp2,sec:results/hyp3}), we still have $\mathcal{D}_s = \mathcal{D}_t$, but they are both OOD with respect to the original ``clean'' dataset used for training the unsupervised models.

Under distribution shifts, the representations $r(\xb)$ might be inaccurate, which might bias our downstream results. Although there is no perfect solution to this issue, we attempt to reduce as much as possible the potential effect of distribution shifts on the training and evaluation of downstream models.
When distribution shifts affect global scene properties, there is no alternative but to train and evaluate the models as usual. 
When distribution shifts affect single objects, however, we can assume that the representations of the ID objects are not as severely affected by the shift, and only use these for training downstream models. 

Here we consider the case where the test dataset $\mathcal{D}_t$ has an object-level distribution shift, and the training dataset $\mathcal{D}_s$ is either the original ``clean'' dataset or the same as $\mathcal{D}_t$. 
At \textbf{train time}, we ignore OOD objects (if any) both when matching slots with objects and when training downstream property prediction models. Note that, when the training dataset $\mathcal{D}_s$ is the original ``clean'' dataset, the downstream models are always trained as usual because there are no OOD objects.
At \textbf{test time}, there are a few cases depending on the matching strategy:
\begin{itemize}[topsep=0pt,itemsep=0pt,partopsep=0pt,parsep=\parskip]
    \item When using mask matching, we consider \emph{all} objects for matching, and evaluate the downstream models on all objects. We then report test results on ID and OOD objects separately.

    \item When using loss matching, we cannot match all ground-truth objects, since the OOD objects might have OOD categorical properties (in our setup, the downstream models cannot predict classes that were not seen during training). Therefore, we resort to a two-step matching approach: we first match slots to all objects using the prediction loss computed only on the properties that are ID for \emph{all} objects. We then keep only the matches for OOD objects, and repeat the usual loss matching with the remaining slots and objects, using all properties. The OOD objects are thus matched in a relatively fair way, while the matching of the ID objects can be refined at a later step using all available properties.

    \item When using deterministic matching, we cannot exactly follow the two-step matching strategy presented above. Instead, we modify the lexicographic order to give a higher weight to OOD features of OOD objects, so the corresponding objects are pushed down in the order while maintaining the order given by more significant (according to the order) properties. Note that the downstream model in this case might be at a disadvantage if it is trained on a dataset with object-level distribution shifts: the model is now trained to predict only ID objects, so at test time there will be one more target object on average.
\end{itemize}

\subsection{Distribution shifts for OOD evaluation}\label{app:evaluation_shifts}

\begin{figure}
    \centering
    \includegraphics[width=0.7\linewidth]{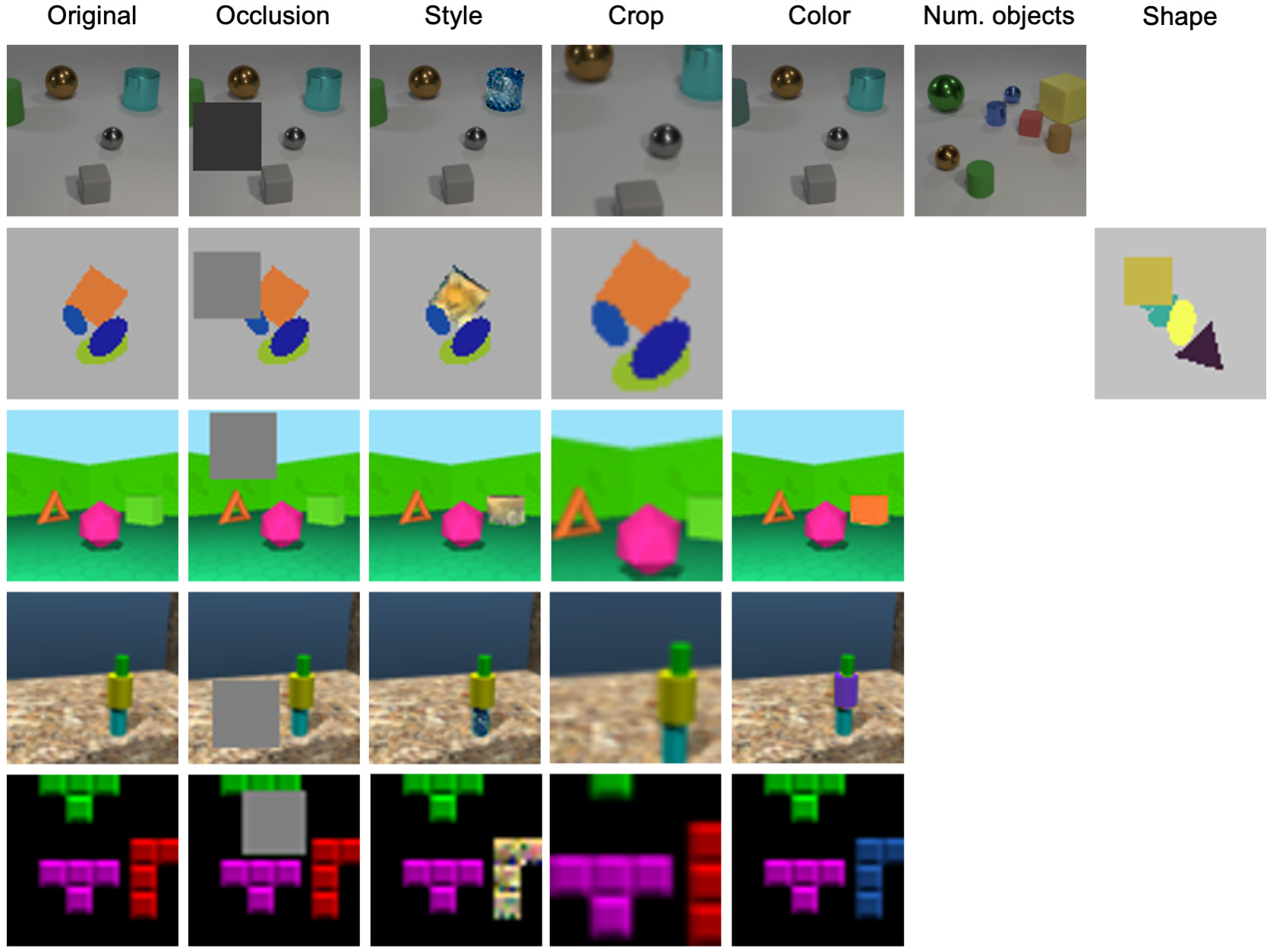}
    \caption{Distribution shifts applied to the different datasets to test generalization.}
    \label{fig:transform/transforms_datasets}
\end{figure}

Here we present more in detail the distribution shifts we apply to images in order to test OOD generalization in different scenarios. Examples are shown in \cref{fig:transform/transforms_datasets}.

\textbf{Occlusion.} A gray square is placed on top of the scene. The position is determined by picking 5 locations uniformly at random (such that the entire square is in the image) and selecting the one that occludes less (in terms of total area) of the foreground objects. The size of the occlusion is $(\lfloor 0.4 \cdot H\rfloor, \lfloor0.4 \cdot W\rfloor)$ with $H$ and $W$ the height and width of the image, respectively. Occluded objects have their mask updated to reflect the occlusion. The occlusion is categorized as background (or first background object in case there are multiple background objects such as in Objects Room). The RGB color of the square is $[0.2, 0.2, 0.2]$ for CLEVR and $[0.5,0.5,0.5]$ for all other datasets.

\textbf{Object color.} An object is selected uniformly at random and its color is changed by randomly adjusting its brightness, contrast, saturation, and hue, using torchvision's \code{ColorJitter} transform with arguments $[0.5, 0.5, 0.5, 0.5]$ for the four above-mentioned parameters. This transformation is not performed on Multi-dSprites, since the object colors in this dataset cover the entire RGB color space. The \texttt{color} and \texttt{material} properties (when relevant) are not used in downstream tasks.

\textbf{Crop.} The image and mask are cropped at the center and resized to match their original size. The crop size is $(\lfloor\frac{2}{3}H\rfloor, \lfloor\frac{2}{3}W\rfloor)$ with $H$ and $W$ the original height and width of the image, respectively. When resizing, we use bilinear interpolation for the image and nearest neighbor for the mask.

\textbf{Object style.} We implement style transfer based on \citet{gatys_neural_2016} and on the PyTorch tutorial by \citet{jacq_neural_nodate}. The first $100$k samples in all datasets are converted using as style image \emph{The Great Wave off Kanagawa} from Hokusai's series \emph{Thirty-six Views of Mount Fuji}. The style is applied only to one foreground object using the object masks. The \texttt{color} and \texttt{material} properties (when relevant) are not used in downstream tasks.

\textbf{Object shape.} For the Multi-dSprites dataset, a triangle is placed on the scene with properties sampled according to the same distributions defined by the Multi-dSprites dataset. This is performed only on the images where at most 4 objects are present, to mimic changing the shape of an existing object. The depth of the triangle in the object stack is selected uniformly at random as an integer in $[1,5]$. All objects from the selected depth and upwards are moved up by one level to place the new shape underneath them. The objects masks are adjusted accordingly for both the added shape and the objects below it.


\clearpage

\section{Additional results}
\label{app:results}

In this section, we report additional quantitative results and show qualitative performance
on all datasets for a selection of object-centric models and VAE baselines.

\subsection{Performance in the training distribution}

\begin{figure}[hb]
    \centering
    \includegraphics[width=0.94\textwidth]{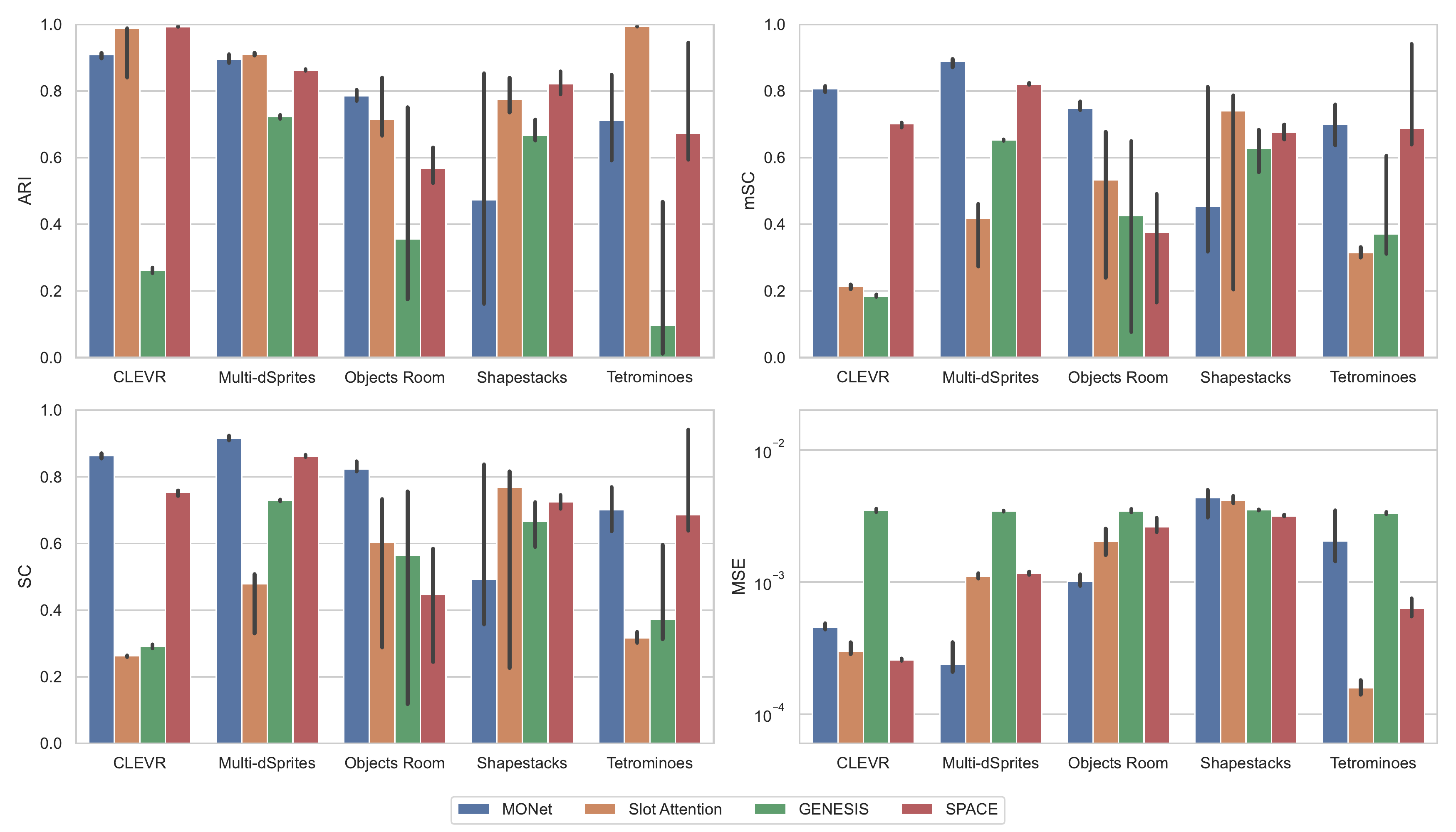}
    \caption{Overview of segmentation metrics (ARI \up, mSC \up, SC \up) and reconstruction MSE (\down) in distribution (test set of $\num{2000}$ images). The bars show medians and 95\% confidence intervals with 10 random seeds.}
    \label{fig:indistrib/metrics/box_plots}
\end{figure}
\begin{figure}
    \centering
    \includegraphics[width=0.85\textwidth]{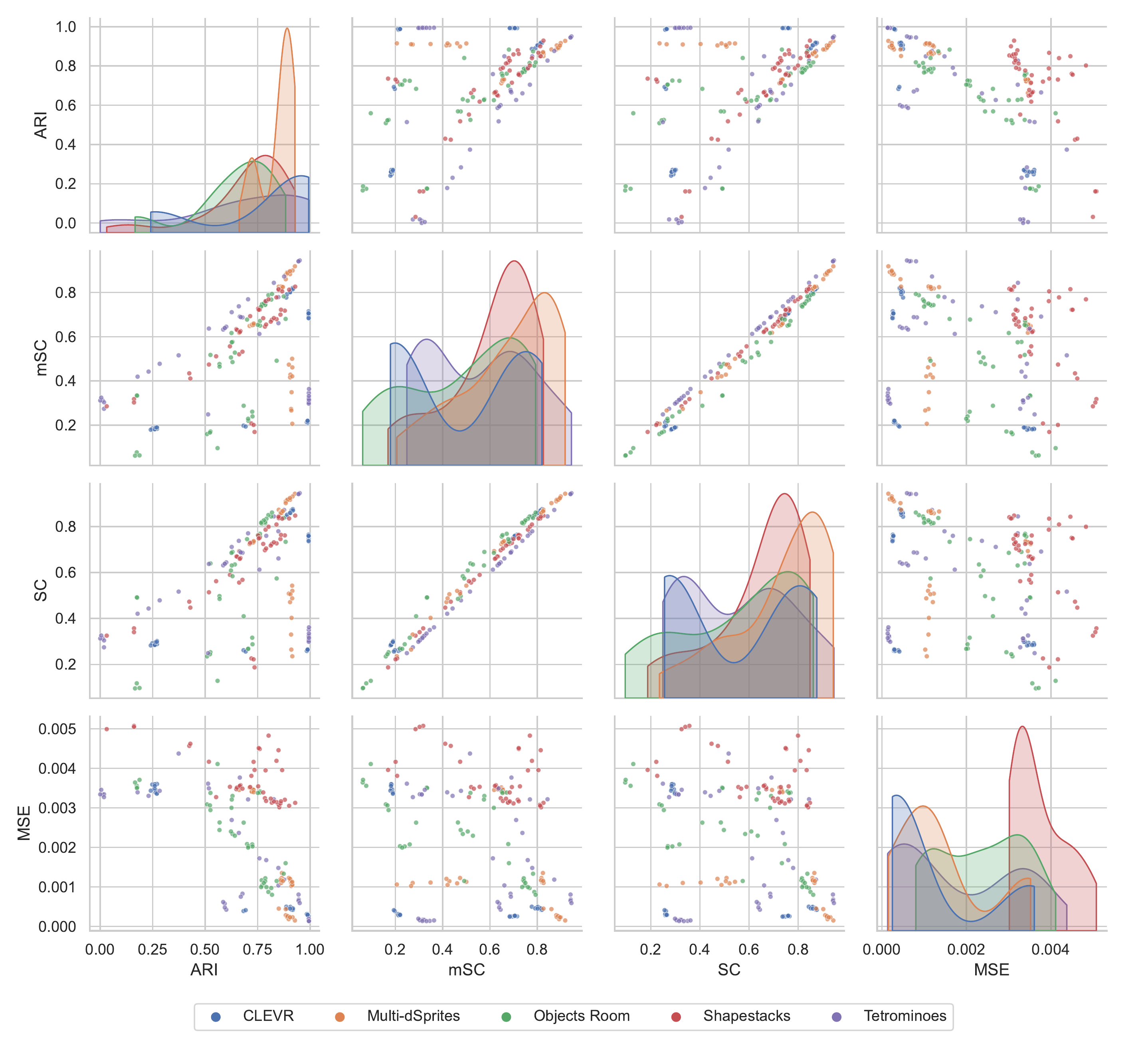}
    \caption{Scatter plots between metrics over all 200 object-centric models, color-coded by dataset. Diagonal plots: kernel density estimation (KDE) of the quantities on the x-axes.}
    \label{fig:indistrib/metrics_metrics/aggregated_datasets/scatter}
\end{figure}

\cref{fig:indistrib/metrics/box_plots} shows the distributions of the reconstruction MSE and all the segmentation metrics, broken down by dataset and model.
The relationship between these metrics is also shown in scatter plots in \cref{fig:indistrib/metrics_metrics/aggregated_datasets/scatter}.
As discussed in \cref{sec:results/metrics}, we observe that segmentation covering metrics are correlated with the ARI only in some cases, and the models are ranked very differently depending on the chosen segmentation metric. In particular, we observe here that Slot Attention achieves a high ARI score and significantly lower (m)SC scores on CLEVR, Multi-dSprites, and Tetrominoes. This is because Slot Attention on these datasets tends to model the background across many slots (see \cref{app:results_qualitative}), which is penalized by the denominator of the IOU in the (m)SC scores (see \cref{eq:sc,eq:iou,eq:msc} in \cref{app:evaluation_metrics}). This behavior should not have a major effect on downstream performance, which is confirmed by the strong and consistent correlation between ARI and downstream performance (see also \cref{sec:results/hyp1} and \cref{fig:indistrib/downstream/corr_metrics_downstream_all}).

\begin{figure}
    \centering
    \includegraphics[width=\textwidth]{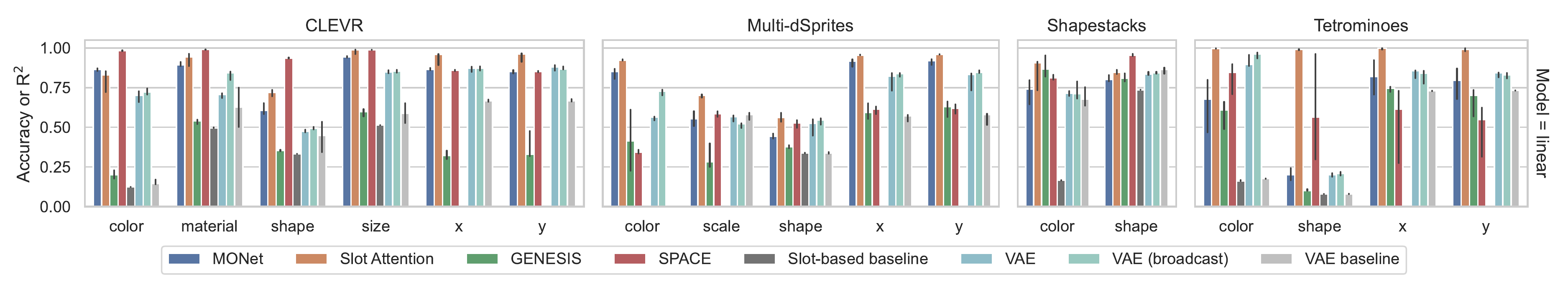}
    
    \vspace{10pt}
    \includegraphics[width=\textwidth]{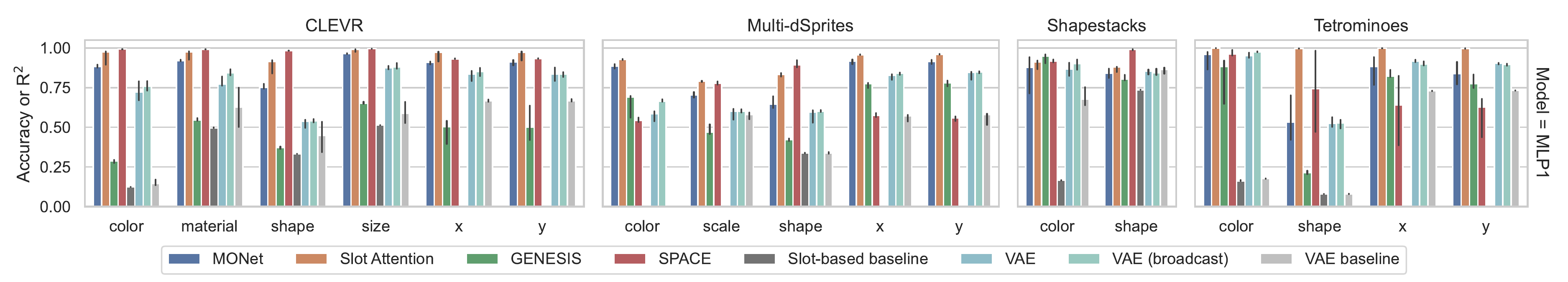}
    
    \vspace{10pt}
    \includegraphics[width=\textwidth]{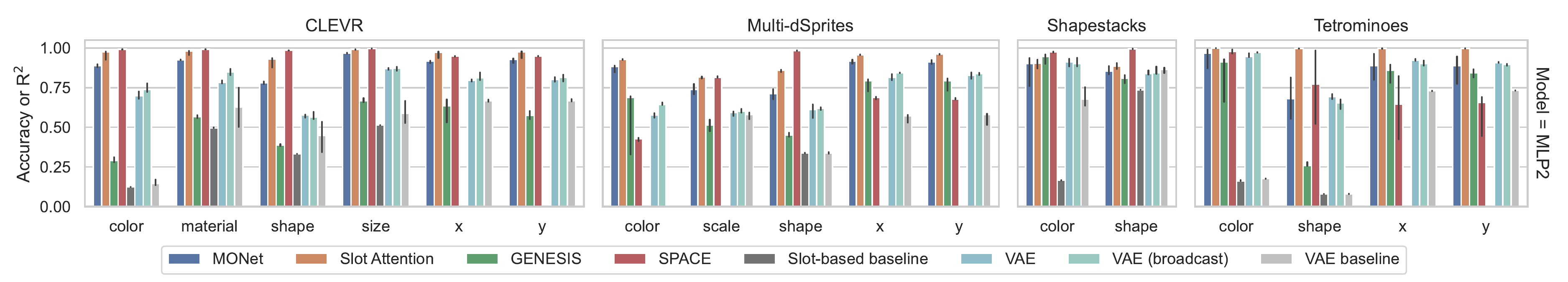}
    
    \vspace{10pt}
    \includegraphics[width=\textwidth]{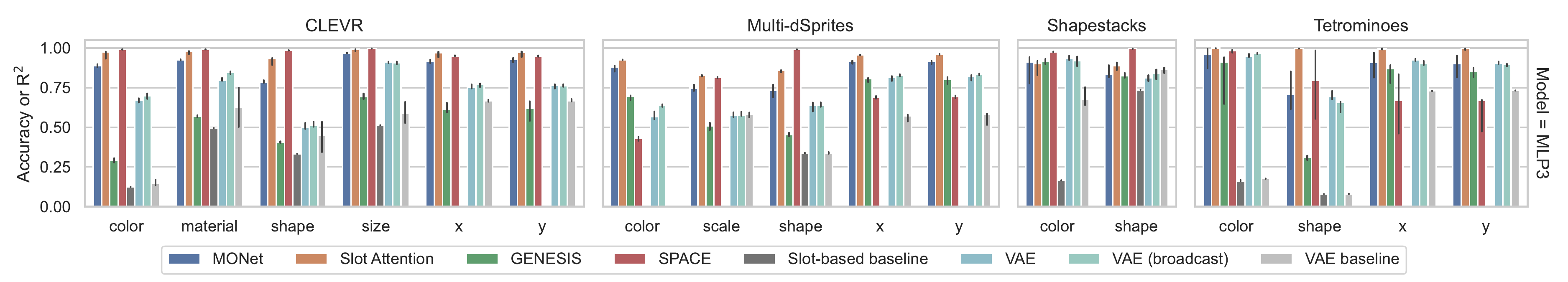}
    \caption{Overview of downstream performance in the training distribution (test set of $\num{2000}$ images) for object-centric models and VAEs, with respective baselines. The metrics on the y-axes are accuracy (\up) for categorical properties and \rsq (\up) for numerical features. Each row show results for a different downstream prediction model. From top to bottom: linear, MLP with 1, 2, and 3 hidden layers (see annotation on the right). We use loss matching (see \cref{sec:experimental_setup}) for all models. The bars show medians and 95\% confidence intervals with 10 random seeds.}
    \label{fig:indistrib/downstream/box_plots}
\end{figure}

\cref{fig:indistrib/downstream/box_plots} shows an overview of downstream factor prediction performance on all labeled datasets (one per column), using as downstream predictors a linear model or an MLP with up to 3 hidden layers (one model per row). The MLP1 results are also shown in \cref{sec:results/hyp1} (\cref{fig:indistrib/downstream/barplots_loss_linear_main}). We report results separately for each object-centric model and for each ground-truth object property. The metrics used here are accuracy for categorical attributes and \rsq for numerical attributes.
We generally observe consistent trends across downstream models. 

\begin{figure}
    \centering
    \includegraphics[height=3.8cm]{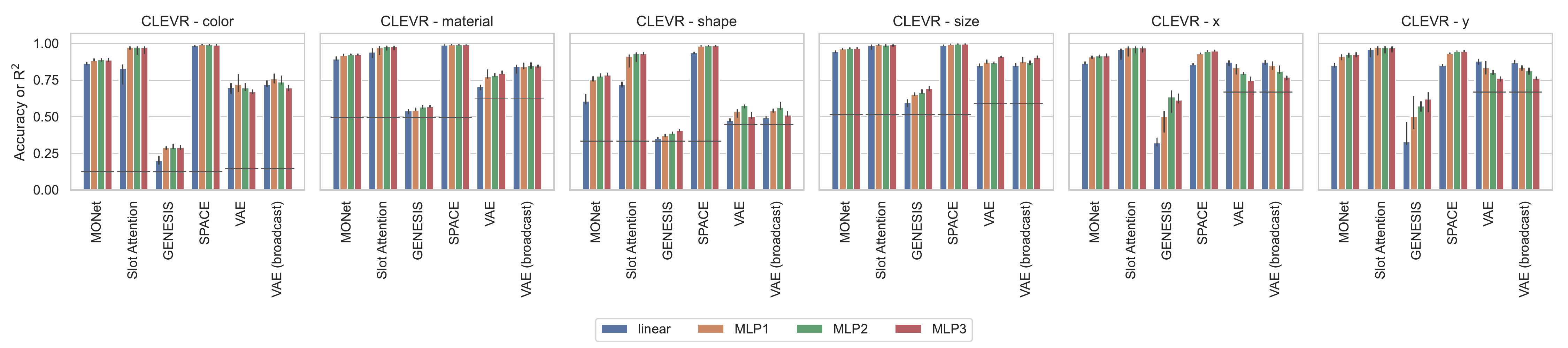}
    
    \vspace{30pt}
    \includegraphics[height=4cm]{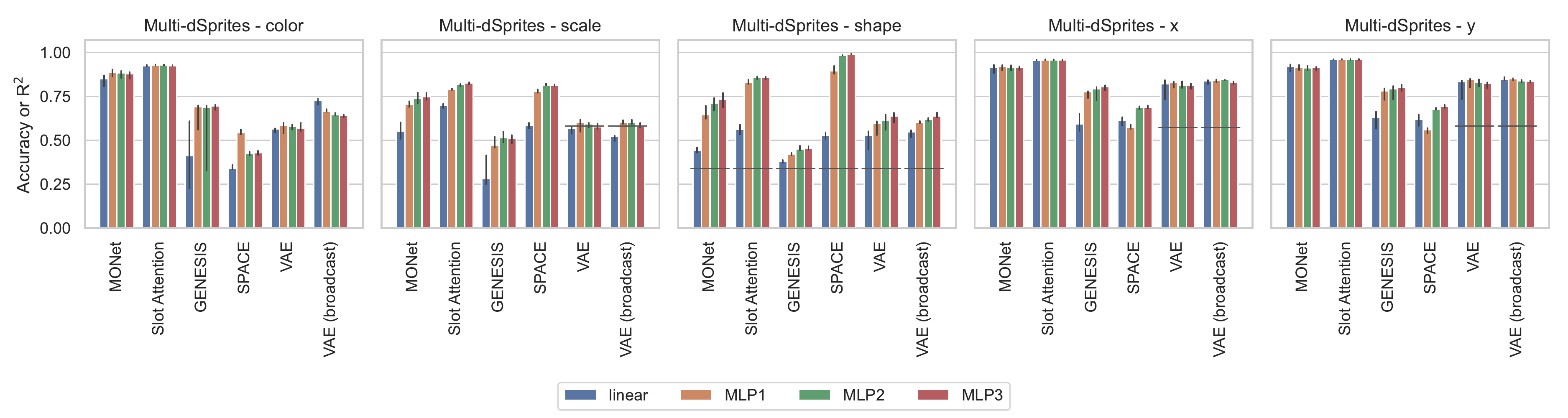}
    
    \vspace{30pt}
    \includegraphics[height=4cm]{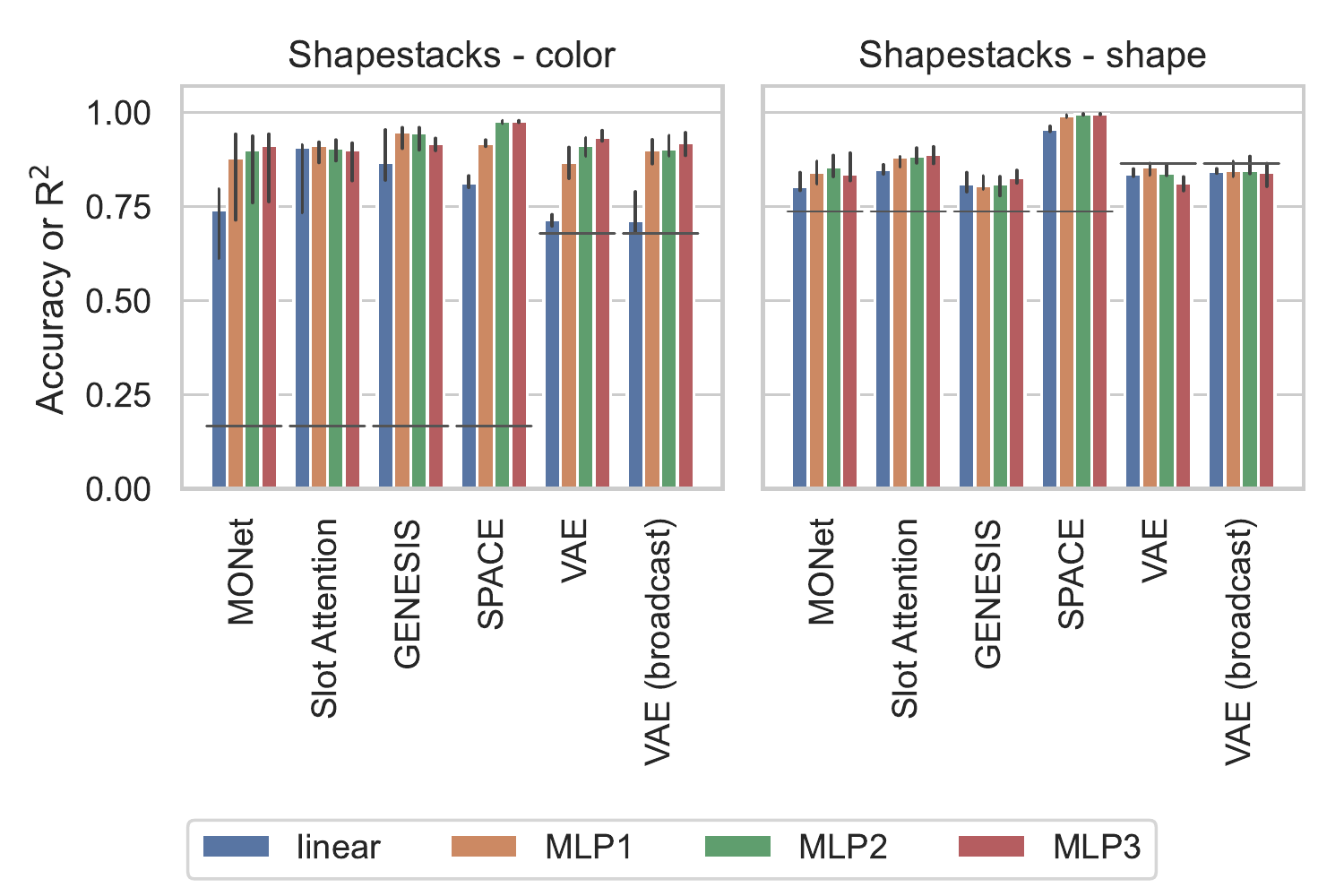}
    
    \vspace{30pt}
    \includegraphics[height=4cm]{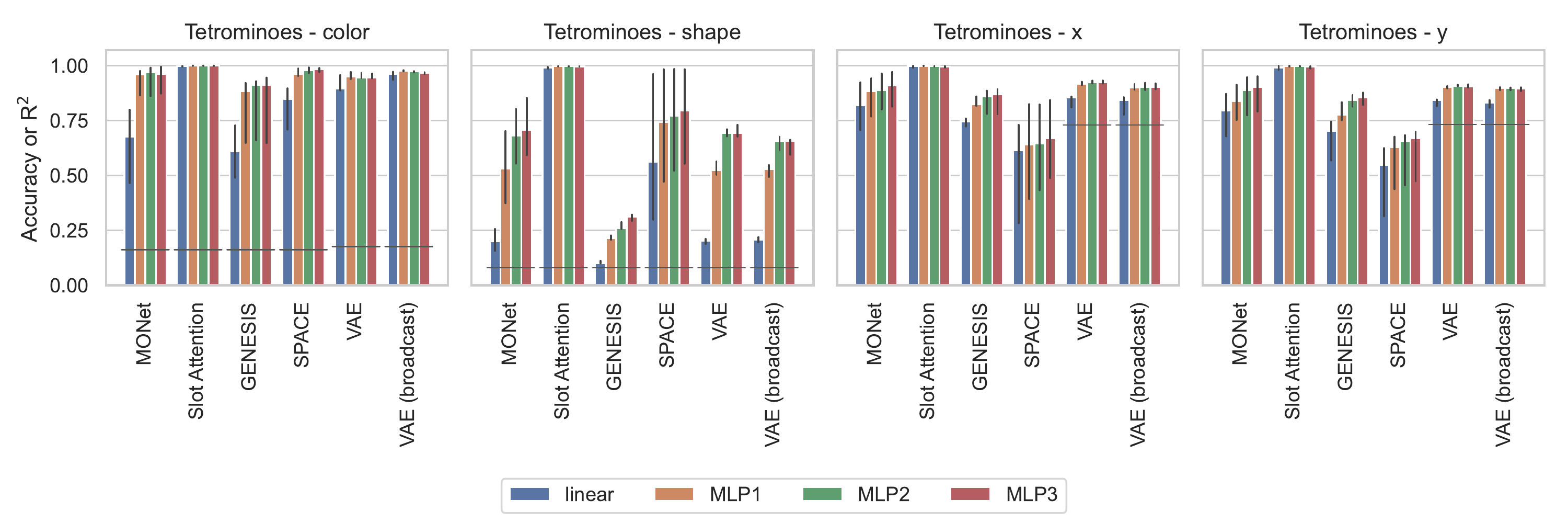}
    
    \caption{Comparing property prediction performance of different downstream models (linear, MLP with 1 to 3 hidden layers), using loss matching (see \cref{sec:experimental_setup}). Results on a test set of $\num{2000}$ images in the training distribution of the upstream unsupervised models. Each plot shows the test performance on one feature of a dataset. We show results for all object-centric models and VAEs, and indicate the baseline (see \cref{sec:experimental_setup}) with a horizontal line (not visible when the baseline is 0).
    The metrics on the y-axes are accuracy (\up) for categorical properties and \rsq (\up) for numerical features.
    The bars show medians and 95\% confidence intervals with 10 random seeds.}
    \label{fig:indistrib/downstream/barplots_compare_downstream_models_loss}
\end{figure}

\begin{figure}
    \centering
    \includegraphics[width=0.65\textwidth]{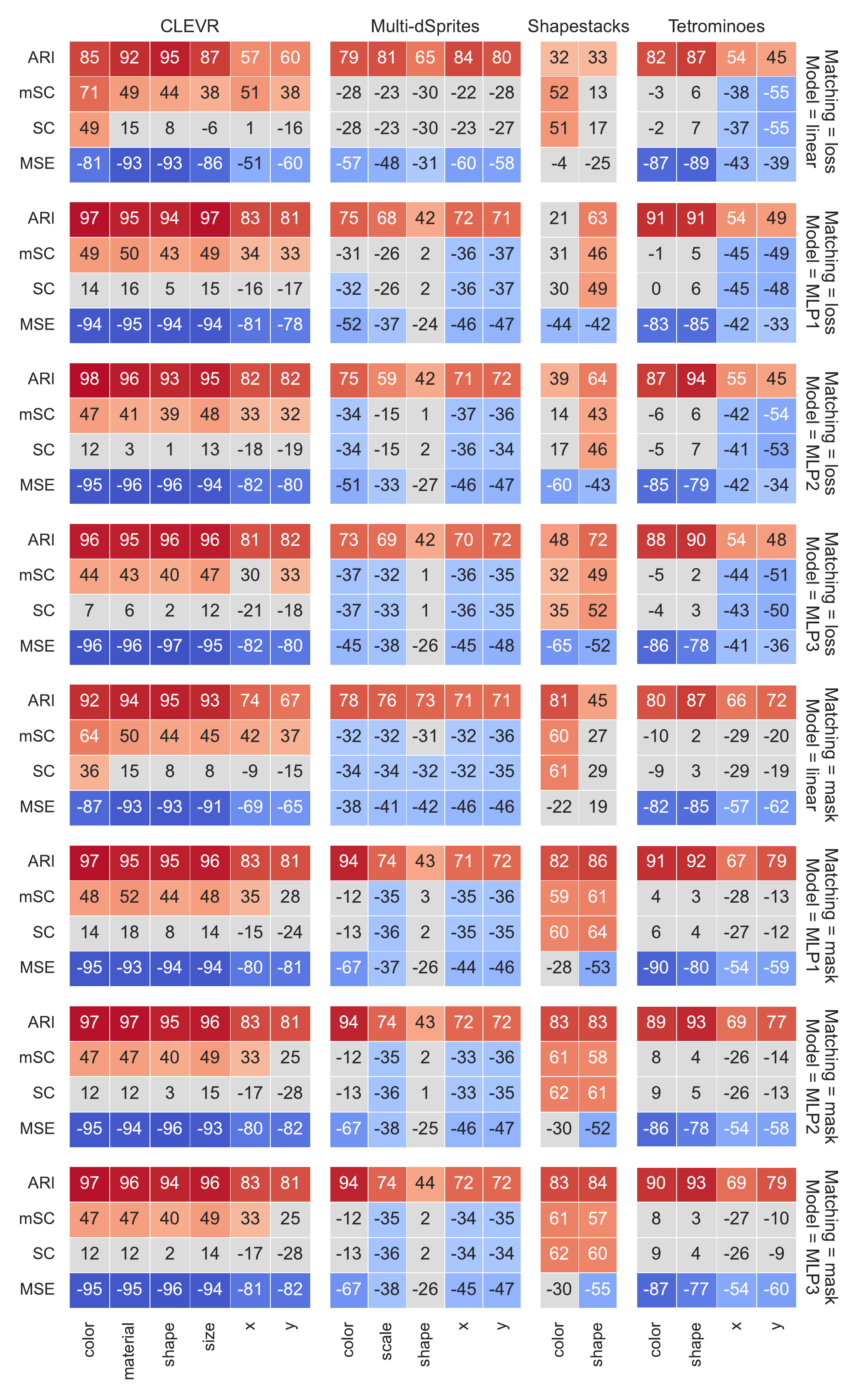}
    \caption{Spearman rank correlations between evaluation metrics and downstream performance with all considered combinations of slot matching (loss- and mask-based) and downstream model (linear, MLP with 1, 2, or 3 hidden layers). The correlations are color-coded only when p\textless 0.05.}
    \label{fig:indistrib/downstream/corr_metrics_downstream_all}
\end{figure}

In \cref{fig:indistrib/downstream/barplots_compare_downstream_models_loss}, we show the same results in a different way, to directly compare downstream models (here the median baselines for slot-based and distributed representations are shown as horizontal lines on top of the relevant bars). Using larger downstream models tends to slightly improve test performance but, interestingly, in many cases the effect is negligible. There are however a few cases in which using a larger model significantly boosts test performance in object property prediction. In some cases it seems sufficient to use a small MLP with one hidden layer instead of a linear model (e.g., color prediction in CLEVR with Slot Attention, shape prediction in CLEVR with MONet and Slot Attention, color prediction in Tetrominoes with MONet, GENESIS, and SPACE, or location prediction in Multi-dSprites with SPACE), while in other cases we get further gains by using even larger models (e.g., shape prediction in Multi-dSprites with SPACE, and shape prediction in Tetrominoes with all models except Slot Attention which already achieves a perfect score with a linear model).

Results for VAEs are generally less interpretable because the performance is often too close to the naive baseline. However, in some cases using deeper downstream models has clear benefits: e.g., shape prediction in Tetrominoes and color prediction in Shapestacks improve from baseline level when using a linear model to a relatively high accuracy when using one or two hidden layers. In other cases, a linear model already works relatively well even from distributed representations---although significantly worse than object-centric representations---and using deeper downstream models is not beneficial (e.g., color and size prediction in CLEVR). In many other cases, larger downstream models do not seem to be sufficient to improve performance from VAE representations, confirming that often the relevant information may not be easily accessible and suggesting that object-centric representations may be generally beneficial.

In \cref{fig:indistrib/downstream/corr_metrics_downstream_all} we show the Spearman rank correlations between evaluation metrics and downstream performance with all considered combinations of slot matching (loss- and mask-based) and downstream model (linear, MLP with 1, 2, or 3 hidden layers). The trends are broadly consistent in all combinations, except that correlations with ARI tend to be stronger (perhaps unsurprisingly) when using mask matching, and when using larger downstream models.

\clearpage
\subsection{Performance under distribution shifts}

\subsubsection{Segmentation and reconstruction}

\begin{figure}[hb]
    \centering
    \begin{minipage}{\linewidth}
        \includegraphics[width=\linewidth]{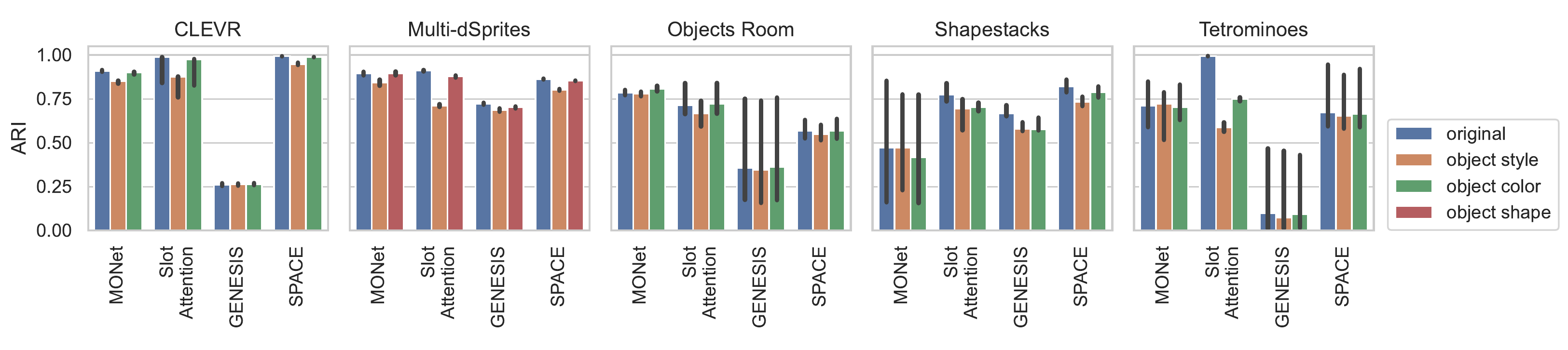}
        
        \vspace{10pt}
        \includegraphics[width=\linewidth]{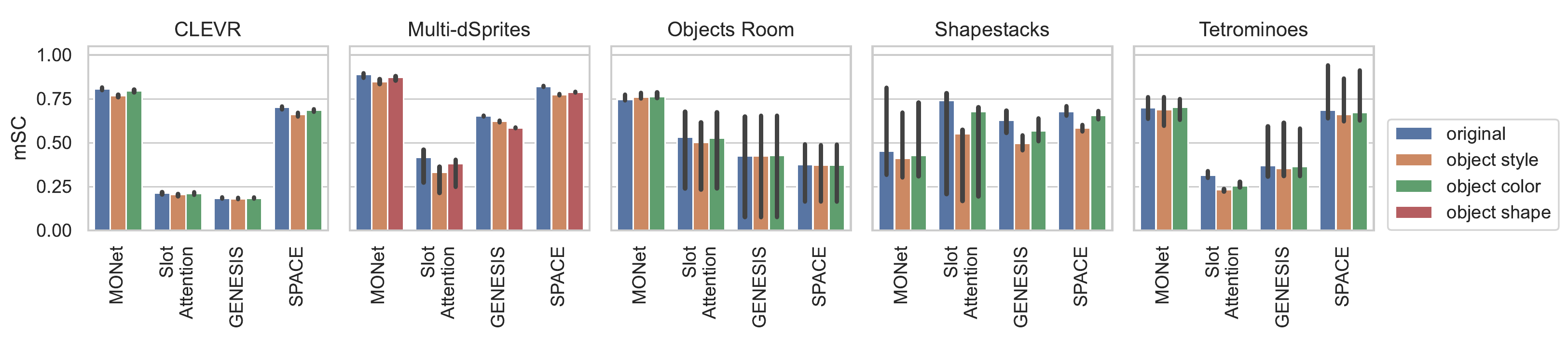}
        
        \vspace{10pt}
        \includegraphics[width=\linewidth]{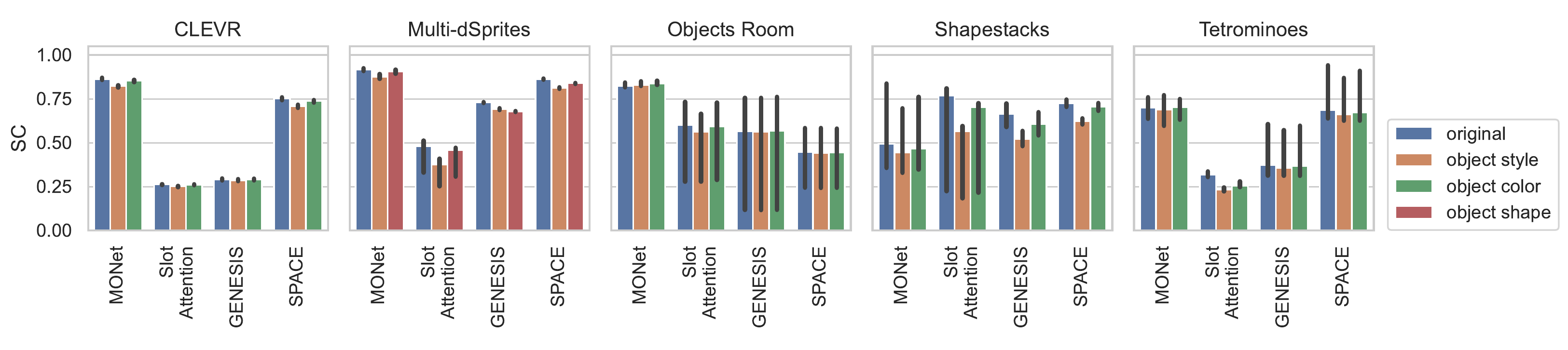}
        
        \vspace{10pt}
        \includegraphics[width=\linewidth]{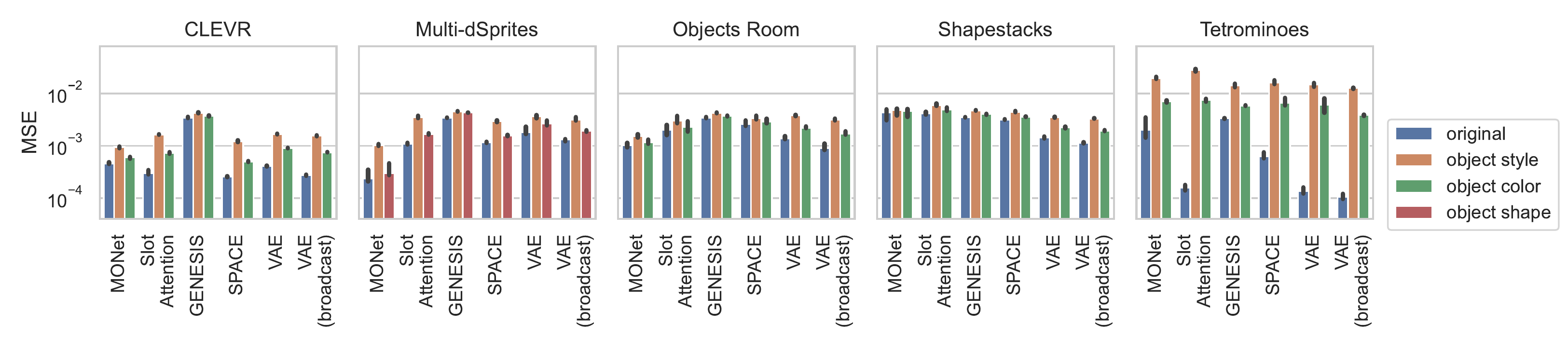}
    \end{minipage}
    \caption{Overview of segmentation metrics (ARI \up, mSC \up, SC \up) and reconstruction MSE (\down) on OOD dataset variants where \textbf{one object} undergoes distribution shift (test set of $\num{2000}$ images). The bars show medians and 95\% confidence intervals with 10 random seeds.}
    \label{fig:ood/metrics/generalization_compositional/box_plots}
\end{figure}

In \cref{fig:ood/metrics/generalization_compositional/box_plots}, we report the distributions of the reconstruction MSE and segmentation metrics in scenarios where one object is OOD. Results are split by dataset, model, and type of distribution shift.
As discussed in \cref{sec:results/hyp2}, the SC and mSC scores show a compatible but less pronounced trend, while the MSE more closely mirrors ARI. Notably, in many cases when we alter object style or color, the reconstruction MSE increases significantly while the ARI is only mildly affected. This suggests that the models are still capable of separating the objects but, unsurprisingly, they fail at reconstructing them properly as they have features that were never encountered during training.

\begin{figure}
    \centering
    \begin{minipage}{\linewidth}
        \includegraphics[width=\linewidth]{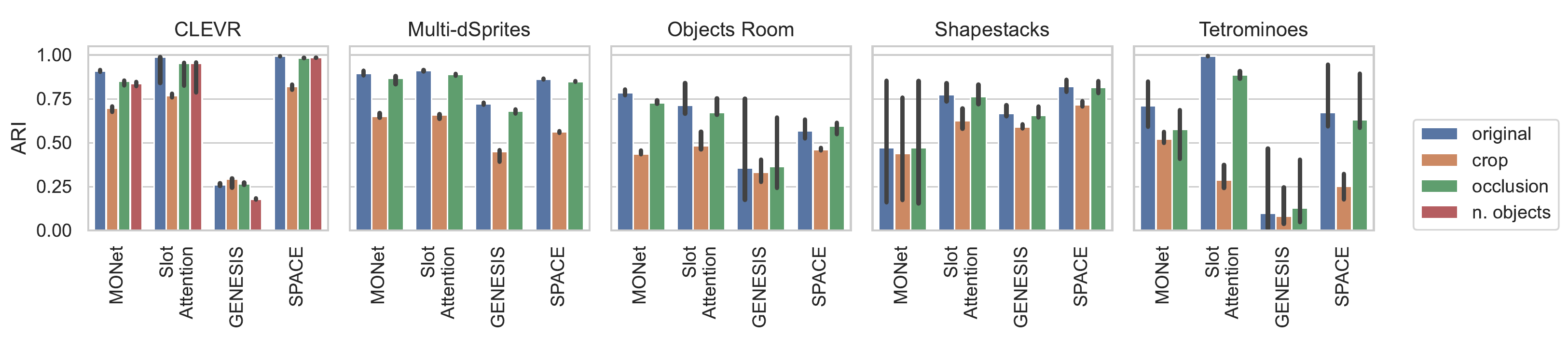}
        
        \vspace{10pt}
        \includegraphics[width=\linewidth]{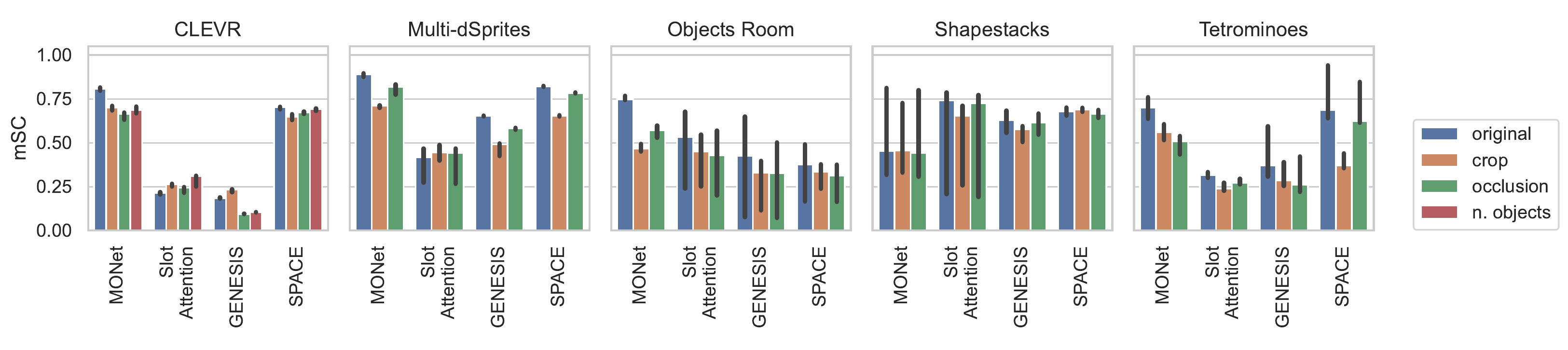}
        
        \vspace{10pt}
        \includegraphics[width=\linewidth]{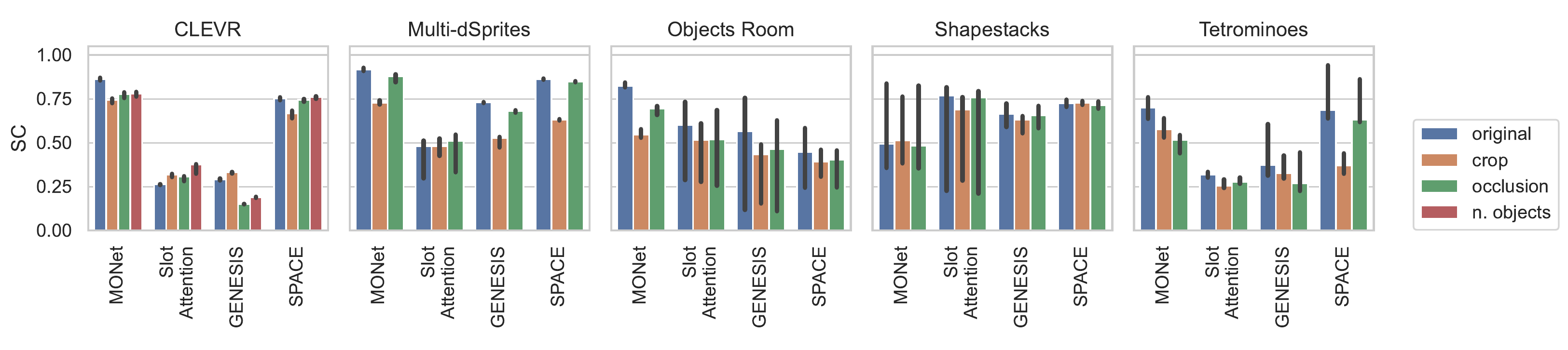}
        
        \vspace{10pt}
        \includegraphics[width=\linewidth]{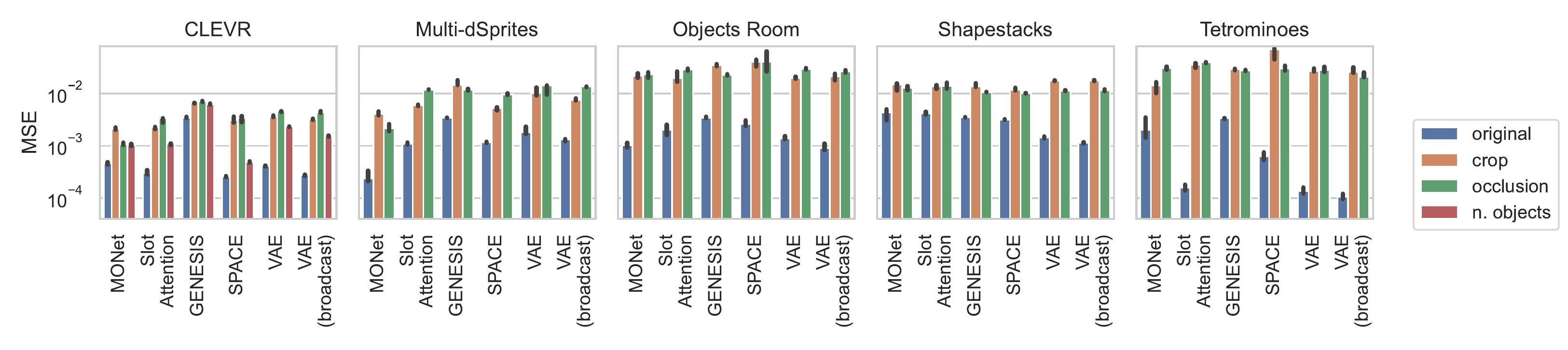}
    \end{minipage}
    \caption{Overview of segmentation metrics (ARI \up, mSC \up, SC \up) and reconstruction MSE (\down) on OOD dataset variants where \textbf{global properties} of the scene are altered (test set of $\num{2000}$ images). The bars show medians and 95\% confidence intervals with 10 random seeds.}
    \label{fig:ood/metrics/generalization_global/box_plots}
\end{figure}

\cref{fig:ood/metrics/generalization_global/box_plots} shows analogous results when the distribution shift affects global scene properties. Here we observe that segmentation performance is relatively robust to occlusion although the MSE increases significantly (as expected, the occlusion cannot be reconstructed properly). Segmentation metrics are also robust to increasing the number of objects in CLEVR---here the MSE also increases, but to a lesser extent, especially for SPACE.

\vspace{20pt}

\clearpage
\subsubsection{Downstream performance}

\begin{figure}[hb]
    \centering
    \includegraphics[width=\textwidth]{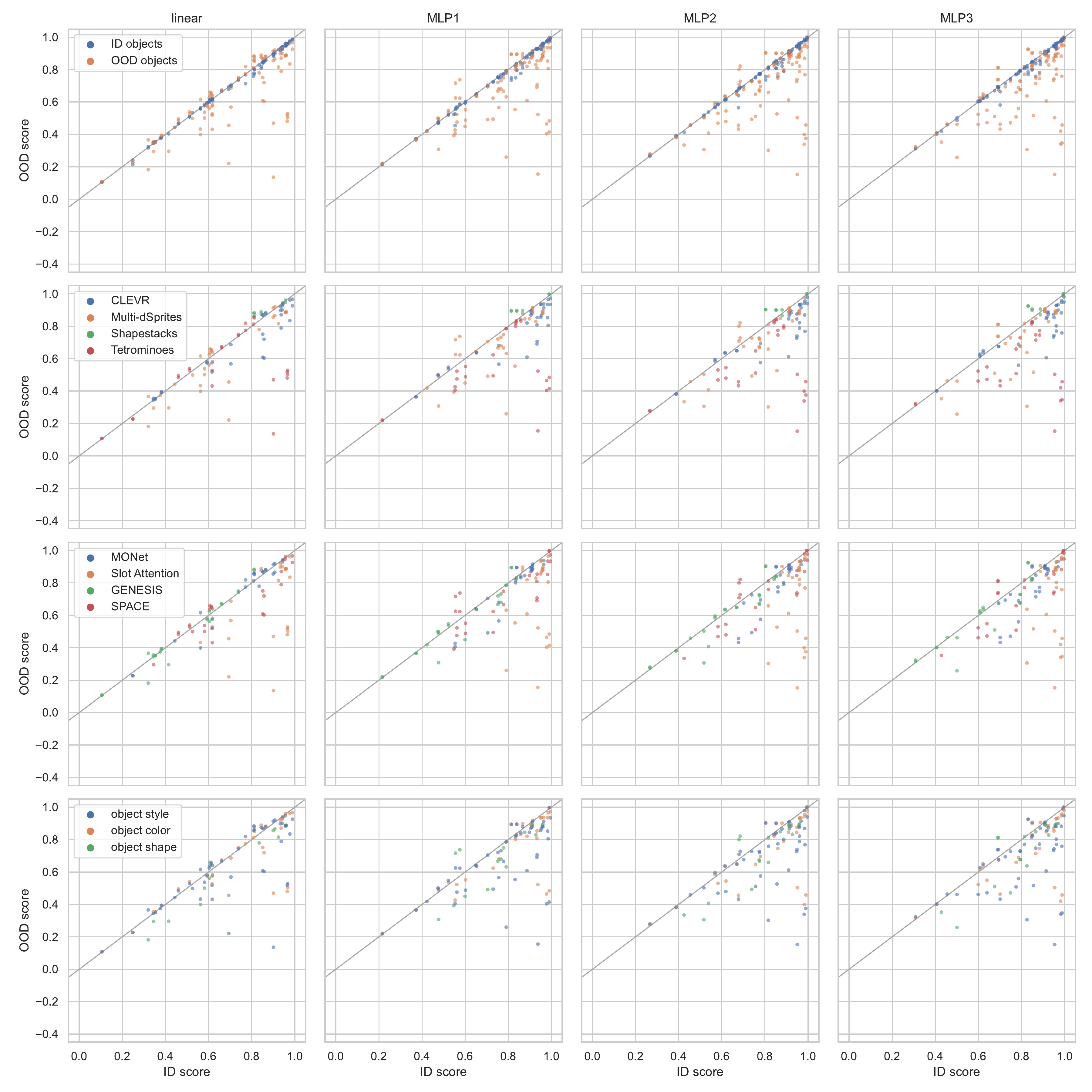}
    \caption{Generalization of \textbf{object-centric representations} in downstream prediction, using \textbf{loss matching} and \textbf{without retraining} the downstream model after the distribution shift. Here the distribution shift affects \textbf{one object}.
    On the x-axis: prediction performance (accuracy or \rsq) for one object property on one dataset, averaged over all objects, on the original training set of the unsupervised object discovery model. On the y-axis: the same metric in OOD scenarios. Each data point corresponds to one representation model (e.g., MONet), one dataset, one object property, one type of distribution shift, and either ID or OOD objects. For each x (performance on one object feature in the training distribution, averaged over objects in a scene and over random seeds of the object-centric models) there are multiple y's, corresponding to different distribution shifts and to ID/OOD objects.
    In the top row, we separately report (color-coded) the performance over ID and OOD objects.
    In the following rows, we only show OOD objects and split according to dataset, model, or type of distribution shift.
    Each column shows analogous results for each of the 4 considered downstream models for property prediction (linear, and MLPs with up to 3 hidden layers).
    }
    \label{fig:ood/downstream/slotted/match_loss/object/retrain_False}
\end{figure}

\begin{figure}
    \centering
    \includegraphics[width=\textwidth]{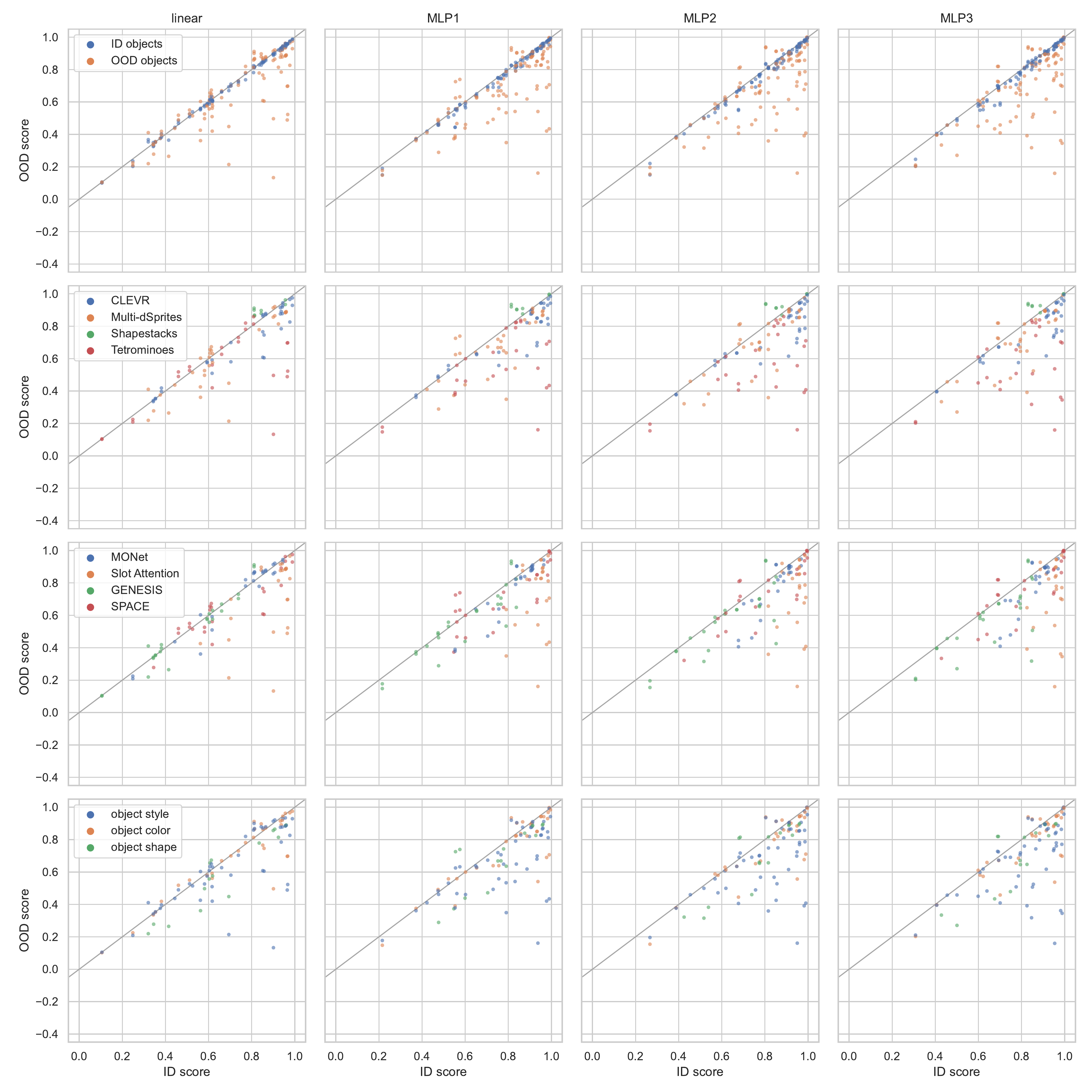}
    \caption{Generalization of \textbf{object-centric representations} in downstream prediction, using \textbf{loss matching} and \textbf{retraining} the downstream model after the distribution shift. Here the distribution shift affects \textbf{one object}.
    On the x-axis: prediction performance (accuracy or \rsq) for one object property on one dataset, averaged over all objects, on the original training set of the unsupervised object discovery model. On the y-axis: the same metric in OOD scenarios. Each data point corresponds to one representation model (e.g., MONet), one dataset, one object property, one type of distribution shift, and either ID or OOD objects. For each x (performance on one object feature in the training distribution, averaged over objects in a scene and over random seeds of the object-centric models) there are multiple y's, corresponding to different distribution shifts and to ID/OOD objects.
    In the top row, we separately report (color-coded) the performance over ID and OOD objects.
    In the following rows, we only show OOD objects and split according to dataset, model, or type of distribution shift.
    Each column shows analogous results for each of the 4 considered downstream models for property prediction (linear, and MLPs with up to 3 hidden layers).
    }
    \label{fig:ood/downstream/slotted/match_loss/object/retrain_True}
\end{figure}

\begin{figure}
    \centering
    \includegraphics[width=\textwidth]{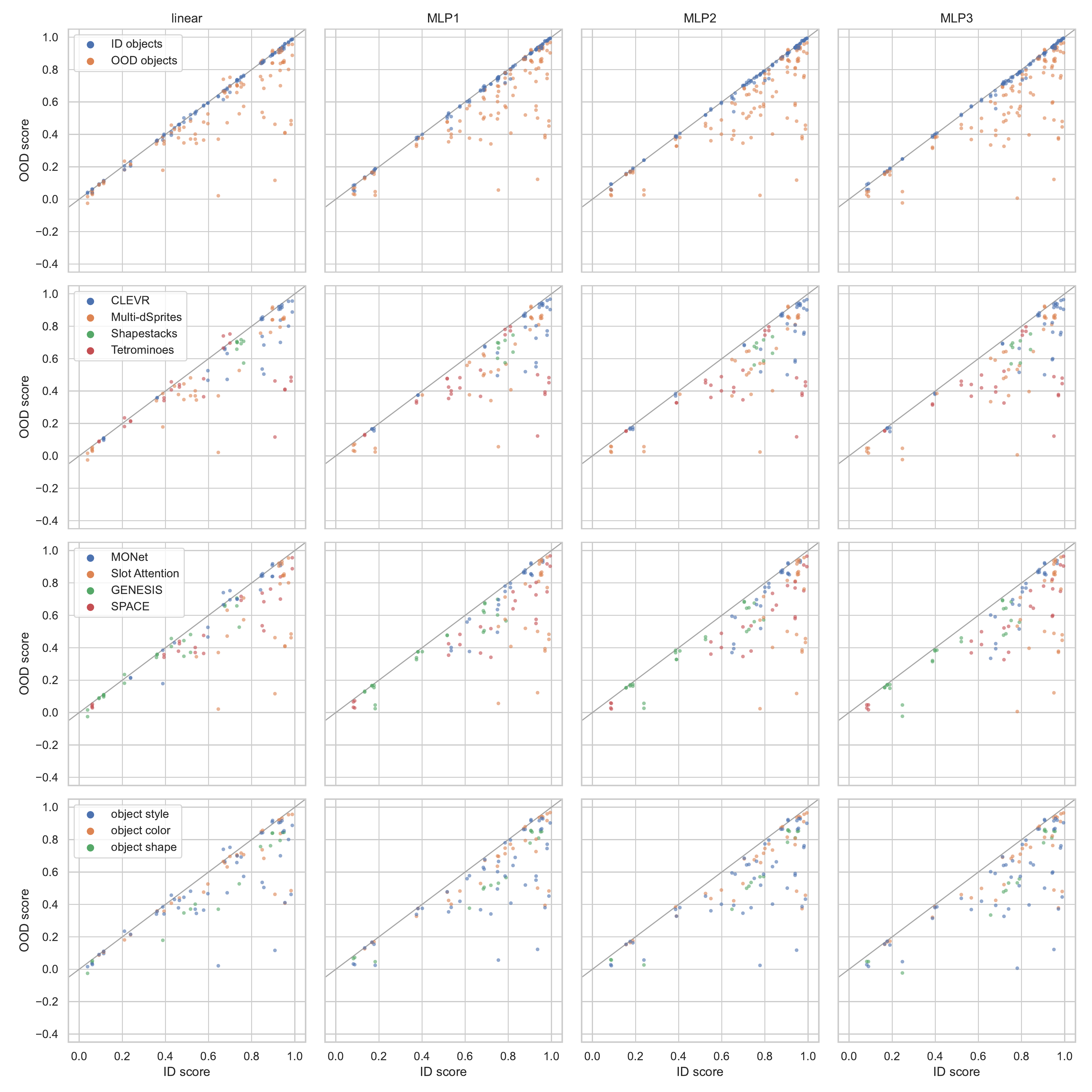}
    \caption{Generalization of \textbf{object-centric representations} in downstream prediction, using \textbf{mask matching} and \textbf{without retraining} the downstream model after the distribution shift. Here the distribution shift affects \textbf{one object}.
    On the x-axis: prediction performance (accuracy or \rsq) for one object property on one dataset, averaged over all objects, on the original training set of the unsupervised object discovery model. On the y-axis: the same metric in OOD scenarios. Each data point corresponds to one representation model (e.g., MONet), one dataset, one object property, one type of distribution shift, and either ID or OOD objects. For each x (performance on one object feature in the training distribution, averaged over objects in a scene and over random seeds of the object-centric models) there are multiple y's, corresponding to different distribution shifts and to ID/OOD objects.
    In the top row, we separately report (color-coded) the performance over ID and OOD objects.
    In the following rows, we only show OOD objects and split according to dataset, model, or type of distribution shift.
    Each column shows analogous results for each of the 4 considered downstream models for property prediction (linear, and MLPs with up to 3 hidden layers).
    }
    \label{fig:ood/downstream/slotted/match_mask/object/retrain_False}
\end{figure}

\begin{figure}
    \centering
    \includegraphics[width=\textwidth]{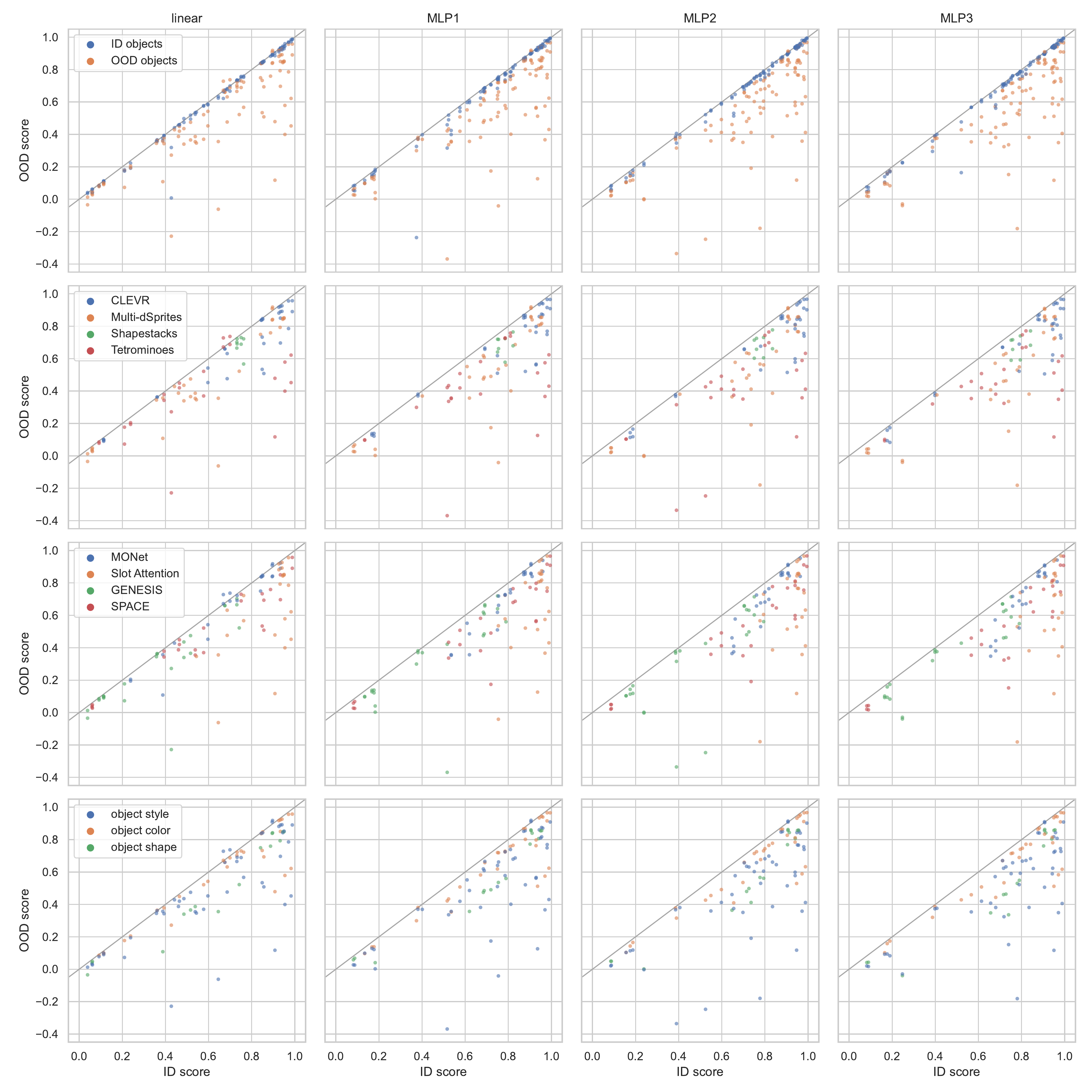}
    \caption{Generalization of \textbf{object-centric representations} in downstream prediction, using \textbf{mask matching} and \textbf{retraining} the downstream model after the distribution shift. Here the distribution shift affects \textbf{one object}.
    On the x-axis: prediction performance (accuracy or \rsq) for one object property on one dataset, averaged over all objects, on the original training set of the unsupervised object discovery model. On the y-axis: the same metric in OOD scenarios. Each data point corresponds to one representation model (e.g., MONet), one dataset, one object property, one type of distribution shift, and either ID or OOD objects. For each x (performance on one object feature in the training distribution, averaged over objects in a scene and over random seeds of the object-centric models) there are multiple y's, corresponding to different distribution shifts and to ID/OOD objects.
    In the top row, we separately report (color-coded) the performance over ID and OOD objects.
    In the following rows, we only show OOD objects and split according to dataset, model, or type of distribution shift.
    Each column shows analogous results for each of the 4 considered downstream models for property prediction (linear, and MLPs with up to 3 hidden layers).
    }
    \label{fig:ood/downstream/slotted/match_mask/object/retrain_True}
\end{figure}

\begin{figure}
    \centering
    \includegraphics[width=\textwidth]{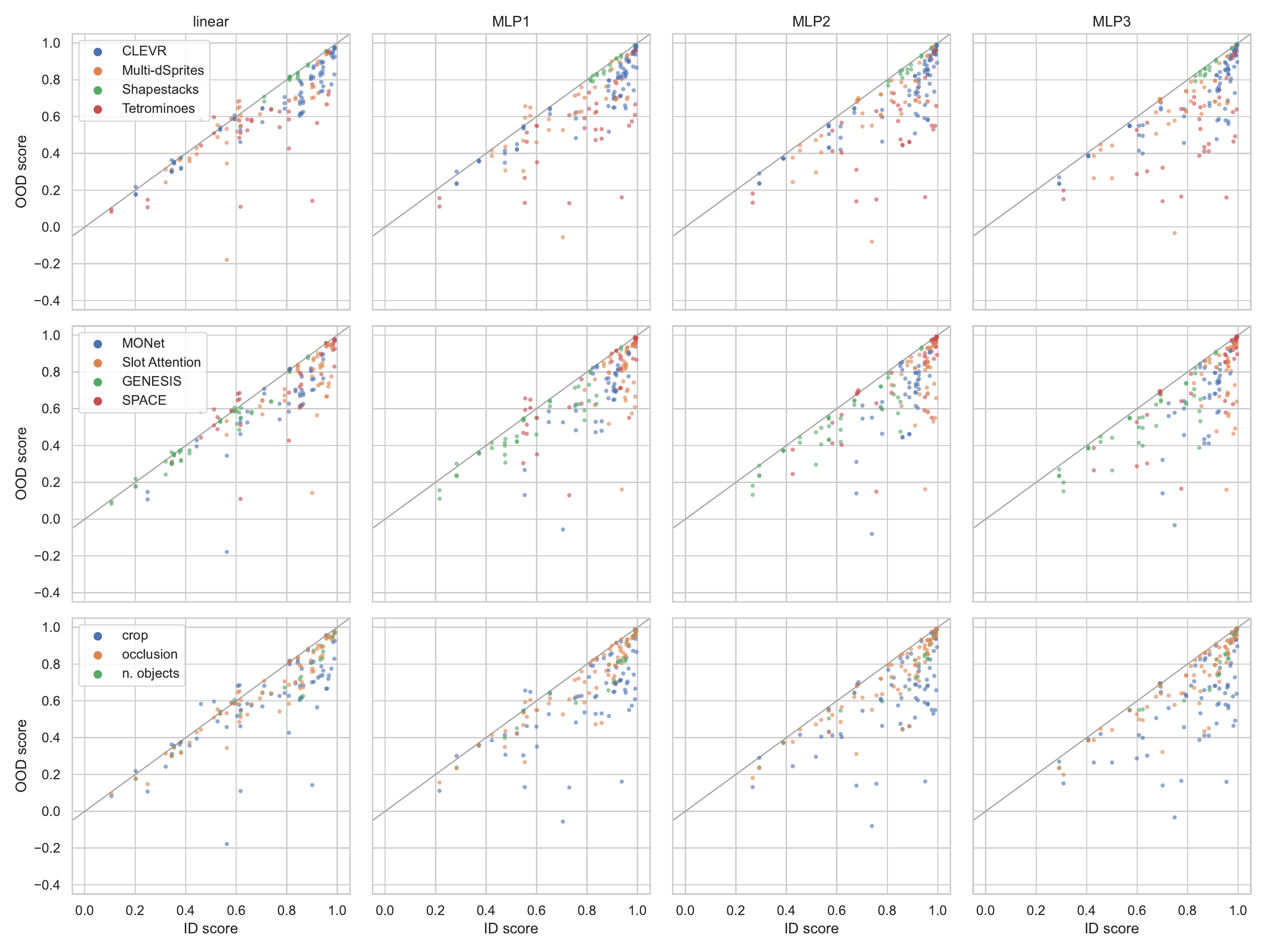}
    \caption{Generalization of \textbf{object-centric representations} in downstream prediction, using \textbf{loss matching} and \textbf{without retraining} the downstream model after the distribution shift. Here the distribution shift affects \textbf{global properties} of the scene.
        On the x-axis: prediction performance (accuracy or \rsq) for one object property on one dataset, averaged over all objects, on the original training set of the unsupervised object discovery model. On the y-axis: the same metric in OOD scenarios. Each data point corresponds to one representation model (e.g., MONet), one dataset, one object property, and one type of distribution shift. For each x (performance on one object feature in the training distribution, averaged over objects in a scene and over random seeds of the object-centric models) there are multiple y's, corresponding to different distribution shifts.
        In each row, we color-code the data according to dataset, model, or type of distribution shift.
        Each column shows analogous results for each of the 4 considered downstream models for property prediction (linear, and MLPs with up to 3 hidden layers).
    }
    \label{fig:ood/downstream/slotted/match_loss/global/retrain_False}
\end{figure}

\begin{figure}
    \centering
    \includegraphics[width=\textwidth]{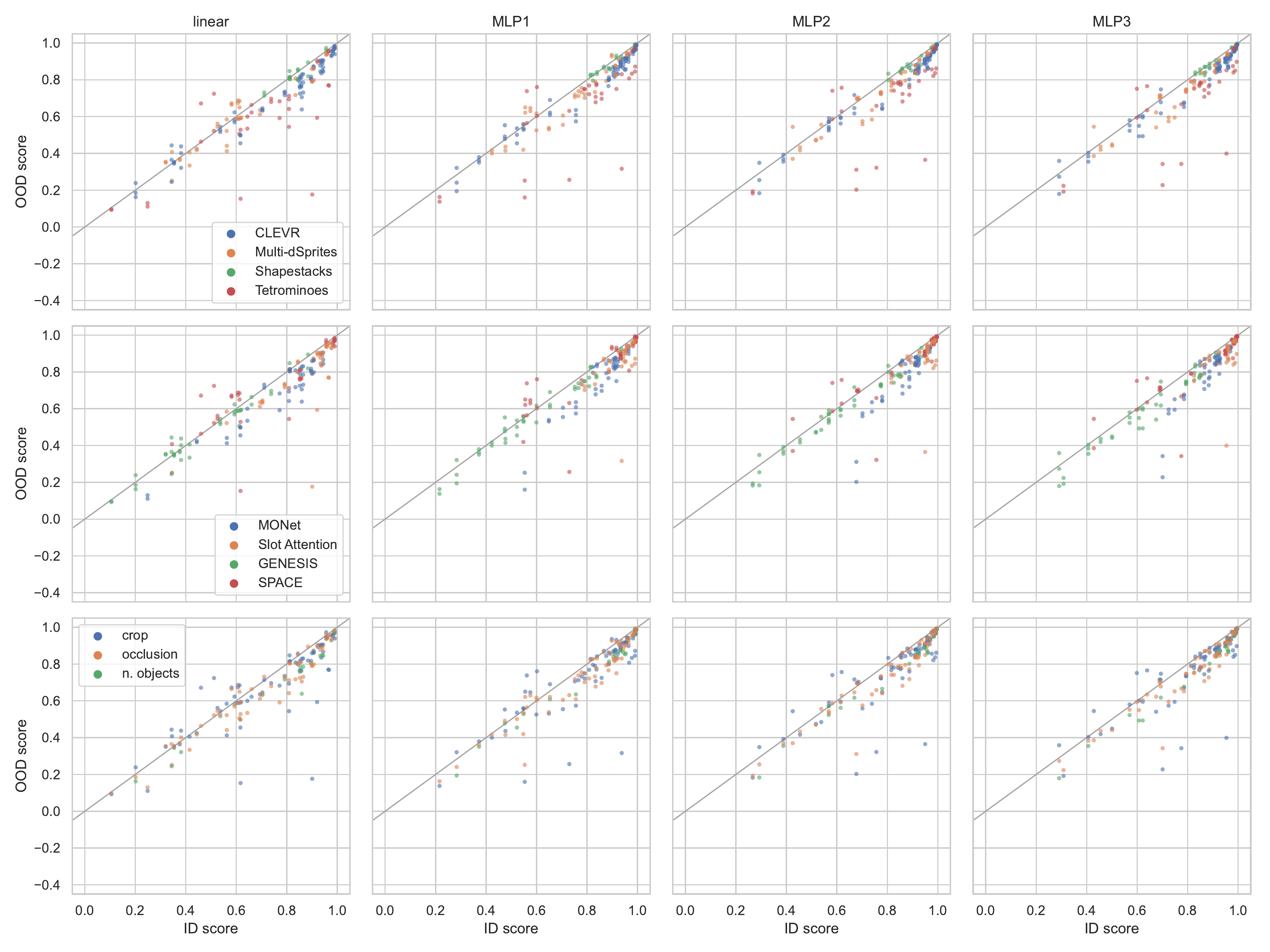}
    \caption{Generalization of \textbf{object-centric representations} in downstream prediction, using \textbf{loss matching} and \textbf{retraining} the downstream model after the distribution shift. Here the distribution shift affects \textbf{global properties} of the scene.
        On the x-axis: prediction performance (accuracy or \rsq) for one object property on one dataset, averaged over all objects, on the original training set of the unsupervised object discovery model. On the y-axis: the same metric in OOD scenarios. Each data point corresponds to one representation model (e.g., MONet), one dataset, one object property, and one type of distribution shift. For each x (performance on one object feature in the training distribution, averaged over objects in a scene and over random seeds of the object-centric models) there are multiple y's, corresponding to different distribution shifts.
        In each row, we color-code the data according to dataset, model, or type of distribution shift.
        Each column shows analogous results for each of the 4 considered downstream models for property prediction (linear, and MLPs with up to 3 hidden layers).
    }
    \label{fig:ood/downstream/slotted/match_loss/global/retrain_True}
\end{figure}

\begin{figure}
    \centering
    \includegraphics[width=\textwidth]{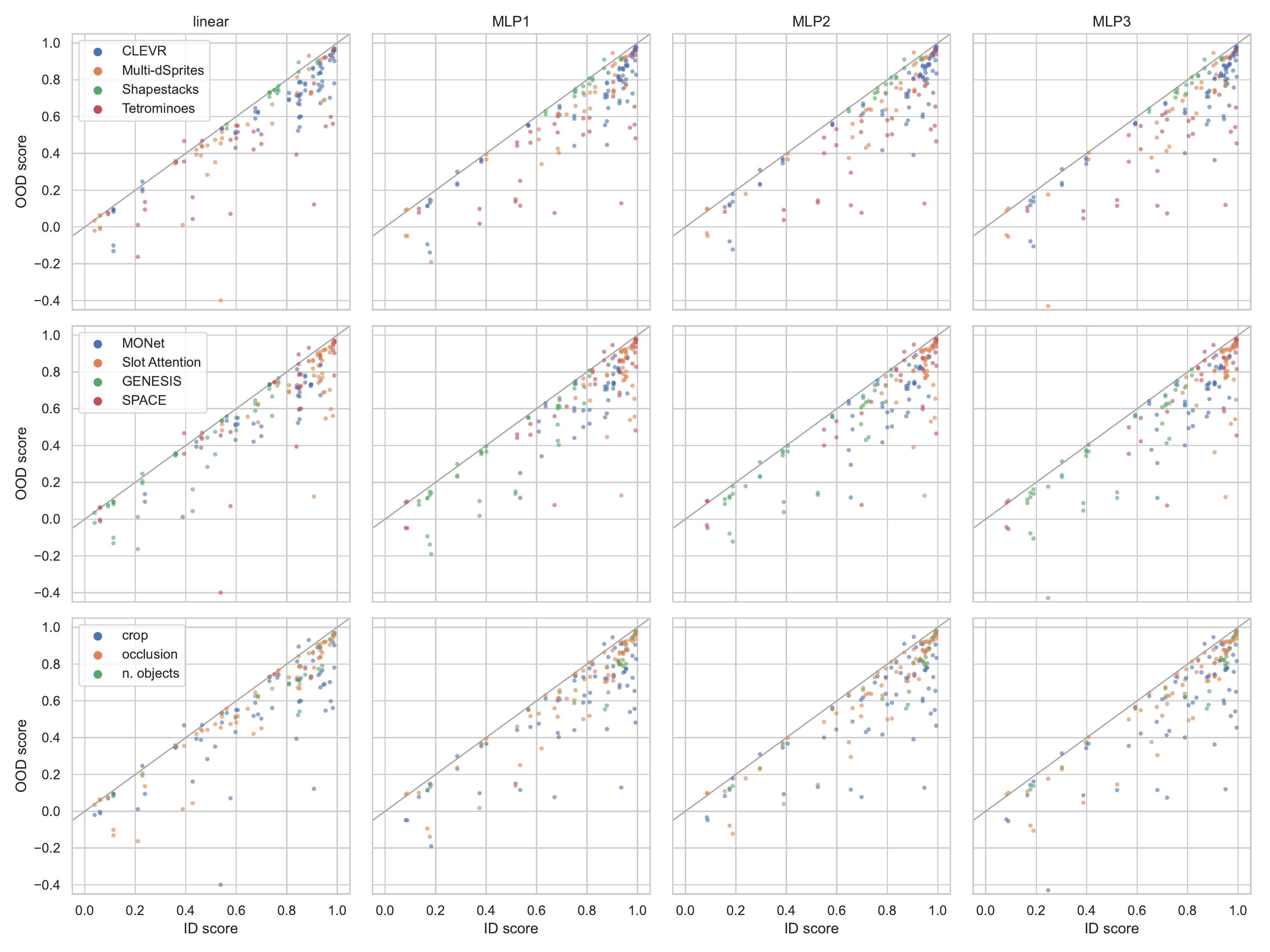}
    \caption{Generalization of \textbf{object-centric representations} in downstream prediction, using \textbf{mask matching} and \textbf{without retraining} the downstream model after the distribution shift. Here the distribution shift affects \textbf{global properties} of the scene.
        On the x-axis: prediction performance (accuracy or \rsq) for one object property on one dataset, averaged over all objects, on the original training set of the unsupervised object discovery model. On the y-axis: the same metric in OOD scenarios. Each data point corresponds to one representation model (e.g., MONet), one dataset, one object property, and one type of distribution shift. For each x (performance on one object feature in the training distribution, averaged over objects in a scene and over random seeds of the object-centric models) there are multiple y's, corresponding to different distribution shifts.
        In each row, we color-code the data according to dataset, model, or type of distribution shift.
        Each column shows analogous results for each of the 4 considered downstream models for property prediction (linear, and MLPs with up to 3 hidden layers).
    }
    \label{fig:ood/downstream/slotted/match_mask/global/retrain_False}
\end{figure}

\begin{figure}
    \centering
    \includegraphics[width=\textwidth]{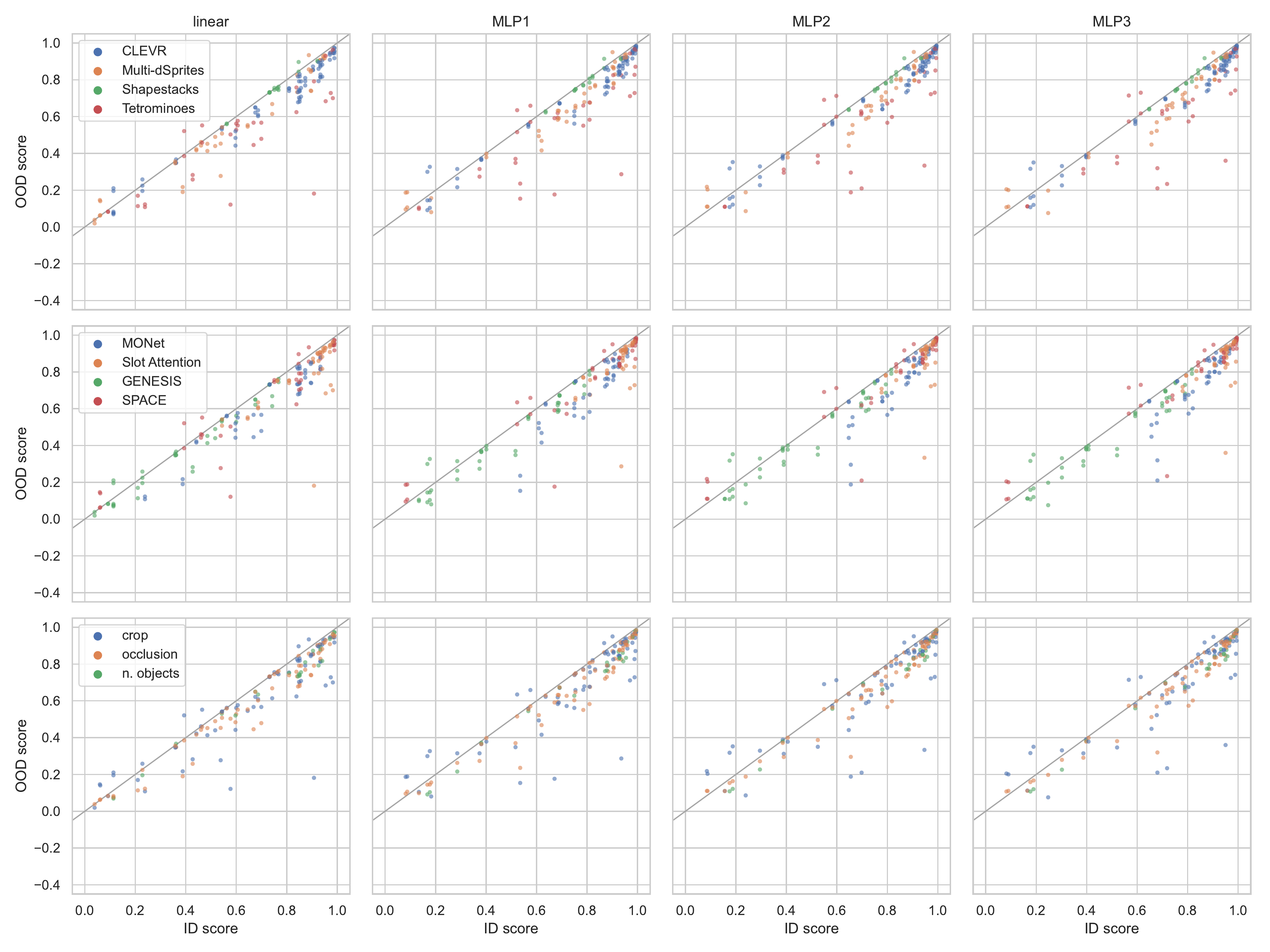}
    \caption{Generalization of \textbf{object-centric representations} in downstream prediction, using \textbf{mask matching} and \textbf{retraining} the downstream model after the distribution shift. Here the distribution shift affects \textbf{global properties} of the scene.
        On the x-axis: prediction performance (accuracy or \rsq) for one object property on one dataset, averaged over all objects, on the original training set of the unsupervised object discovery model. On the y-axis: the same metric in OOD scenarios. Each data point corresponds to one representation model (e.g., MONet), one dataset, one object property, and one type of distribution shift. For each x (performance on one object feature in the training distribution, averaged over objects in a scene and over random seeds of the object-centric models) there are multiple y's, corresponding to different distribution shifts.
        In each row, we color-code the data according to dataset, model, or type of distribution shift.
        Each column shows analogous results for each of the 4 considered downstream models for property prediction (linear, and MLPs with up to 3 hidden layers).
    }
    \label{fig:ood/downstream/slotted/match_mask/global/retrain_True}
\end{figure}

\begin{figure}
    \centering
    \includegraphics[width=\textwidth]{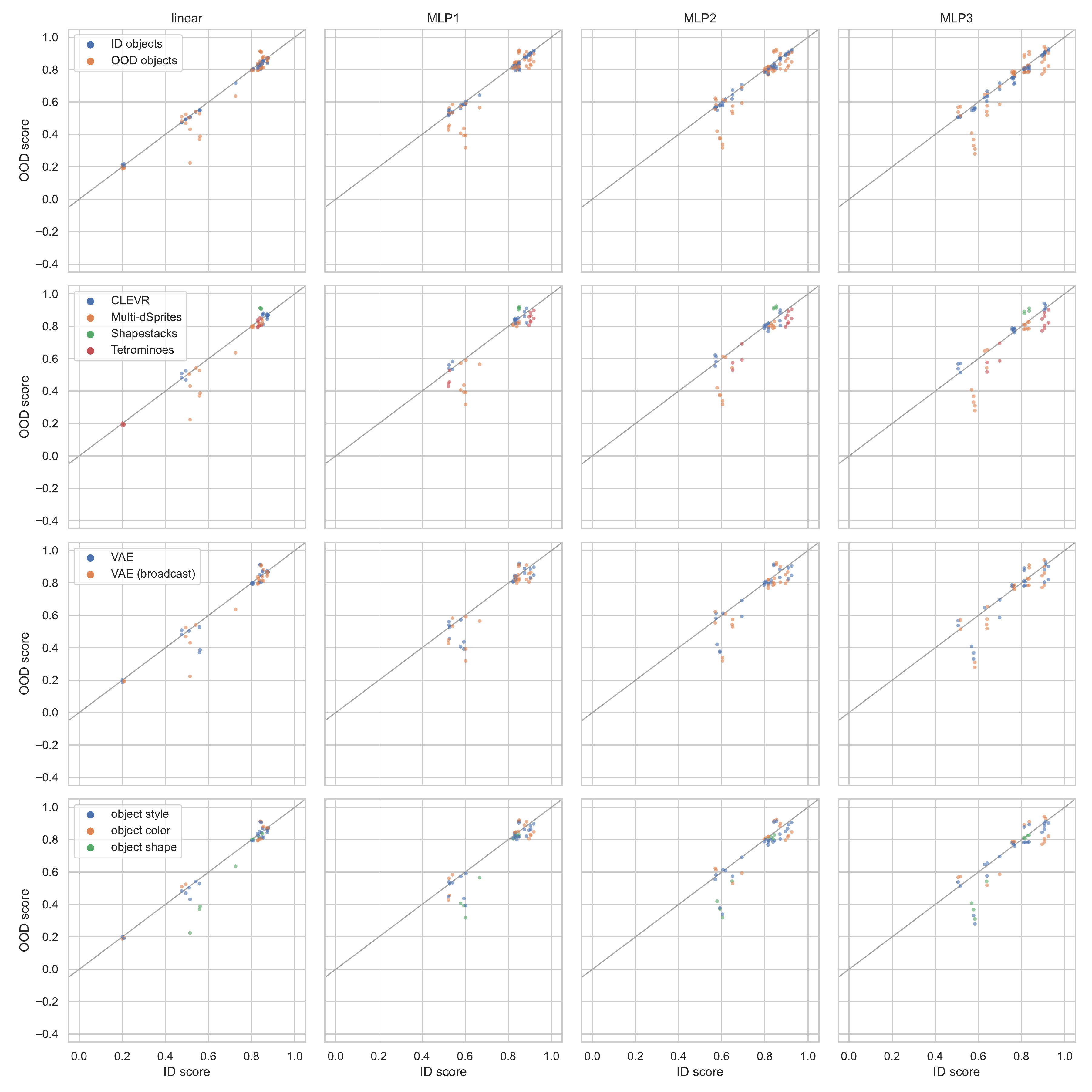}
    \caption{Generalization of \textbf{distributed representations} in downstream prediction, using \textbf{loss matching} and \textbf{without retraining} the downstream model after the distribution shift. Here the distribution shift affects \textbf{one object}.
    On the x-axis: prediction performance (accuracy or \rsq) for one object property on one dataset, averaged over all objects, on the original training set of the unsupervised object discovery model. On the y-axis: the same metric in OOD scenarios. Each data point corresponds to one representation model (e.g., MONet), one dataset, one object property, one type of distribution shift, and either ID or OOD objects. For each x (performance on one object feature in the training distribution, averaged over objects in a scene and over random seeds of the object-centric models) there are multiple y's, corresponding to different distribution shifts and to ID/OOD objects.
    In the top row, we separately report (color-coded) the performance over ID and OOD objects.
    In the following rows, we only show OOD objects and split according to dataset, model, or type of distribution shift.
    Each column shows analogous results for each of the 4 considered downstream models for property prediction (linear, and MLPs with up to 3 hidden layers).
    }
    \label{fig:ood/downstream/vae/match_loss/object/retrain_False}
\end{figure}

\begin{figure}
    \centering
    \includegraphics[width=\textwidth]{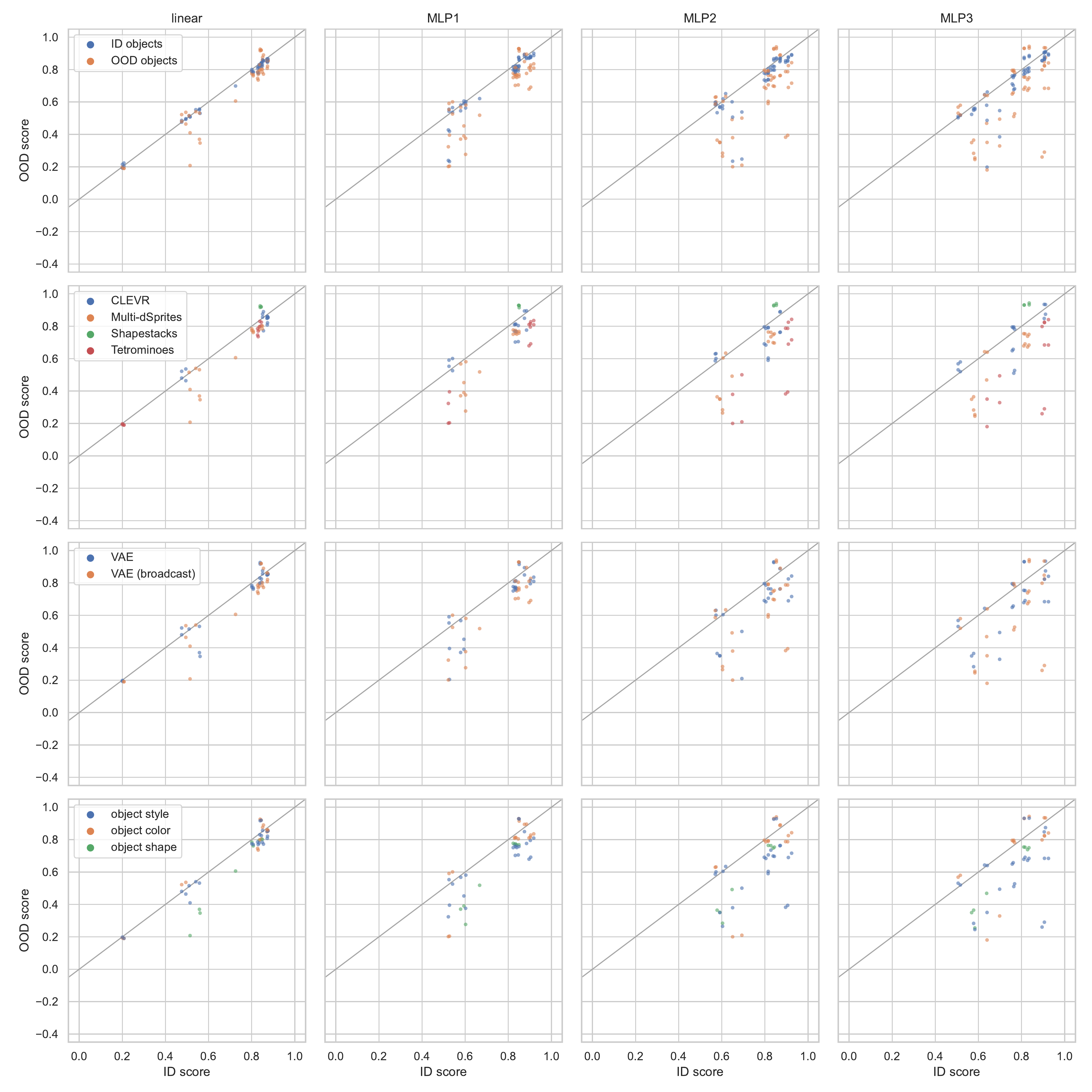}
    \caption{Generalization of \textbf{distributed representations} in downstream prediction, using \textbf{loss matching} and \textbf{retraining} the downstream model after the distribution shift. Here the distribution shift affects \textbf{one object}.
    On the x-axis: prediction performance (accuracy or \rsq) for one object property on one dataset, averaged over all objects, on the original training set of the unsupervised object discovery model. On the y-axis: the same metric in OOD scenarios. Each data point corresponds to one representation model (e.g., MONet), one dataset, one object property, one type of distribution shift, and either ID or OOD objects. For each x (performance on one object feature in the training distribution, averaged over objects in a scene and over random seeds of the object-centric models) there are multiple y's, corresponding to different distribution shifts and to ID/OOD objects.
    In the top row, we separately report (color-coded) the performance over ID and OOD objects.
    In the following rows, we only show OOD objects and split according to dataset, model, or type of distribution shift.
    Each column shows analogous results for each of the 4 considered downstream models for property prediction (linear, and MLPs with up to 3 hidden layers).
    }
    \label{fig:ood/downstream/vae/match_loss/object/retrain_True}
\end{figure}

\begin{figure}
    \centering
    \includegraphics[width=\textwidth]{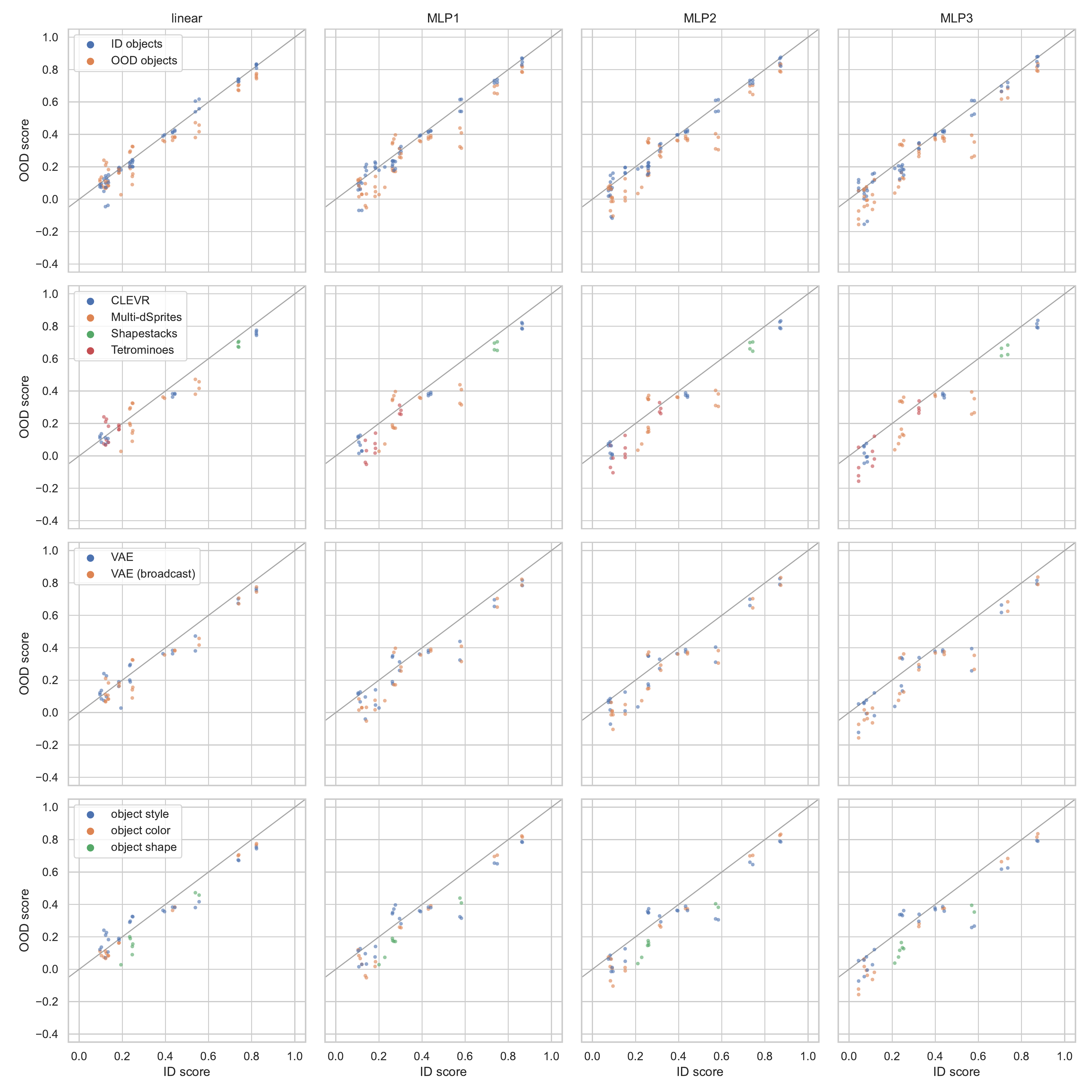}
    \caption{Generalization of \textbf{distributed representations} in downstream prediction, using \textbf{deterministic matching} and \textbf{without retraining} the downstream model after the distribution shift. Here the distribution shift affects \textbf{one object}.
    On the x-axis: prediction performance (accuracy or \rsq) for one object property on one dataset, averaged over all objects, on the original training set of the unsupervised object discovery model. On the y-axis: the same metric in OOD scenarios. Each data point corresponds to one representation model (e.g., MONet), one dataset, one object property, one type of distribution shift, and either ID or OOD objects. For each x (performance on one object feature in the training distribution, averaged over objects in a scene and over random seeds of the object-centric models) there are multiple y's, corresponding to different distribution shifts and to ID/OOD objects.
    In the top row, we separately report (color-coded) the performance over ID and OOD objects.
    In the following rows, we only show OOD objects and split according to dataset, model, or type of distribution shift.
    Each column shows analogous results for each of the 4 considered downstream models for property prediction (linear, and MLPs with up to 3 hidden layers).
    }
    \label{fig:ood/downstream/vae/match_deterministic/object/retrain_False}
\end{figure}

\begin{figure}
    \centering
    \includegraphics[width=\textwidth]{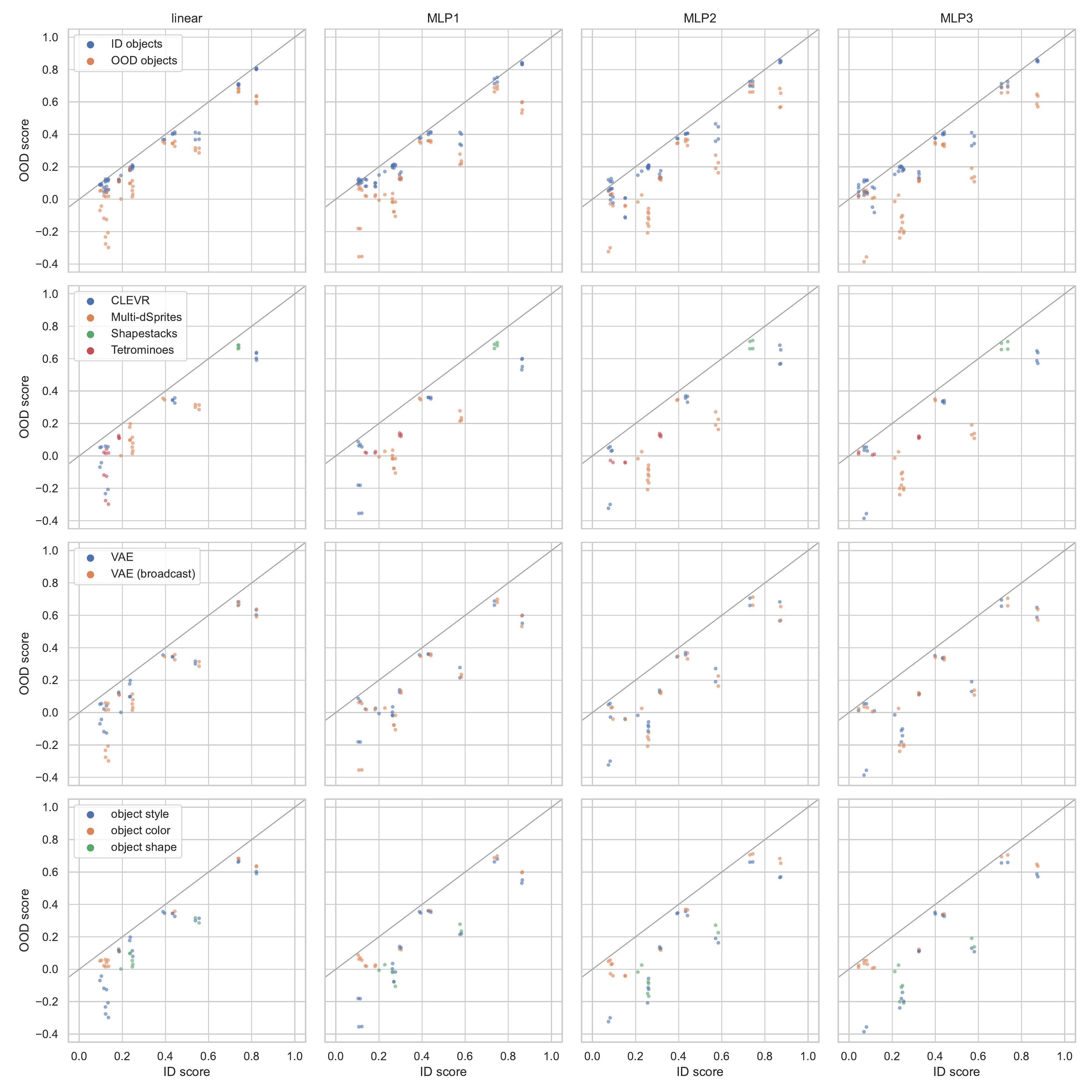}
    \caption{Generalization of \textbf{distributed representations} in downstream prediction, using \textbf{deterministic matching} and \textbf{retraining} the downstream model after the distribution shift. Here the distribution shift affects \textbf{one object}.
    On the x-axis: prediction performance (accuracy or \rsq) for one object property on one dataset, averaged over all objects, on the original training set of the unsupervised object discovery model. On the y-axis: the same metric in OOD scenarios. Each data point corresponds to one representation model (e.g., MONet), one dataset, one object property, one type of distribution shift, and either ID or OOD objects. For each x (performance on one object feature in the training distribution, averaged over objects in a scene and over random seeds of the object-centric models) there are multiple y's, corresponding to different distribution shifts and to ID/OOD objects.
    In the top row, we separately report (color-coded) the performance over ID and OOD objects.
    In the following rows, we only show OOD objects and split according to dataset, model, or type of distribution shift.
    Each column shows analogous results for each of the 4 considered downstream models for property prediction (linear, and MLPs with up to 3 hidden layers).
    }
    \label{fig:ood/downstream/vae/match_deterministic/object/retrain_True}
\end{figure}

\begin{figure}
    \centering
    \includegraphics[width=\textwidth]{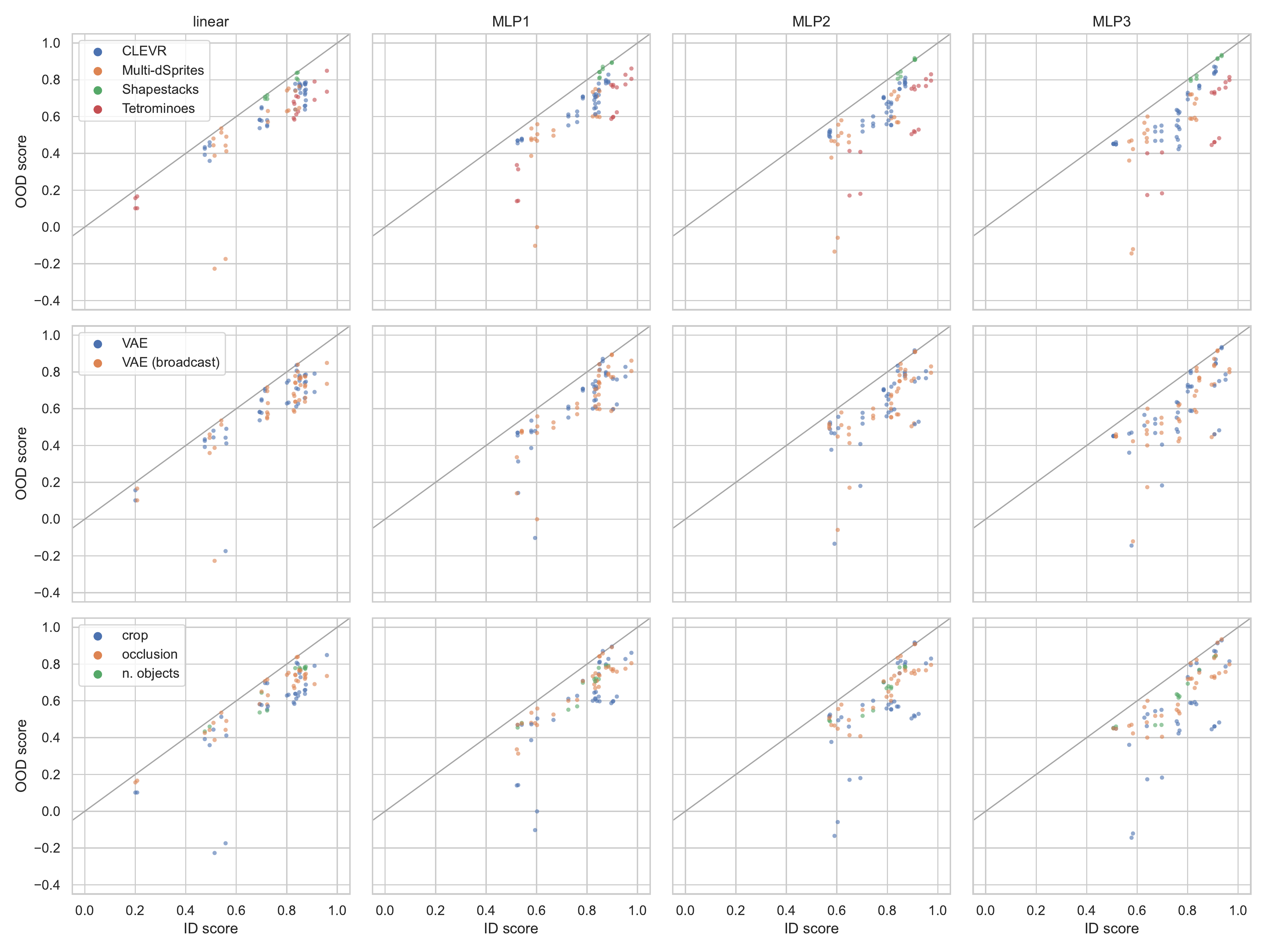}
    \caption{Generalization of \textbf{distributed representations} in downstream prediction, using \textbf{loss matching} and \textbf{without retraining} the downstream model after the distribution shift. Here the distribution shift affects \textbf{global properties} of the scene.
        On the x-axis: prediction performance (accuracy or \rsq) for one object property on one dataset, averaged over all objects, on the original training set of the unsupervised object discovery model. On the y-axis: the same metric in OOD scenarios. Each data point corresponds to one representation model (e.g., MONet), one dataset, one object property, and one type of distribution shift. For each x (performance on one object feature in the training distribution, averaged over objects in a scene and over random seeds of the object-centric models) there are multiple y's, corresponding to different distribution shifts.
        In each row, we color-code the data according to dataset, model, or type of distribution shift.
        Each column shows analogous results for each of the 4 considered downstream models for property prediction (linear, and MLPs with up to 3 hidden layers).
    }
    \label{fig:ood/downstream/vae/match_loss/global/retrain_False/scatter}
\end{figure}

\begin{figure}
    \centering
    \includegraphics[width=\textwidth]{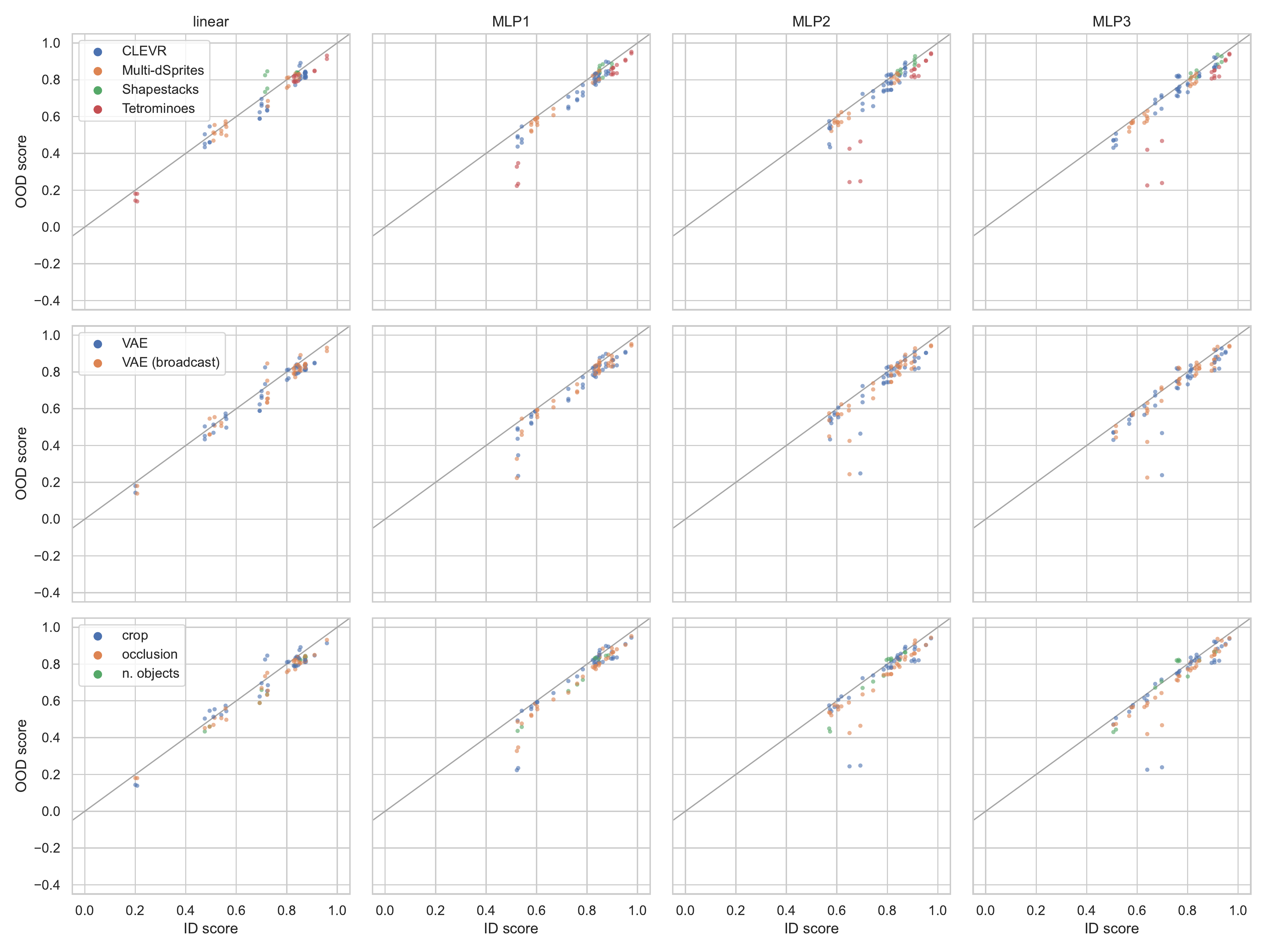}
    \caption{Generalization of \textbf{distributed representations} in downstream prediction, using \textbf{loss matching} and \textbf{retraining} the downstream model after the distribution shift. Here the distribution shift affects \textbf{global properties} of the scene.
        On the x-axis: prediction performance (accuracy or \rsq) for one object property on one dataset, averaged over all objects, on the original training set of the unsupervised object discovery model. On the y-axis: the same metric in OOD scenarios. Each data point corresponds to one representation model (e.g., MONet), one dataset, one object property, and one type of distribution shift. For each x (performance on one object feature in the training distribution, averaged over objects in a scene and over random seeds of the object-centric models) there are multiple y's, corresponding to different distribution shifts.
        In each row, we color-code the data according to dataset, model, or type of distribution shift.
        Each column shows analogous results for each of the 4 considered downstream models for property prediction (linear, and MLPs with up to 3 hidden layers).
    }
    \label{fig:ood/downstream/vae/match_loss/global/retrain_True}
\end{figure}

\begin{figure}
    \centering
    \includegraphics[width=\textwidth]{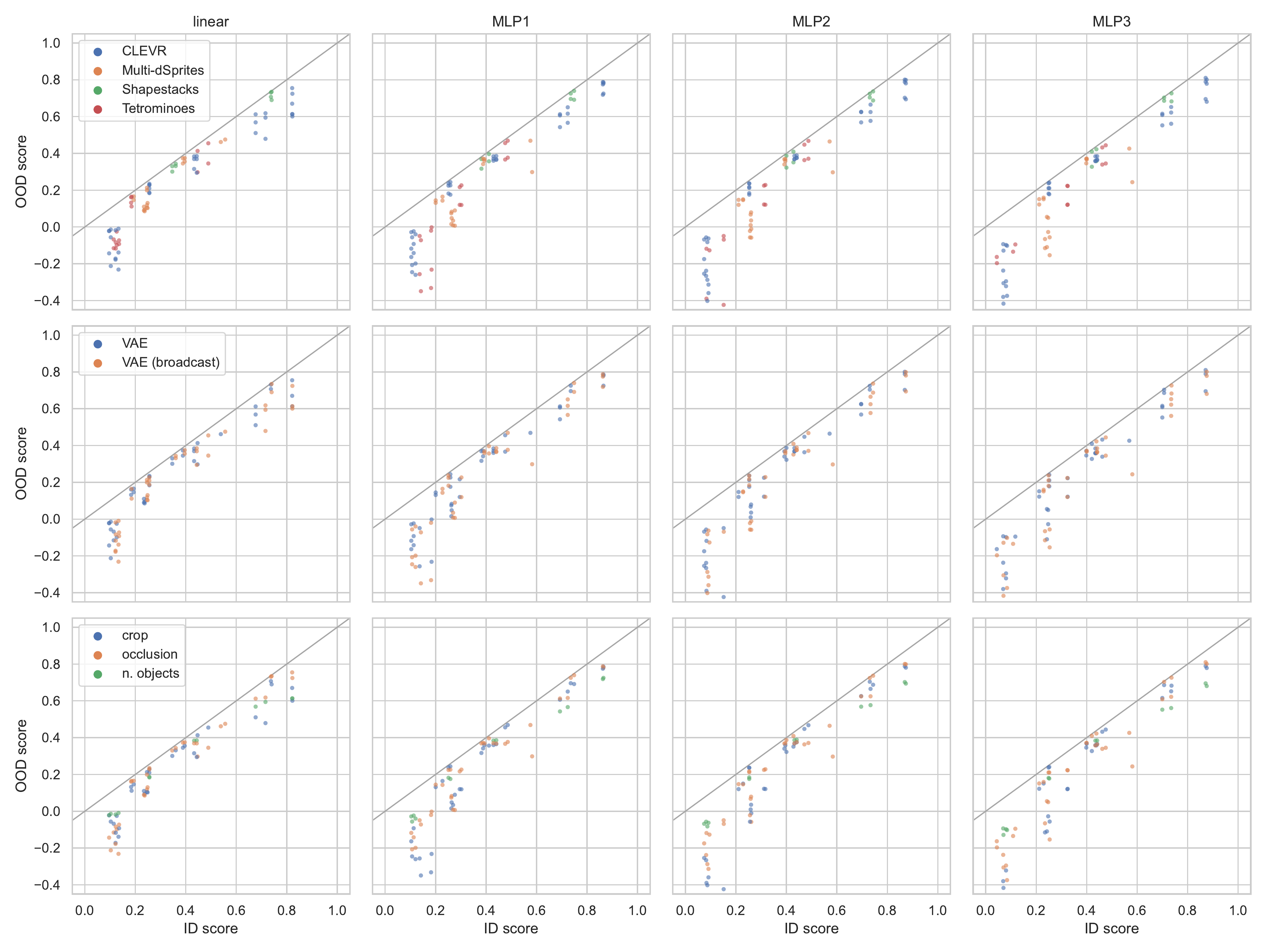}
    \caption{Generalization of \textbf{distributed representations} in downstream prediction, using \textbf{deterministic matching} and \textbf{without retraining} the downstream model after the distribution shift. Here the distribution shift affects \textbf{global properties} of the scene.
        On the x-axis: prediction performance (accuracy or \rsq) for one object property on one dataset, averaged over all objects, on the original training set of the unsupervised object discovery model. On the y-axis: the same metric in OOD scenarios. Each data point corresponds to one representation model (e.g., MONet), one dataset, one object property, and one type of distribution shift. For each x (performance on one object feature in the training distribution, averaged over objects in a scene and over random seeds of the object-centric models) there are multiple y's, corresponding to different distribution shifts.
        In each row, we color-code the data according to dataset, model, or type of distribution shift.
        Each column shows analogous results for each of the 4 considered downstream models for property prediction (linear, and MLPs with up to 3 hidden layers).
    }
    \label{fig:ood/downstream/vae/match_deterministic/global/retrain_False}
\end{figure}

\begin{figure}
    \centering
    \includegraphics[width=\textwidth]{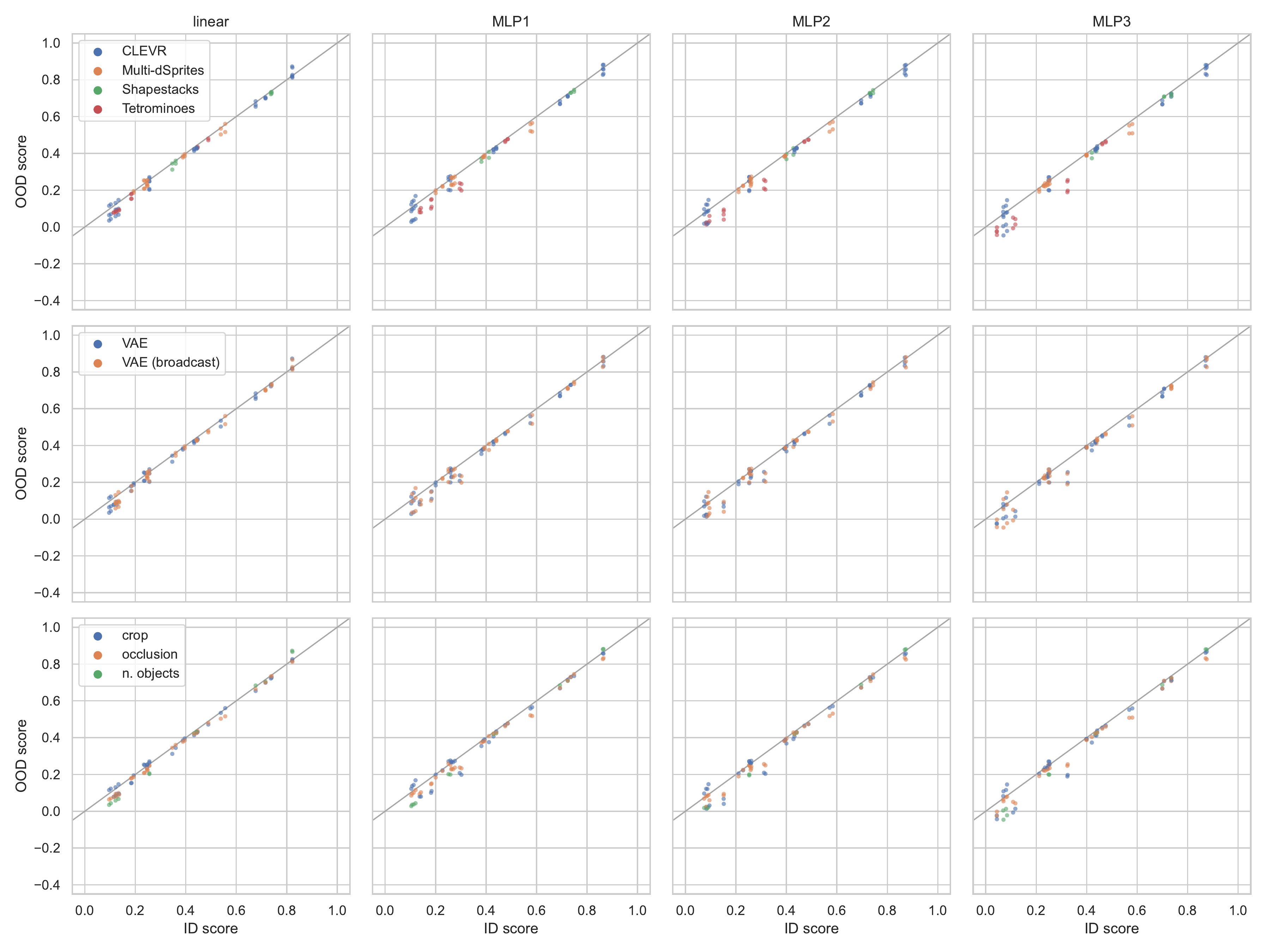}
    \caption{Generalization of \textbf{distributed representations} in downstream prediction, using \textbf{deterministic matching} and \textbf{retraining} the downstream model after the distribution shift. Here the distribution shift affects \textbf{global properties} of the scene.
        On the x-axis: prediction performance (accuracy or \rsq) for one object property on one dataset, averaged over all objects, on the original training set of the unsupervised object discovery model. On the y-axis: the same metric in OOD scenarios. Each data point corresponds to one representation model (e.g., MONet), one dataset, one object property, and one type of distribution shift. For each x (performance on one object feature in the training distribution, averaged over objects in a scene and over random seeds of the object-centric models) there are multiple y's, corresponding to different distribution shifts.
        In each row, we color-code the data according to dataset, model, or type of distribution shift.
        Each column shows analogous results for each of the 4 considered downstream models for property prediction (linear, and MLPs with up to 3 hidden layers).
    }
    \label{fig:ood/downstream/vae/match_deterministic/global/retrain_True}
\end{figure}

In \crefrange{fig:ood/downstream/slotted/match_loss/object/retrain_False}{fig:ood/downstream/vae/match_deterministic/global/retrain_True}, we show the relationship between ID and OOD downstream prediction performance for the same model, dataset, downstream predictor, and object property. Assume a pretrained unsupervised object discovery model is given, and a downstream model is trained from said model's representations to predict object properties. These plots answer the following question: given that the downstream model predicts a particular object property (e.g., size in CLEVR) with a certain accuracy (on average over all objects in all test images), how well is it going to predict the same property when the scene undergoes one of the possible distribution shifts considered in this study? And in case the distribution shift only affects one object, how well is it going to predict that property in the ID objects as opposed to the OOD objects?
These 16 figures show all combinations of the following 4 factors (hierarchically in this order): object-centric/distributed representations; loss/mask matching for object-centric representations or loss/deterministic for distributed representations; without/with retraining of the downstream model after the distribution shift has occurred; single-object/global distribution shifts.
In each figure, we show results for each of the 4 downstream models considered in this study (linear, and MLP with up to 3 hidden layers). For each of these, we show splits in terms of ID/OOD objects (when applicable), dataset, upstream model, type of distribution shift.





\clearpage
\subsection{Qualitative results}\label{app:results_qualitative}

\begin{figure}
    \centering
    \includegraphics[height=4cm]{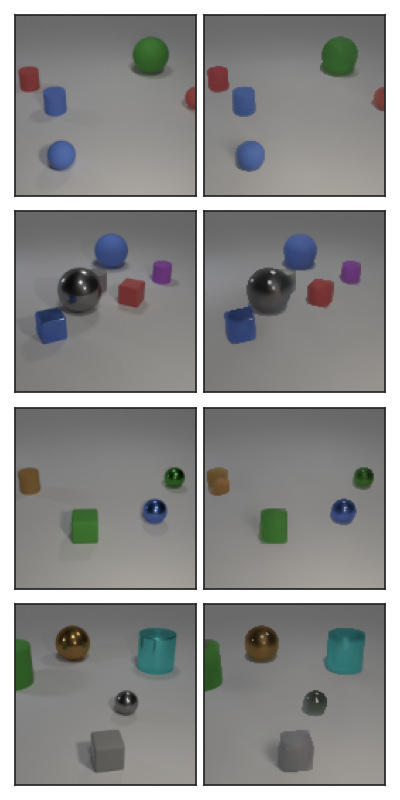}
    \hspace{0.2cm}
    \includegraphics[height=4cm]{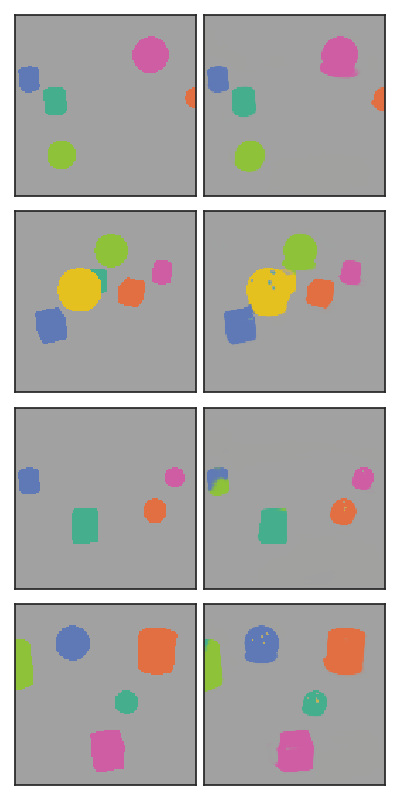}
    \hspace{0.2cm}
    \includegraphics[height=4cm]{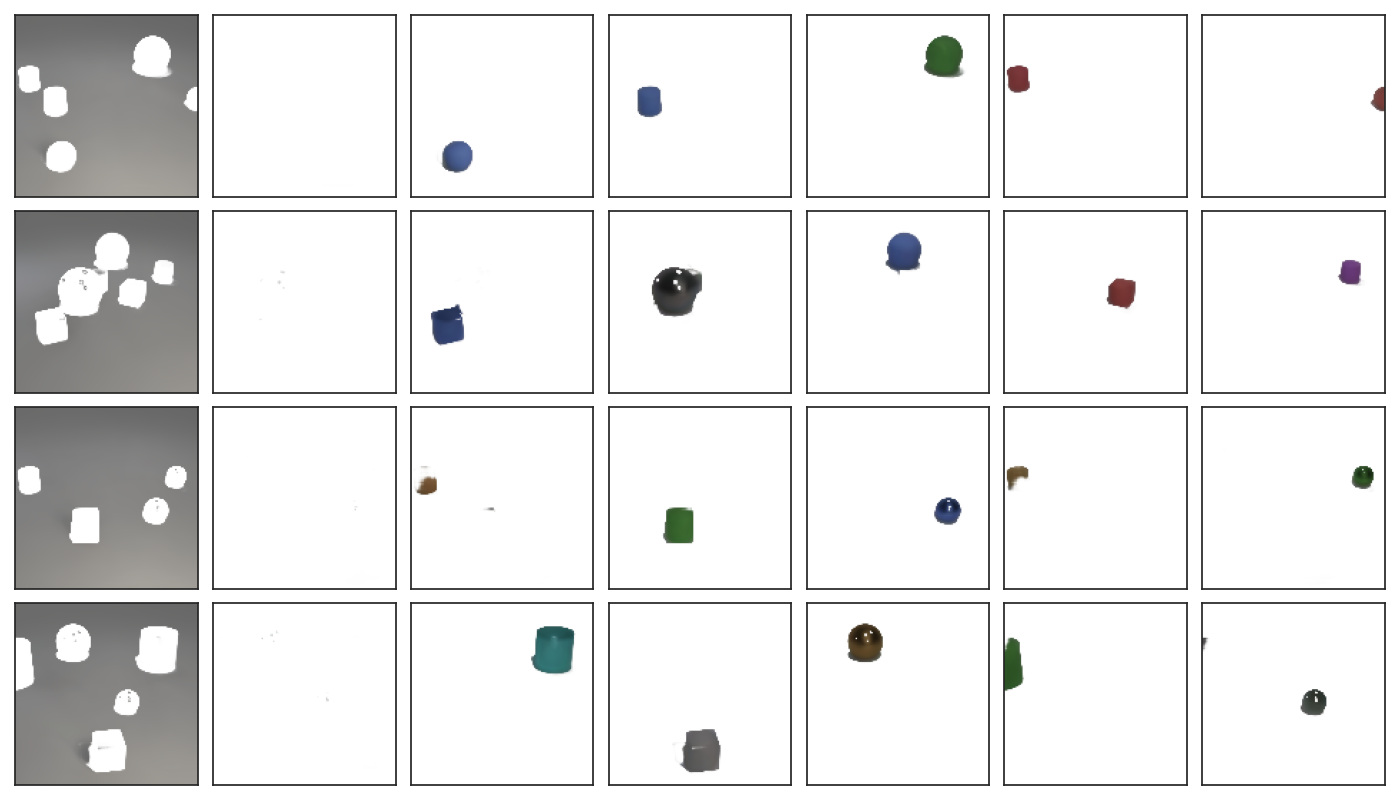}
    
    \vspace{0.45cm}
    
    \includegraphics[height=4cm]{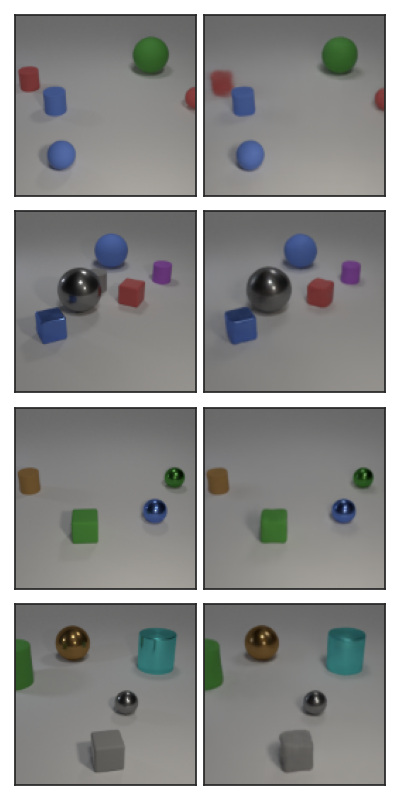}
    \hspace{0.2cm}
    \includegraphics[height=4cm]{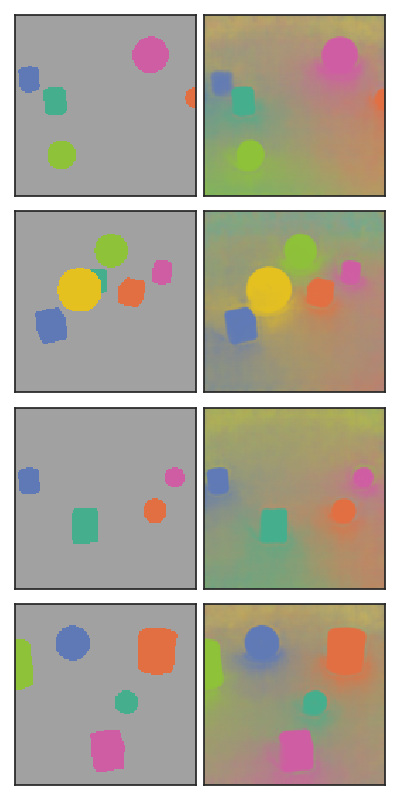}
    \hspace{0.2cm}
    \includegraphics[height=4cm]{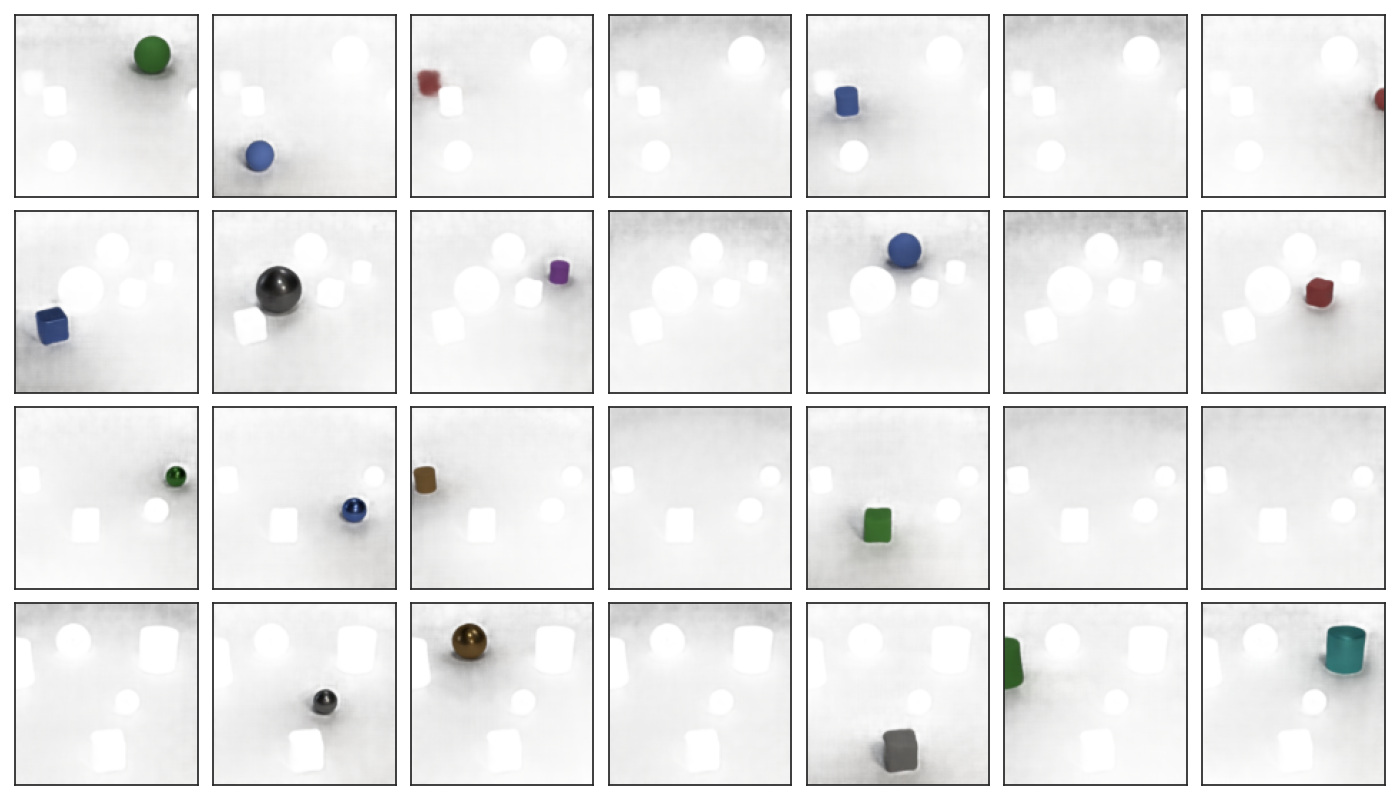}
    
    \vspace{0.45cm}
    
    \includegraphics[height=4cm]{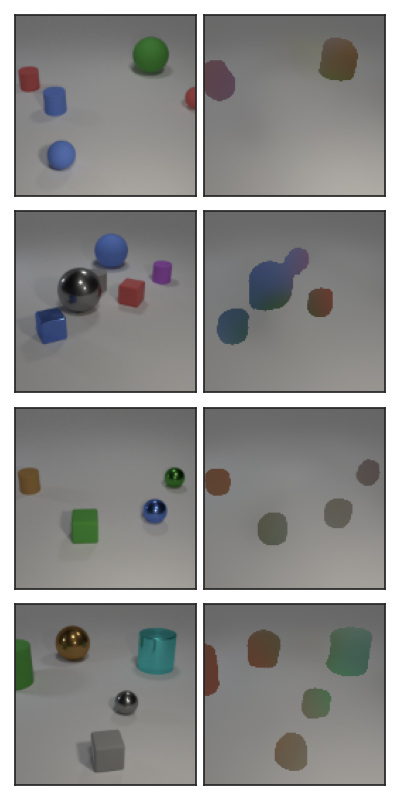}
    \hspace{0.2cm}
    \includegraphics[height=4cm]{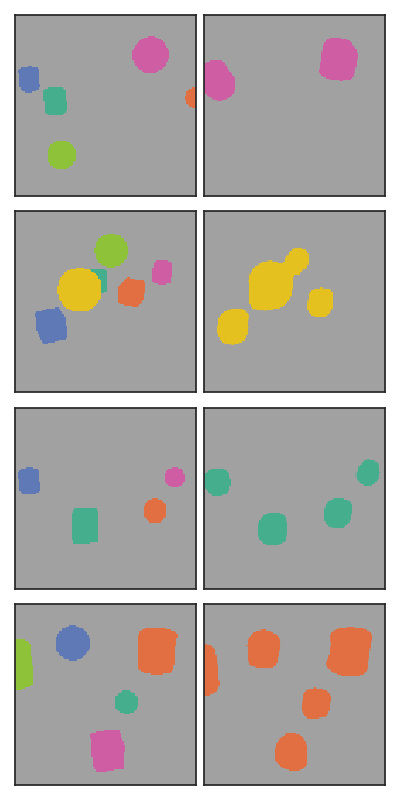}
    \hspace{0.2cm}
    \includegraphics[height=4cm]{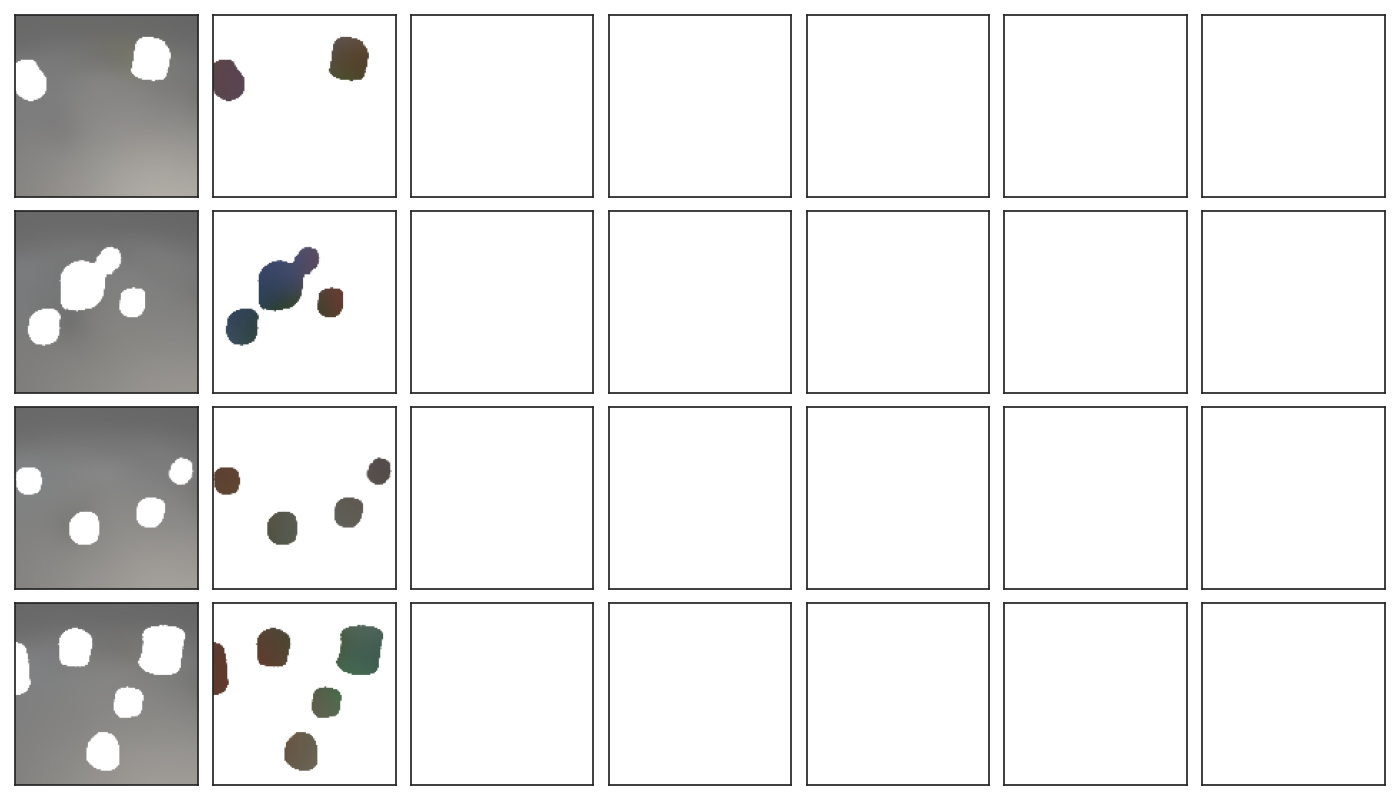}
    
    \vspace{0.45cm}
    
    \includegraphics[height=4cm]{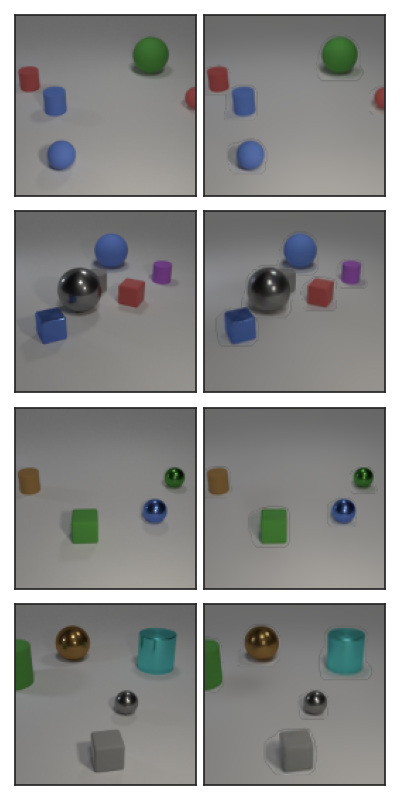}
    \hspace{0.2cm}
    \includegraphics[height=4cm]{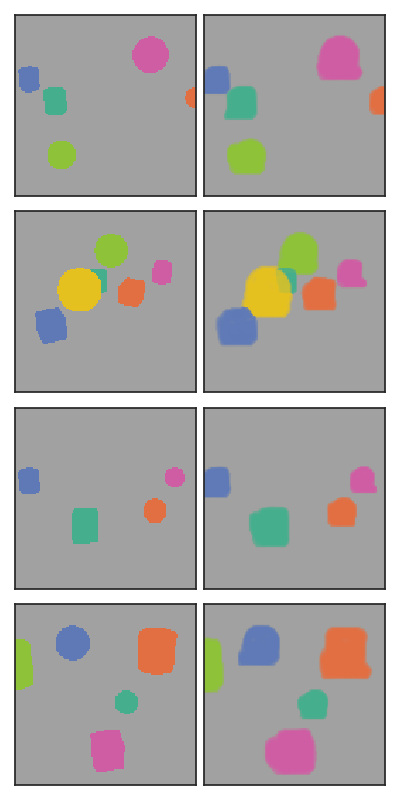}
    \hspace{0.2cm}
    \includegraphics[height=4cm]{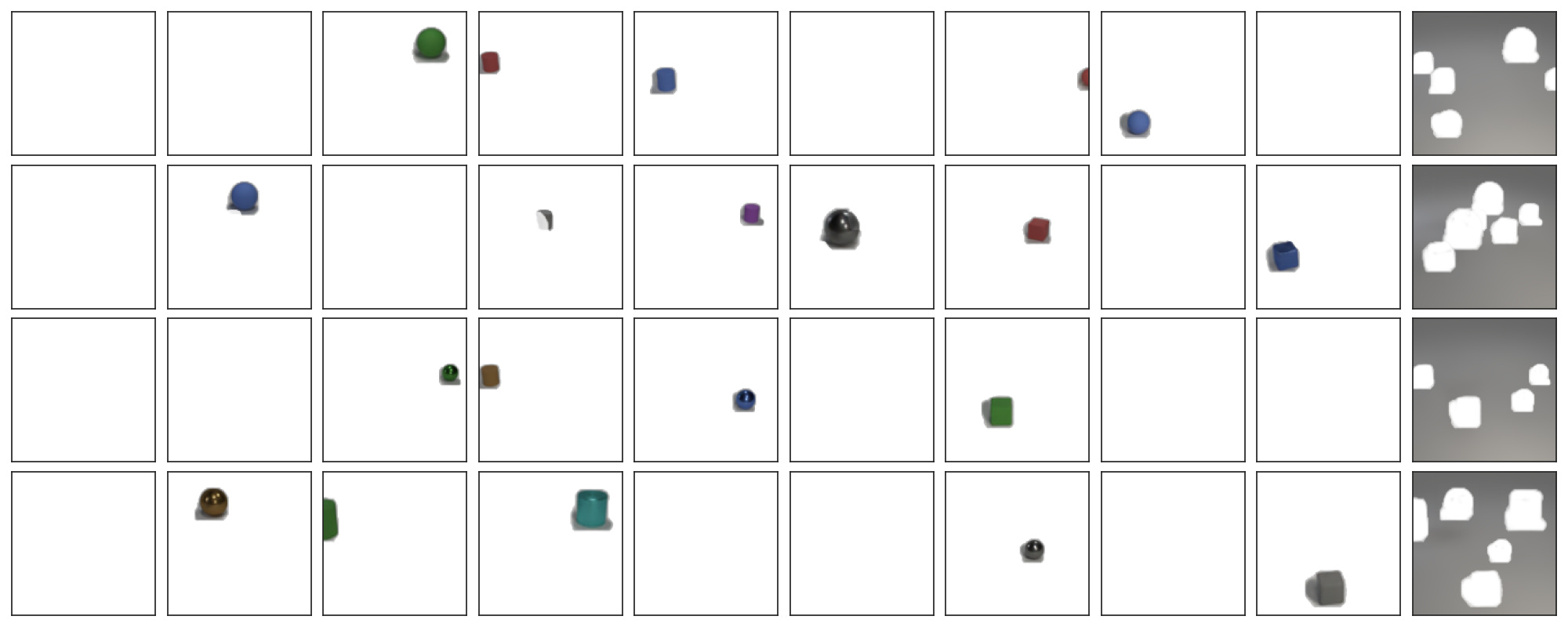}
    
    \vspace{0.45cm}
    \caption{\textbf{Reconstruction and segmentation} of 4 random images from the held-out test set of \textbf{CLEVR6}. Top to bottom: MONet, Slot Attention, GENESIS, SPACE. Left to right: input, reconstruction, ground-truth masks, predicted (soft) masks, slot-wise reconstructions (masked with the predicted masks). As explained in the text, for SPACE we select the 10 most salient slots using the predicted masks. As explained in the text, for SPACE we select the 10 most salient slots using the predicted masks. For each model type, we visualize the specific model with the highest ARI score in the \textit{validation} set. The images shown here are from the \textit{test} set and were not used for model selection.}
    \label{fig:app_model_viz_clevr}
\end{figure}

\begin{figure}
    \centering
    \includegraphics[height=4cm]{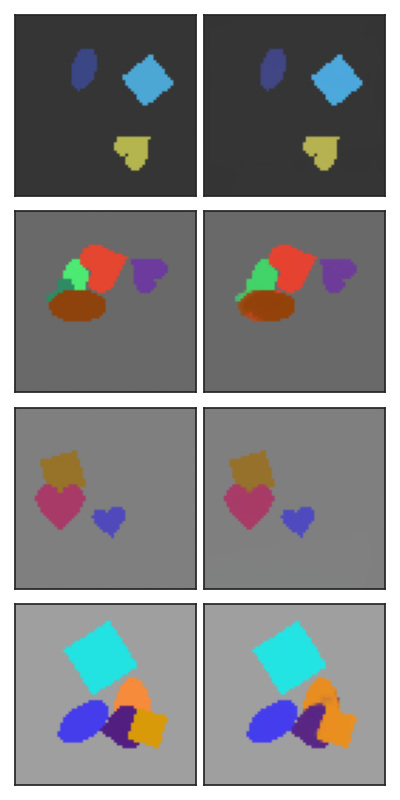}
    \hspace{0.2cm}
    \includegraphics[height=4cm]{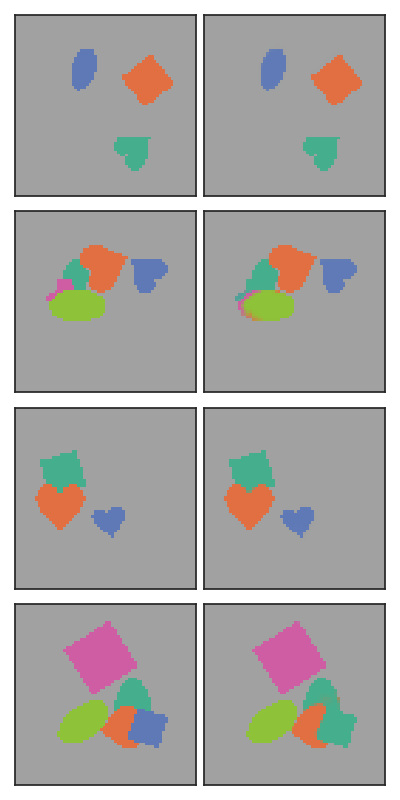}
    \hspace{0.2cm}
    \includegraphics[height=4cm]{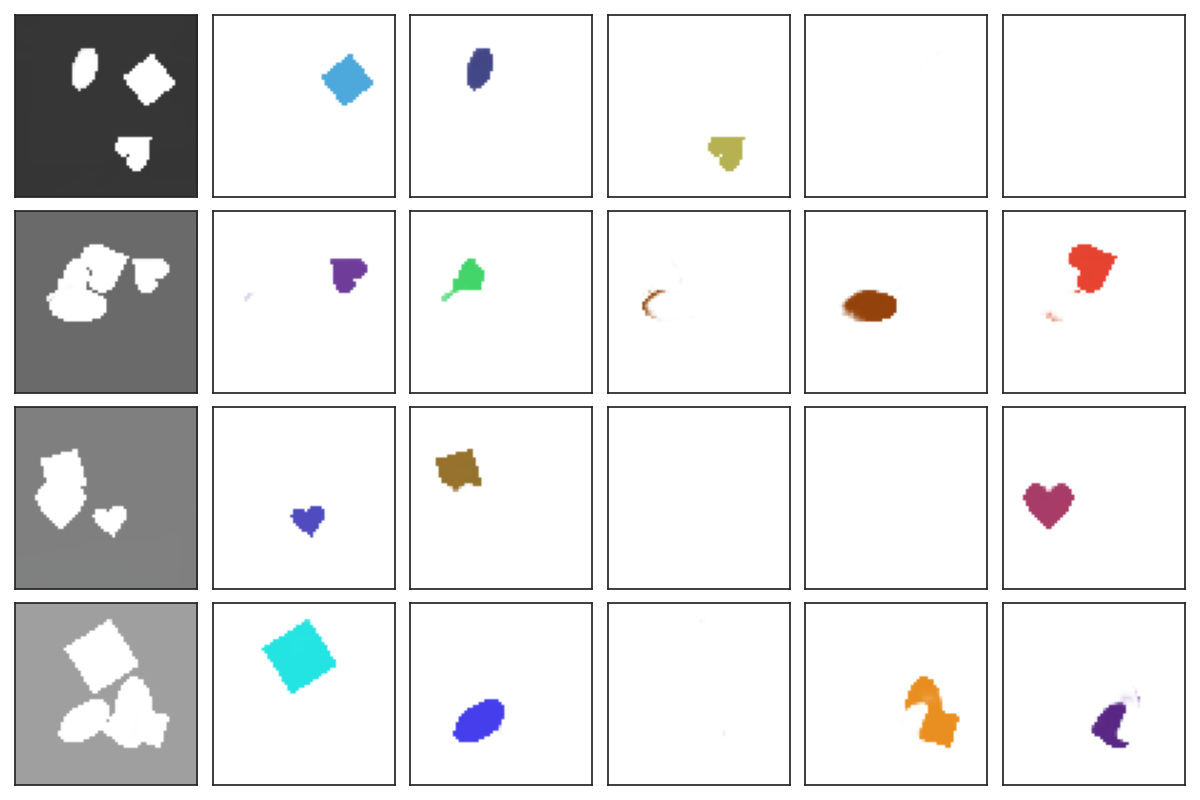}
    
    \vspace{0.45cm}
    
    \includegraphics[height=4cm]{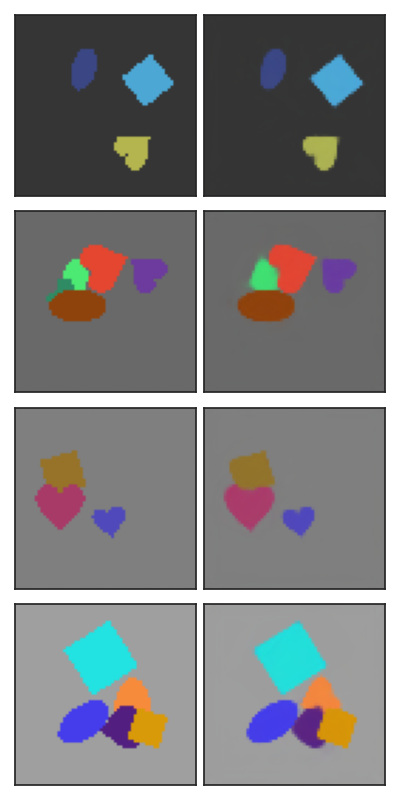}
    \hspace{0.2cm}
    \includegraphics[height=4cm]{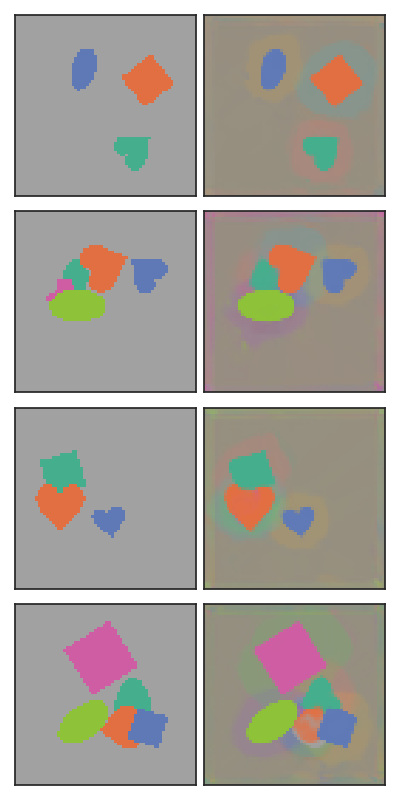}
    \hspace{0.2cm}
    \includegraphics[height=4cm]{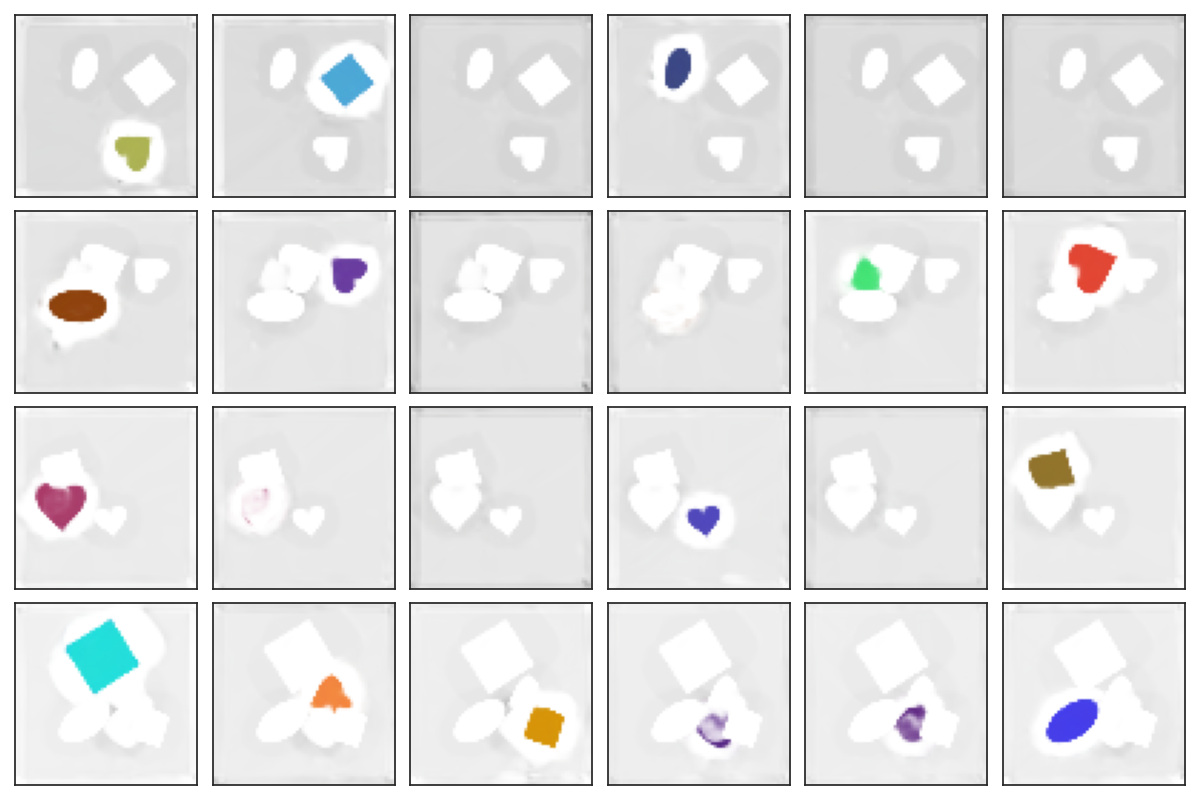}
    
    \vspace{0.45cm}
    
    \includegraphics[height=4cm]{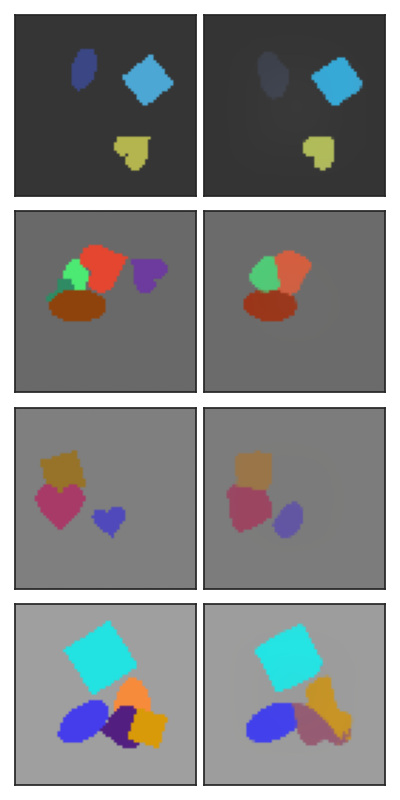}
    \hspace{0.2cm}
    \includegraphics[height=4cm]{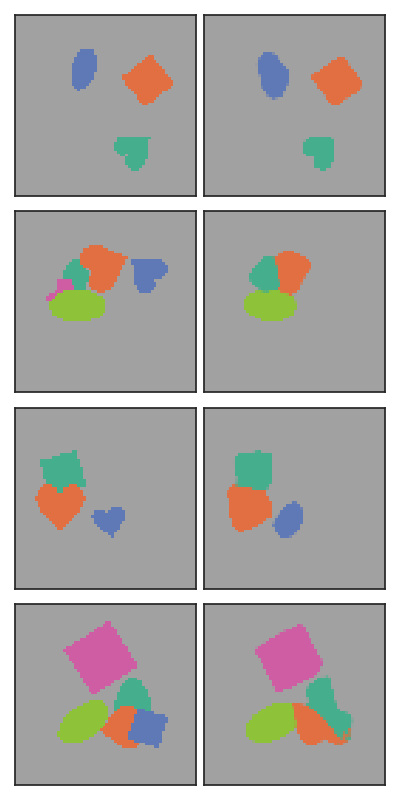}
    \hspace{0.2cm}
    \includegraphics[height=4cm]{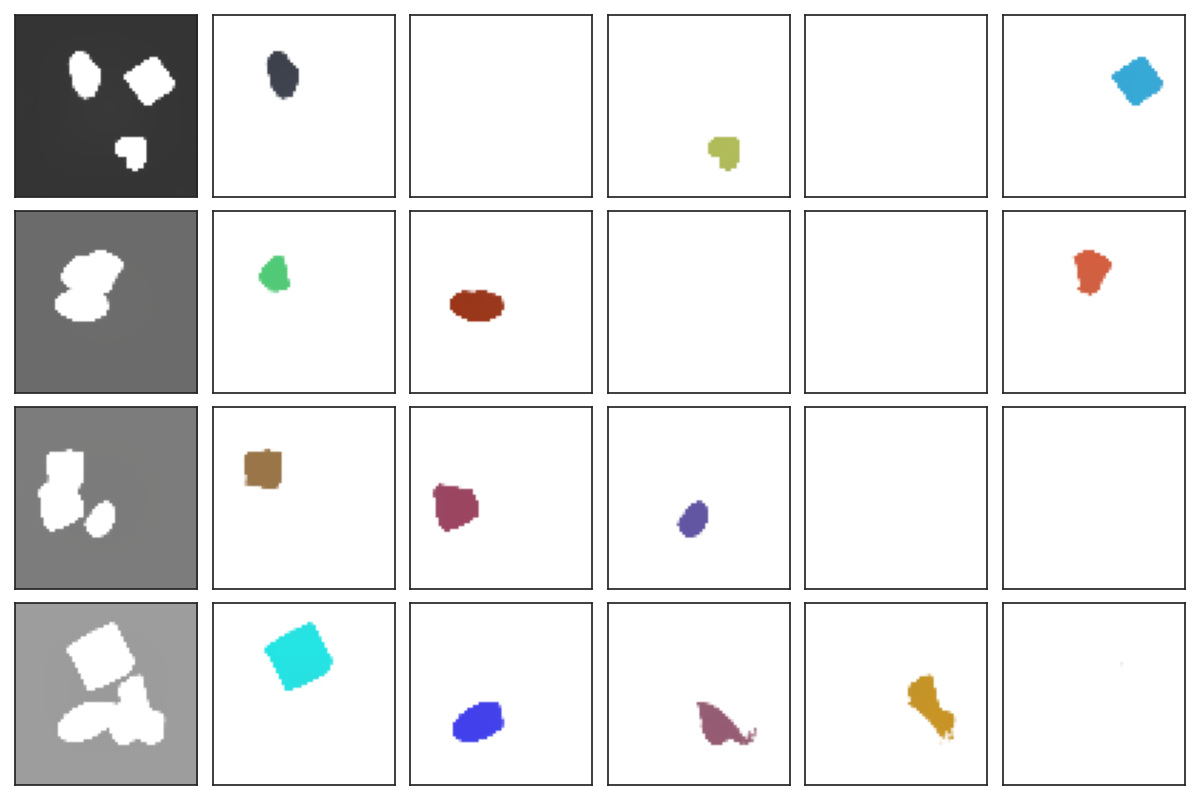}
    
    \vspace{0.45cm}
    
    \includegraphics[height=4cm]{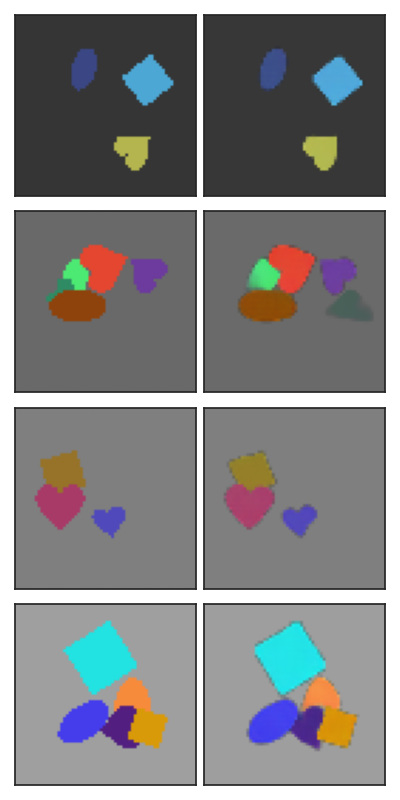}
    \hspace{0.2cm}
    \includegraphics[height=4cm]{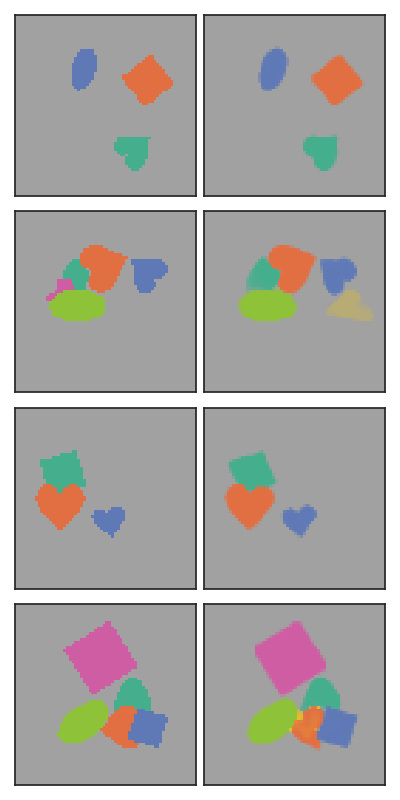}
    \hspace{0.2cm}
    \includegraphics[height=4cm]{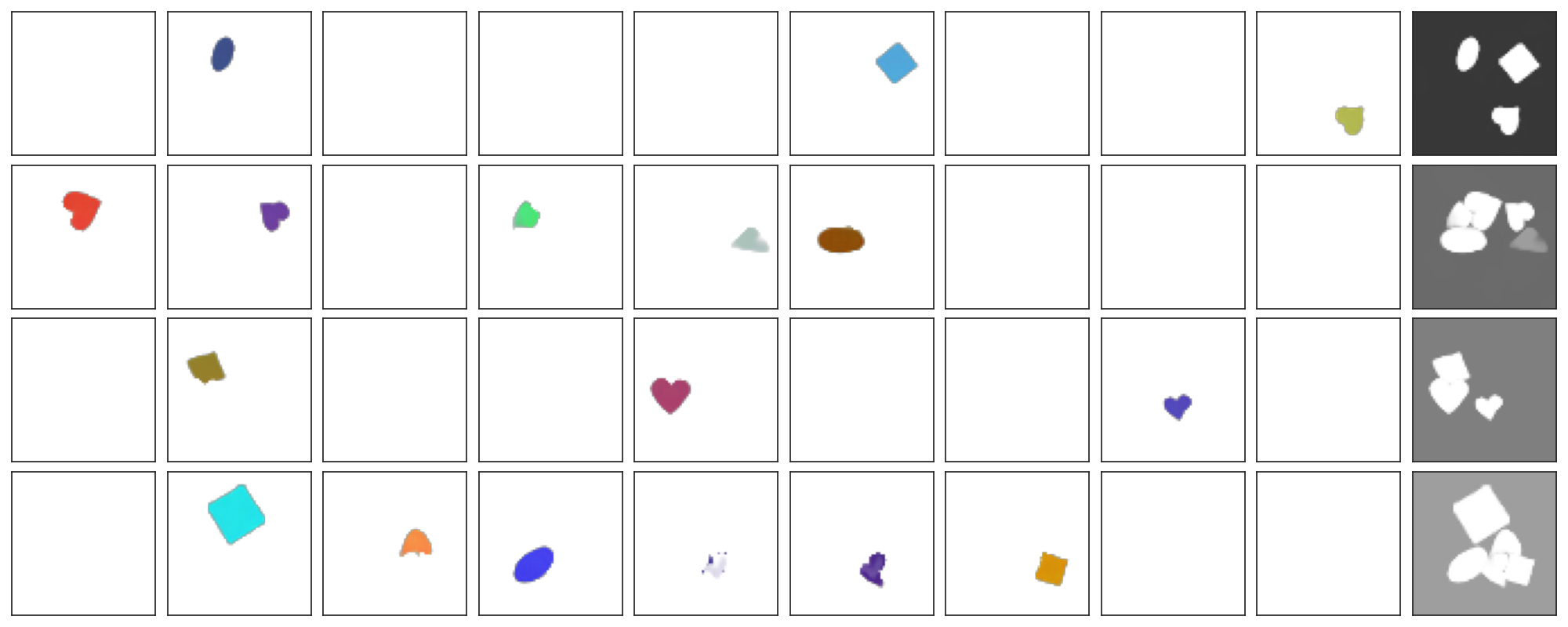}
    
    \vspace{0.45cm}
    \caption{\textbf{Reconstruction and segmentation} of 4 random images from the held-out test set of \textbf{Multi-dSprites}. Top to bottom: MONet, Slot Attention, GENESIS, SPACE. Left to right: input, reconstruction, ground-truth masks, predicted (soft) masks, slot-wise reconstructions (masked with the predicted masks). As explained in the text, for SPACE we select the 10 most salient slots using the predicted masks. For each model type, we visualize the specific model with the highest ARI score in the \textit{validation} set. The images shown here are from the \textit{test} set and were not used for model selection.}
    \label{fig:app_model_viz_multidsprites}
\end{figure}

\begin{figure}
    \centering
    \includegraphics[height=4cm]{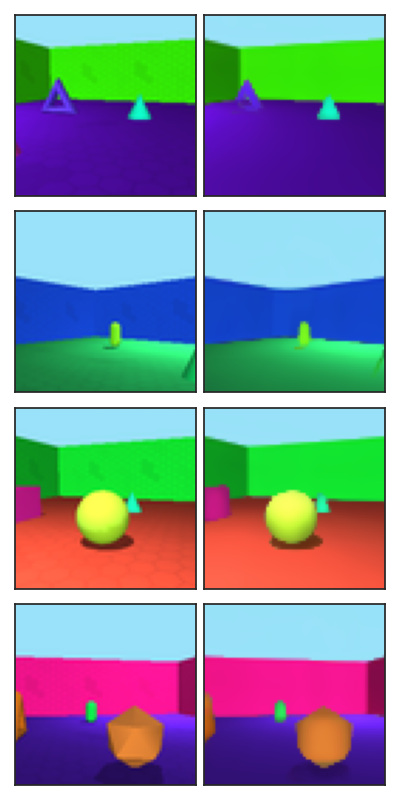}
    \hspace{0.2cm}
    \includegraphics[height=4cm]{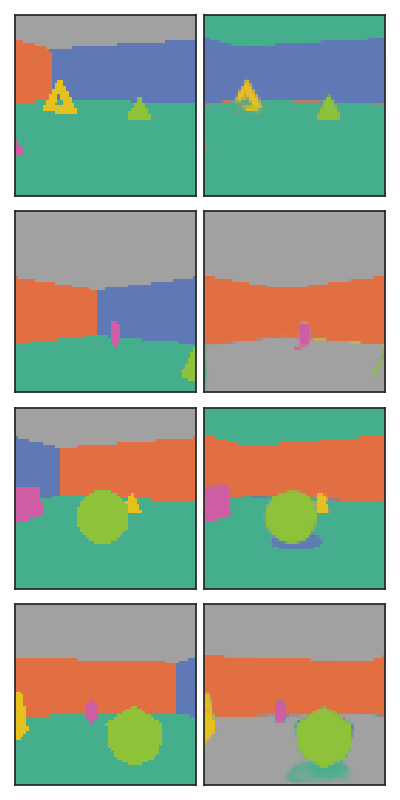}
    \hspace{0.2cm}
    \includegraphics[height=4cm]{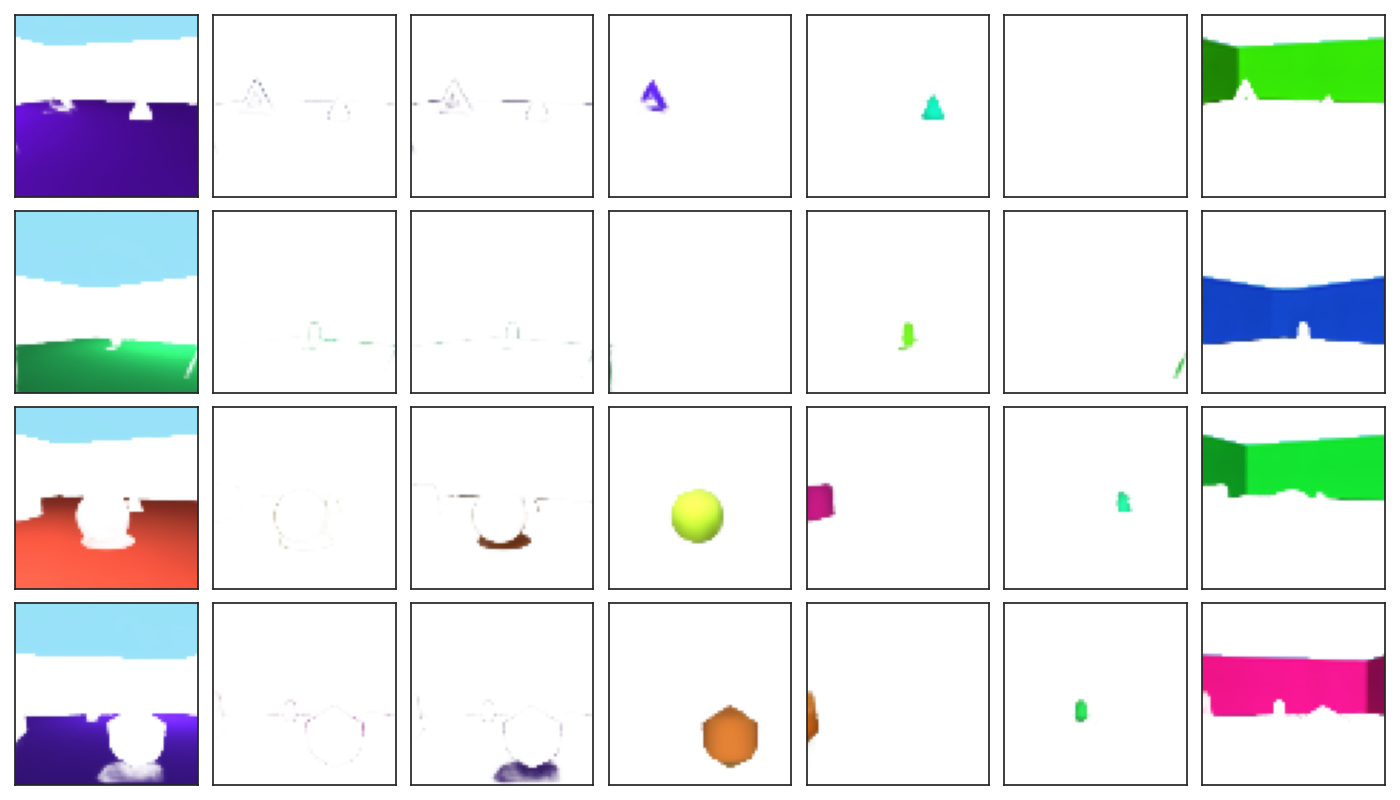}
    
    \vspace{0.45cm}
    
    \includegraphics[height=4cm]{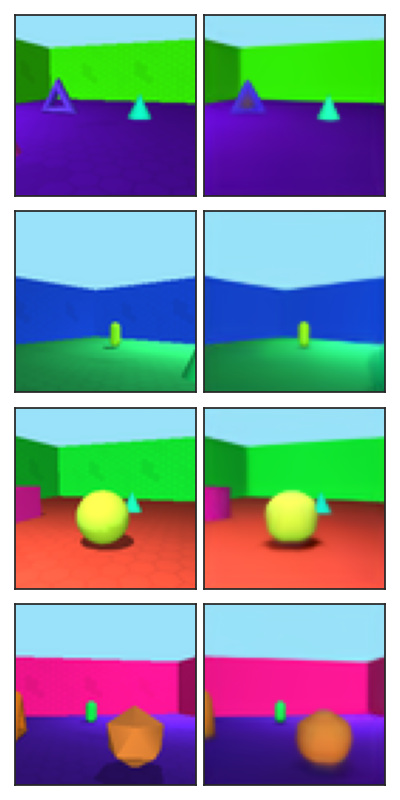}
    \hspace{0.2cm}
    \includegraphics[height=4cm]{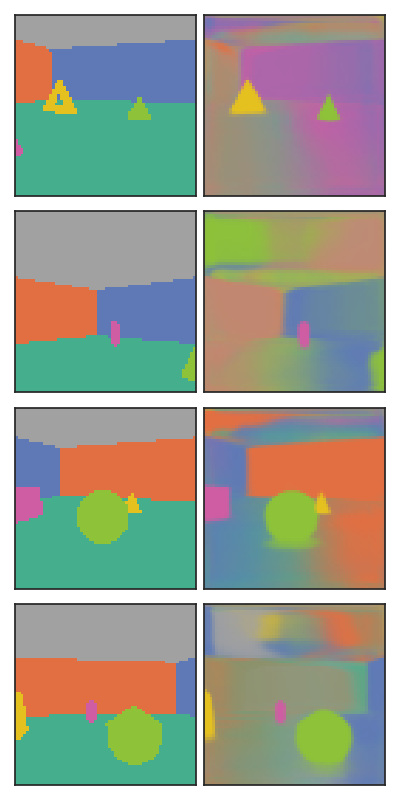}
    \hspace{0.2cm}
    \includegraphics[height=4cm]{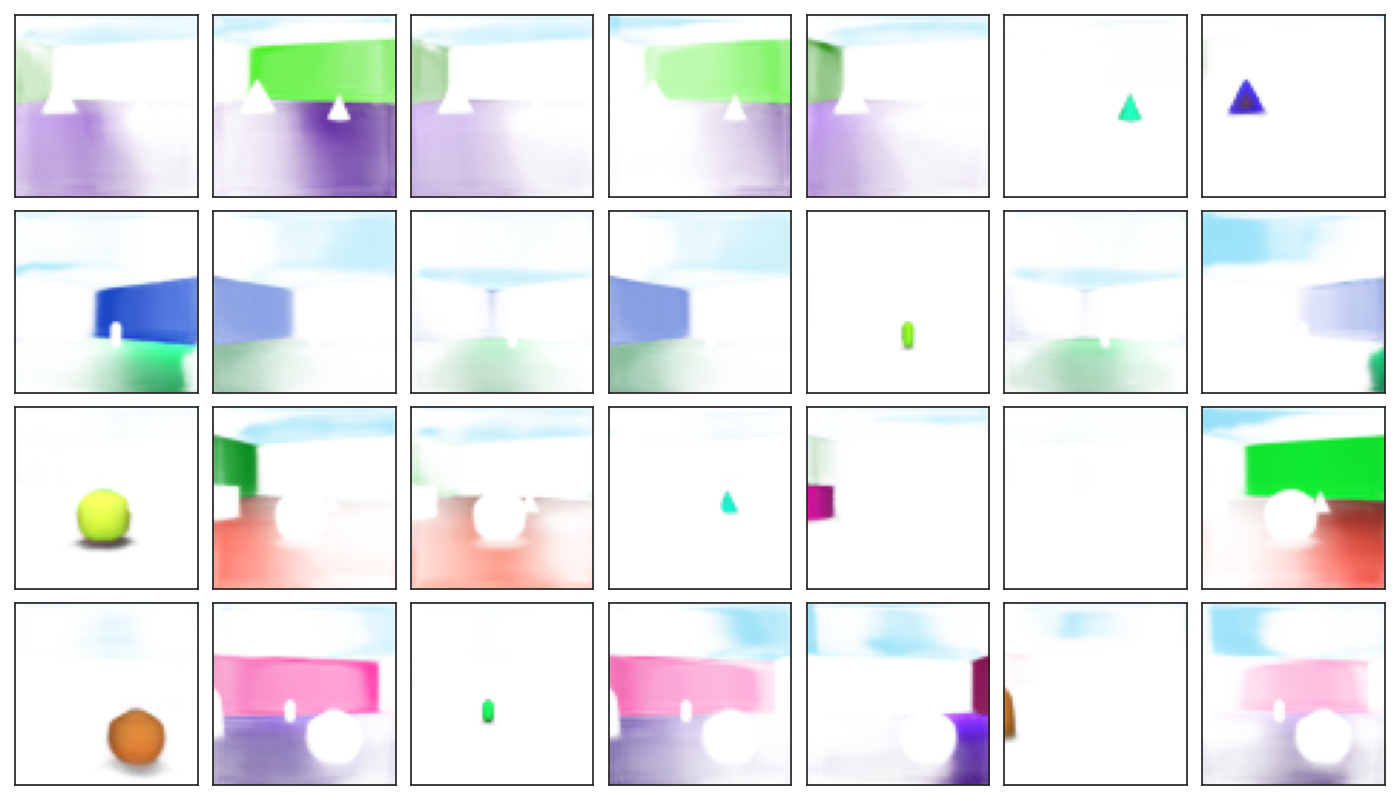}
    
    \vspace{0.45cm}
    
    \includegraphics[height=4cm]{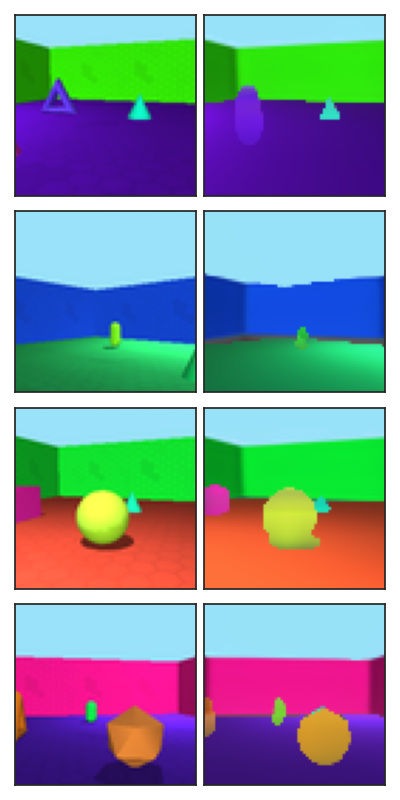}
    \hspace{0.2cm}
    \includegraphics[height=4cm]{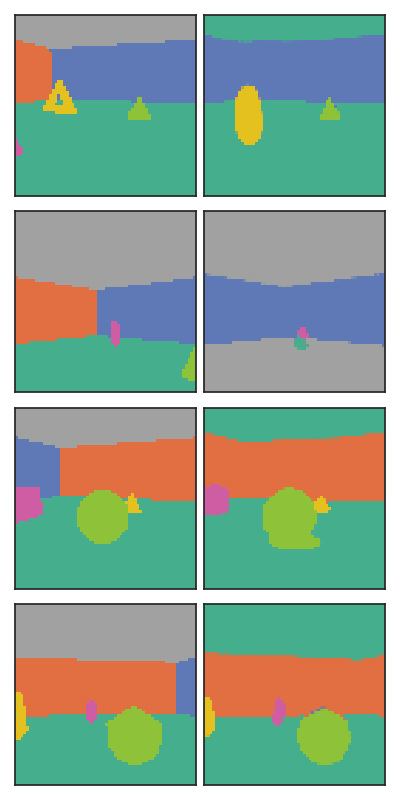}
    \hspace{0.2cm}
    \includegraphics[height=4cm]{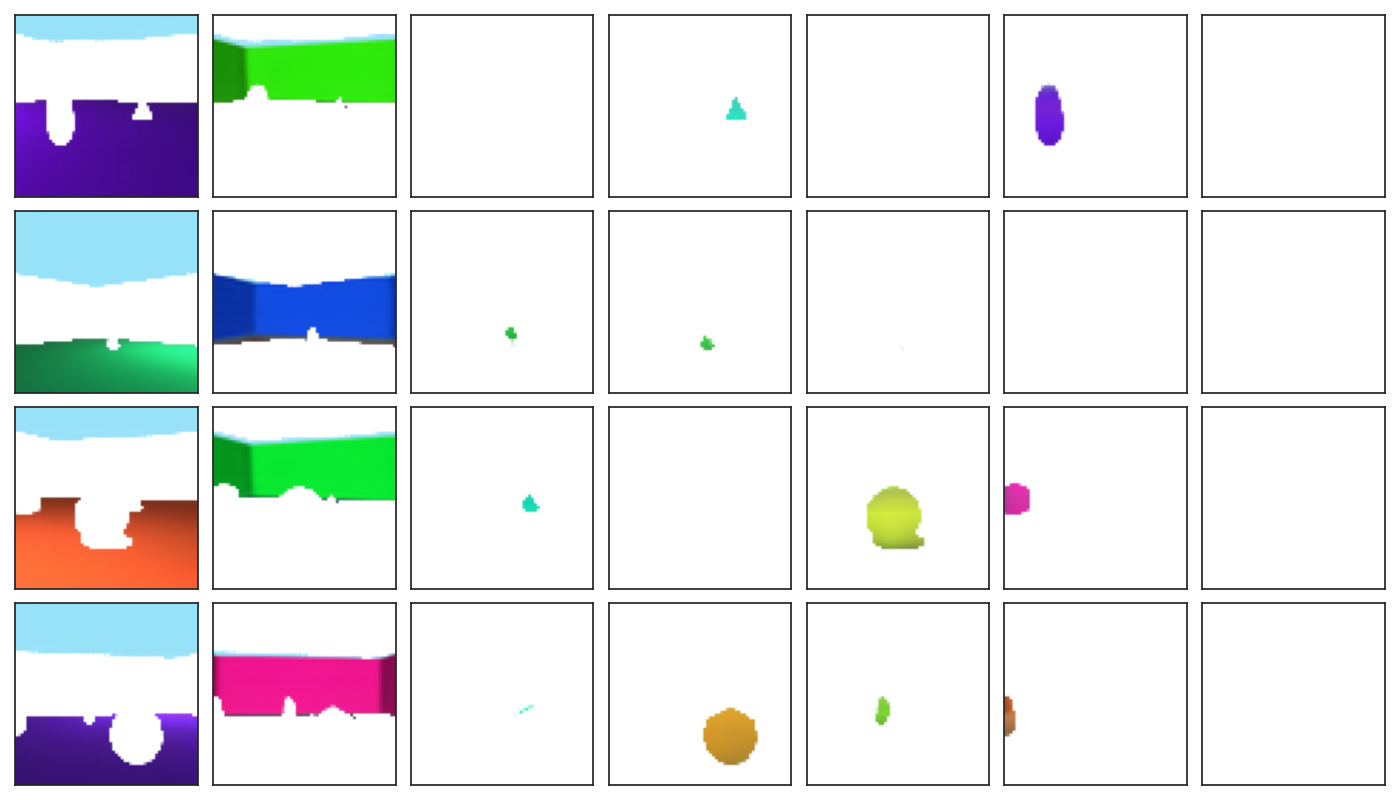}
    
    \vspace{0.45cm}
    
    \includegraphics[height=4cm]{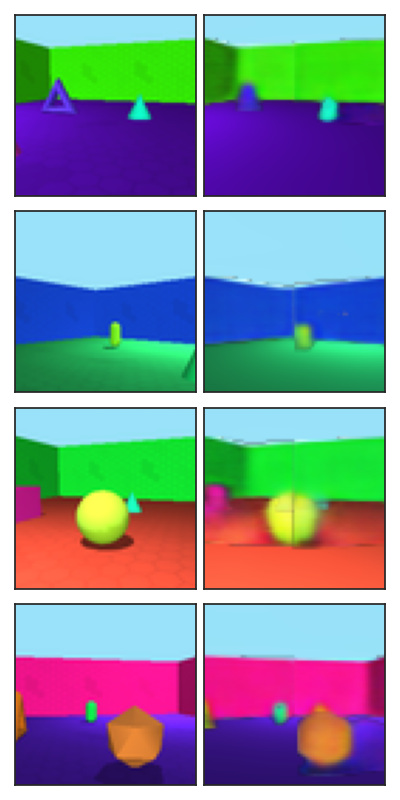}
    \hspace{0.2cm}
    \includegraphics[height=4cm]{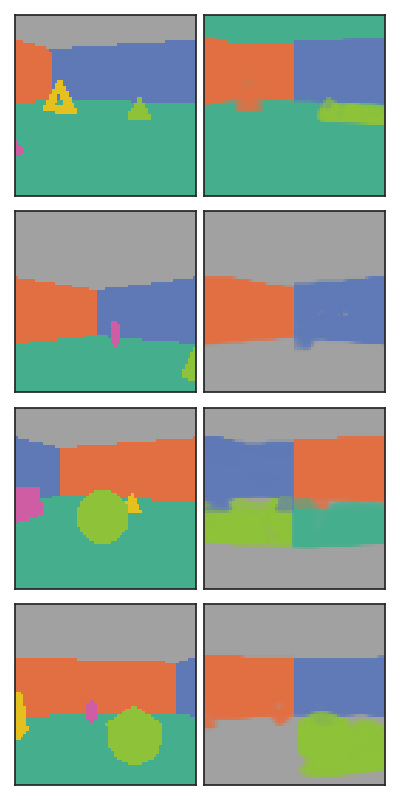}
    \hspace{0.2cm}
    \includegraphics[height=4cm]{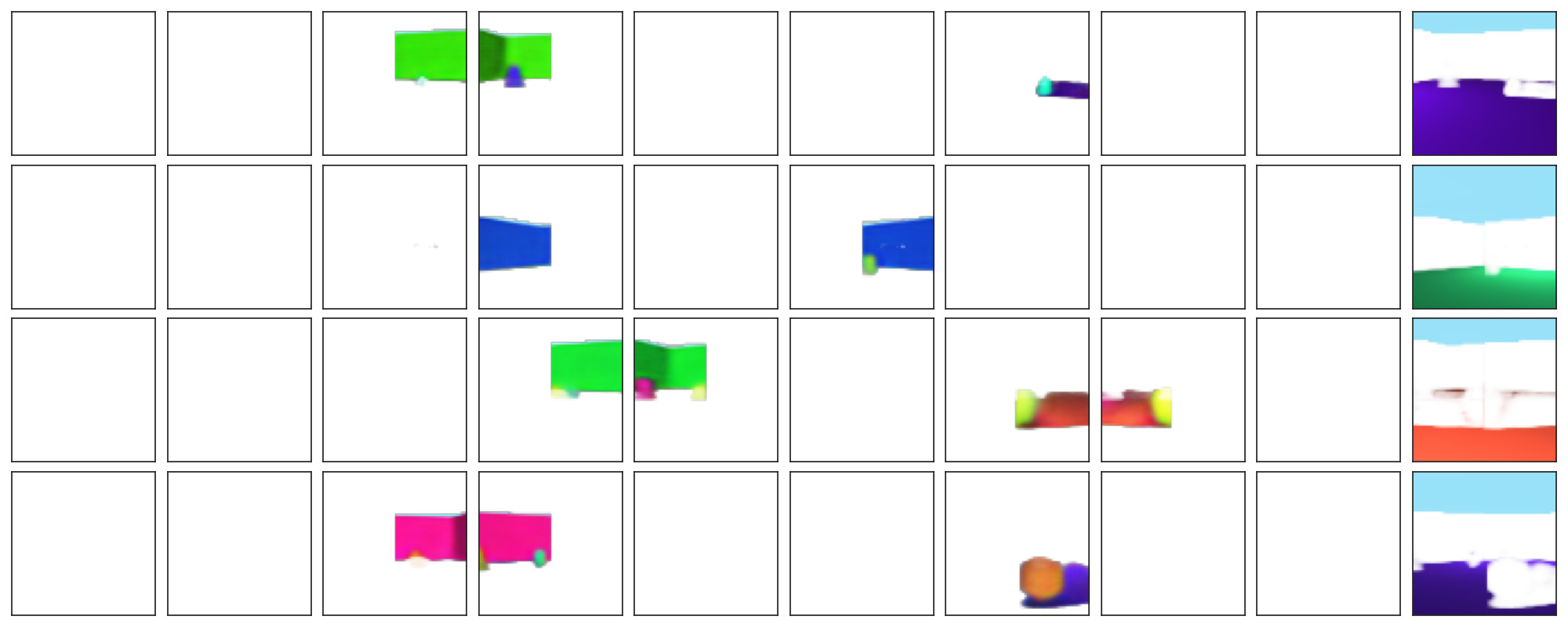}
    
    \vspace{0.45cm}
    \caption{\textbf{Reconstruction and segmentation} of 4 random images from the held-out test set of \textbf{Objects Room}. Top to bottom: MONet, Slot Attention, GENESIS, SPACE. Left to right: input, reconstruction, ground-truth masks, predicted (soft) masks, slot-wise reconstructions (masked with the predicted masks). As explained in the text, for SPACE we select the 10 most salient slots using the predicted masks. For each model type, we visualize the specific model with the highest ARI score in the \textit{validation} set. The images shown here are from the \textit{test} set and were not used for model selection.}
    \label{fig:app_model_viz_objectsroom}
\end{figure}

\begin{figure}
    \centering
    \includegraphics[height=4cm]{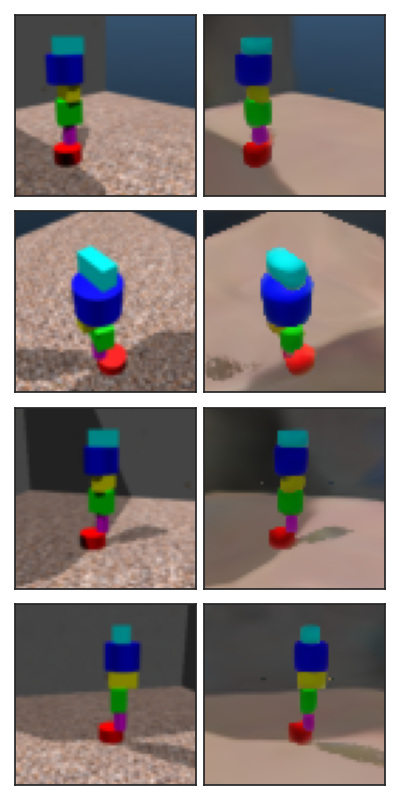}
    \hspace{0.2cm}
    \includegraphics[height=4cm]{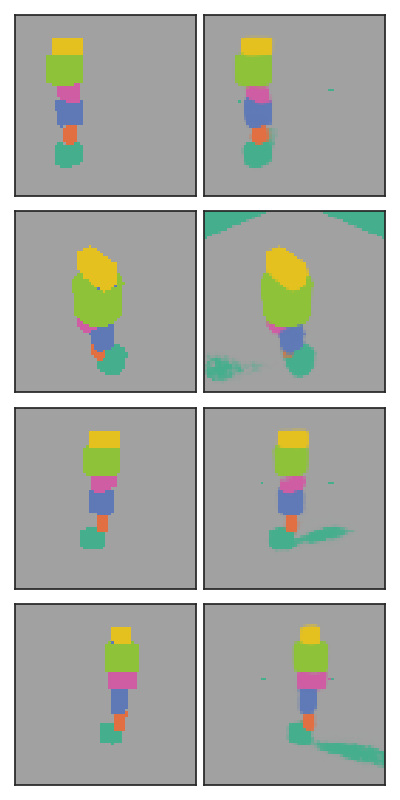}
    \hspace{0.2cm}
    \includegraphics[height=4cm]{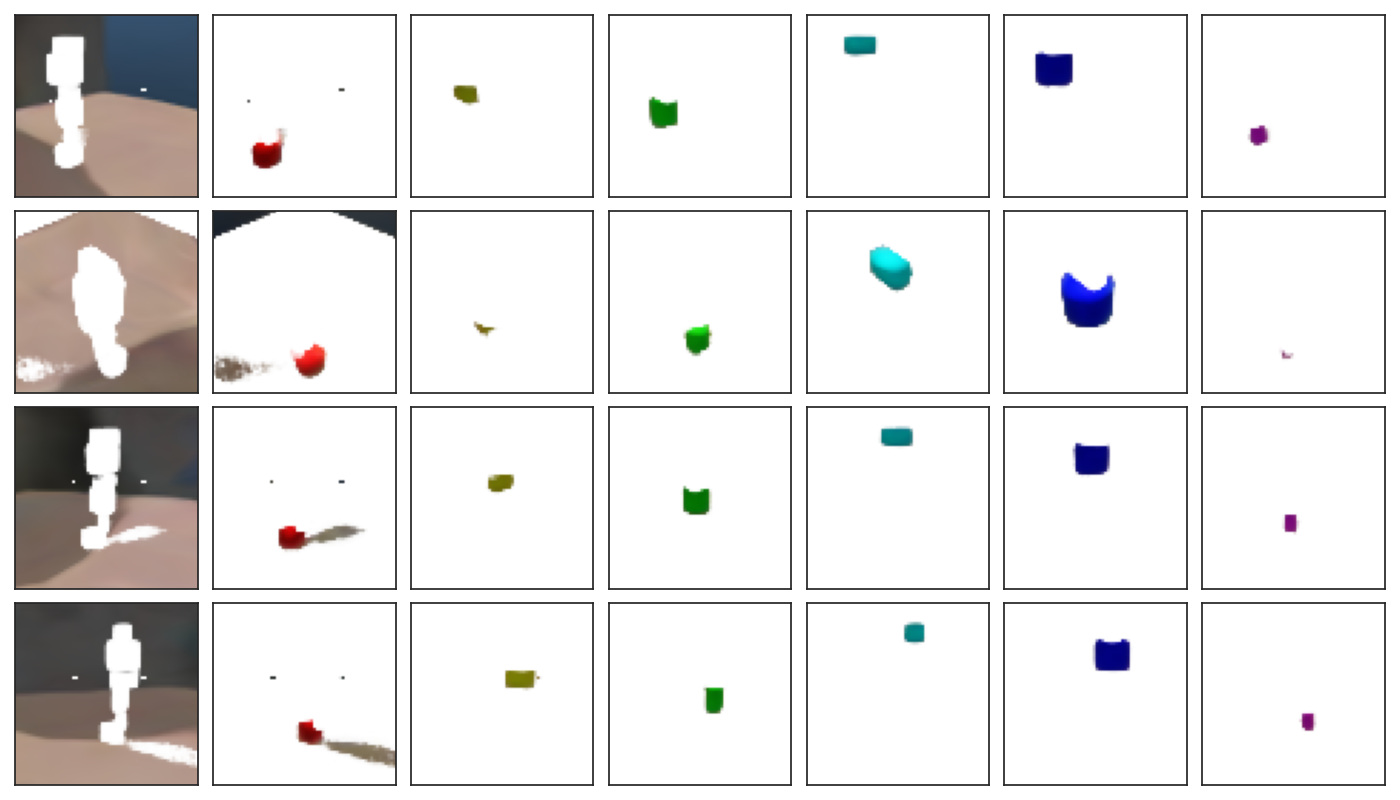}
    
    \vspace{0.45cm}
    
    \includegraphics[height=4cm]{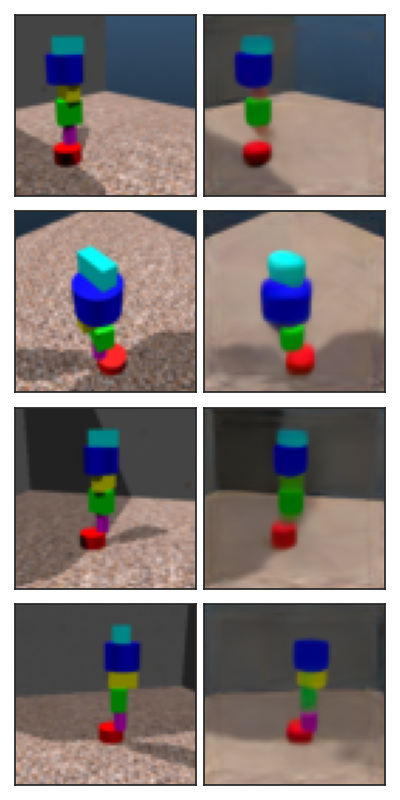}
    \hspace{0.2cm}
    \includegraphics[height=4cm]{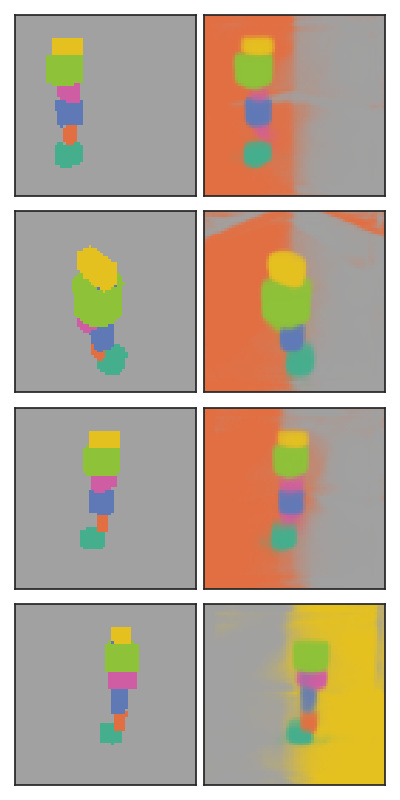}
    \hspace{0.2cm}
    \includegraphics[height=4cm]{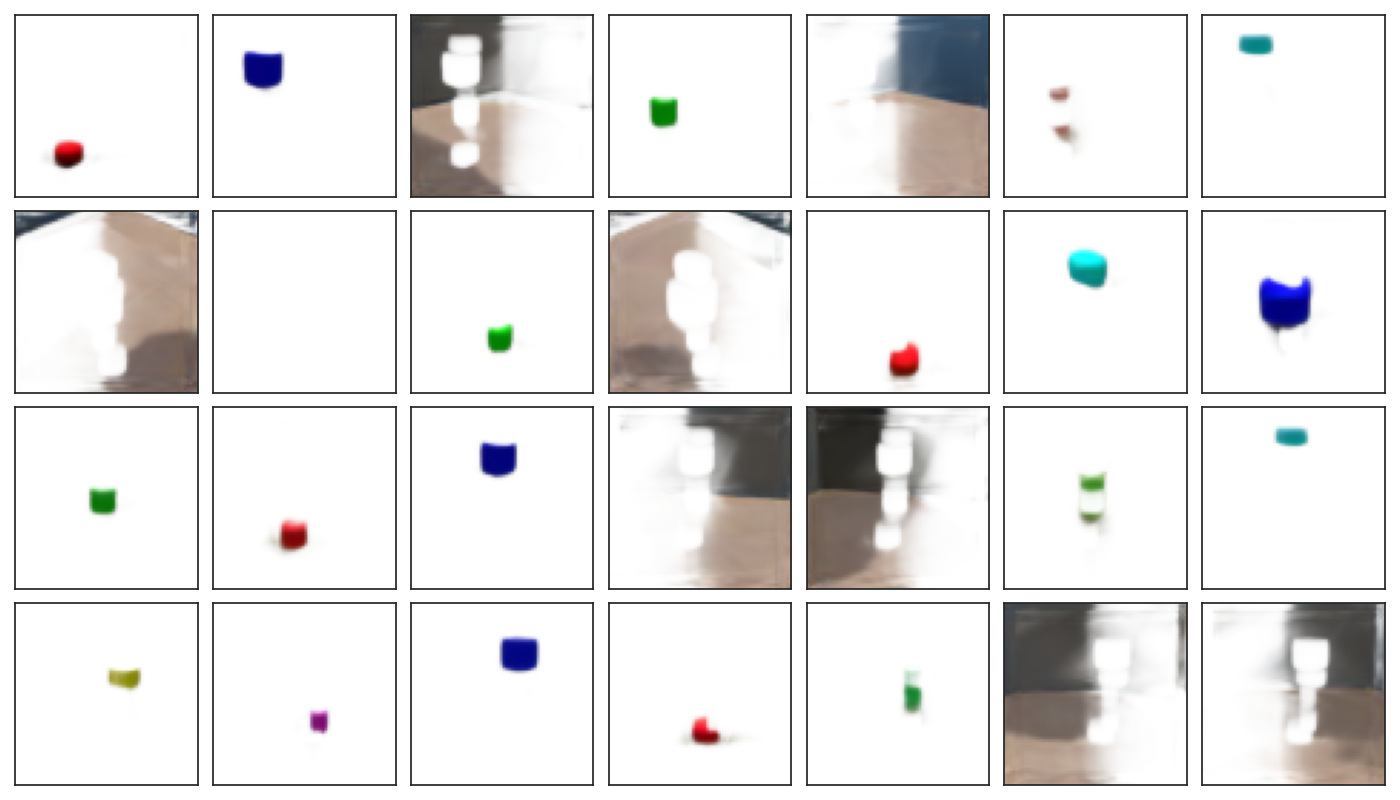}
    
    \vspace{0.45cm}
    
    \includegraphics[height=4cm]{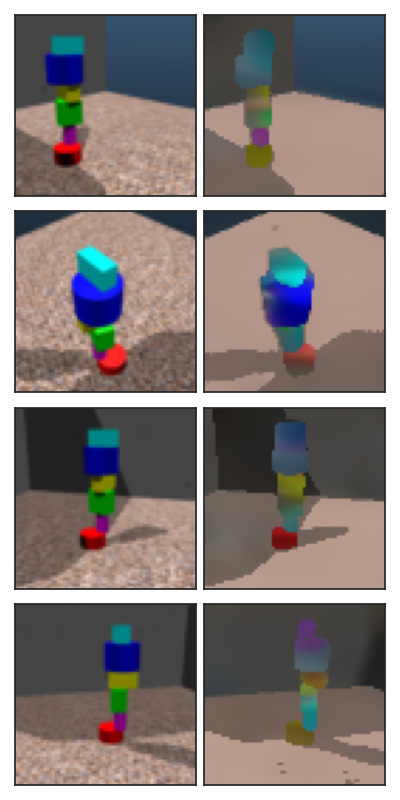}
    \hspace{0.2cm}
    \includegraphics[height=4cm]{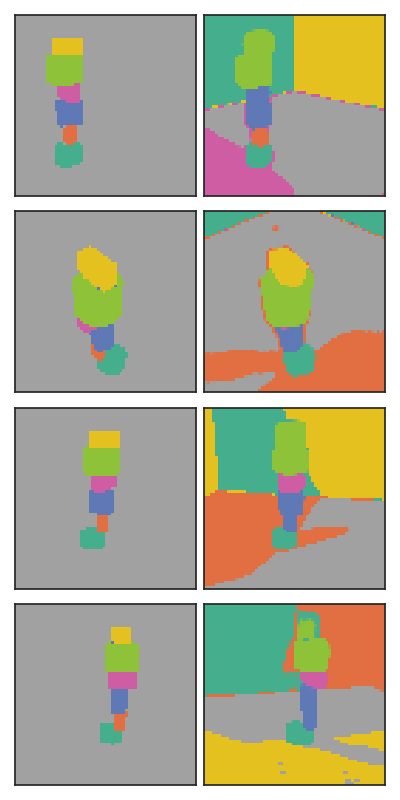}
    \hspace{0.2cm}
    \includegraphics[height=4cm]{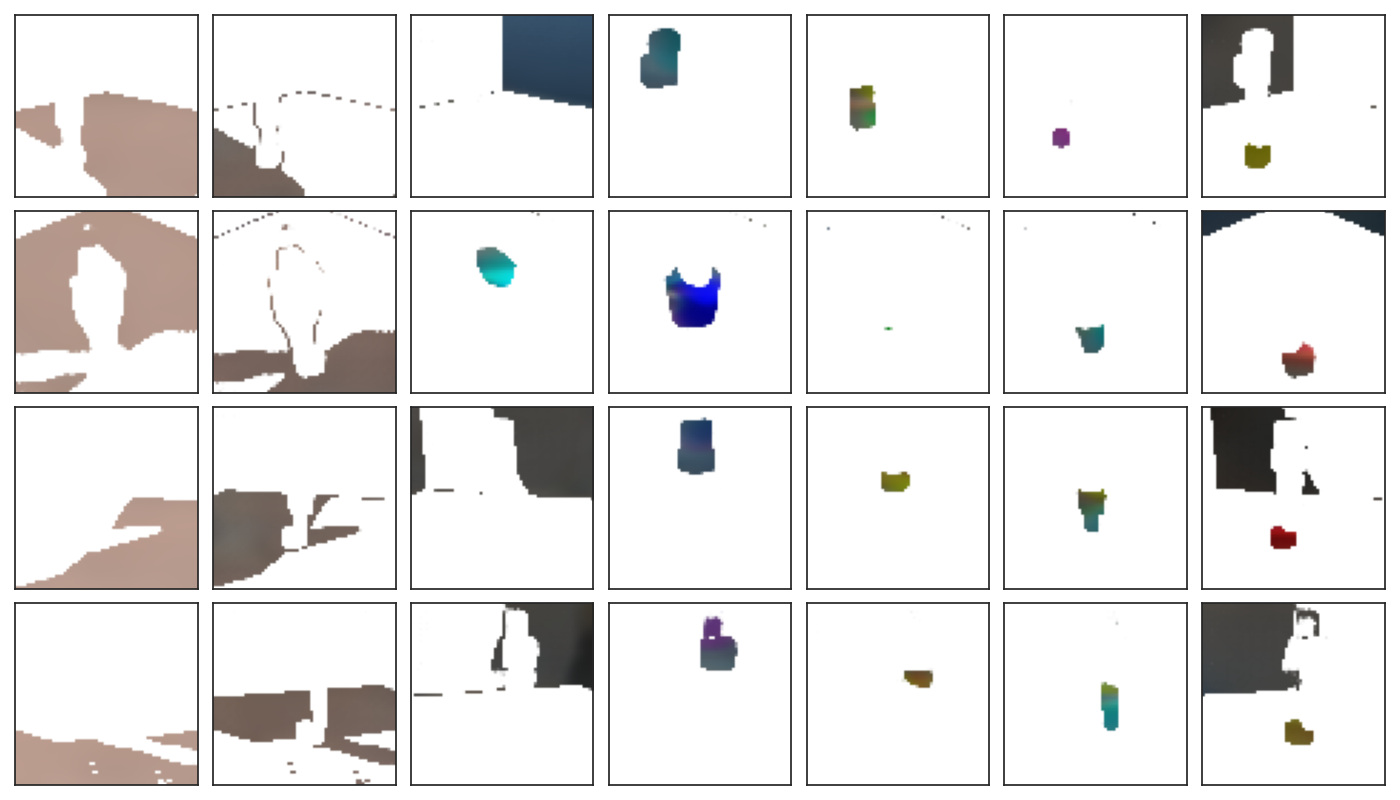}
    
    \vspace{0.45cm}
    
    \includegraphics[height=4cm]{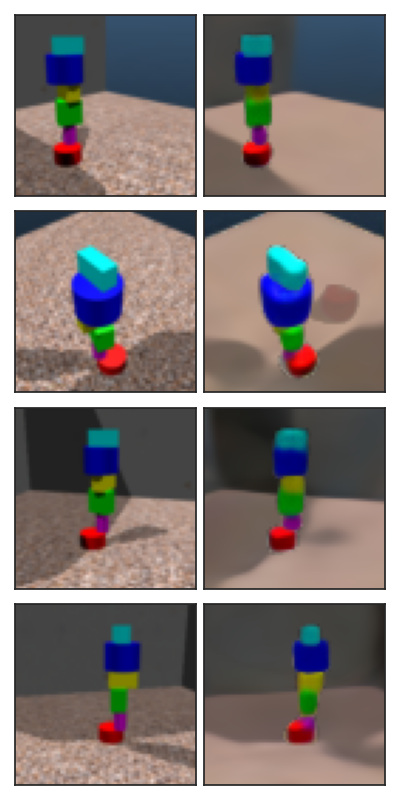}
    \hspace{0.2cm}
    \includegraphics[height=4cm]{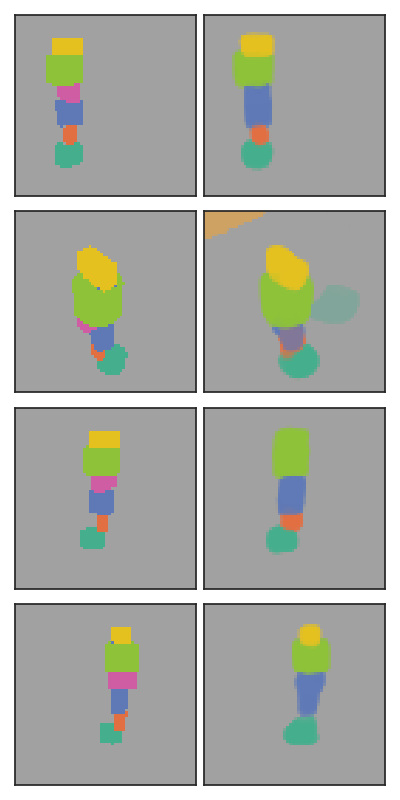}
    \hspace{0.2cm}
    \includegraphics[height=4cm]{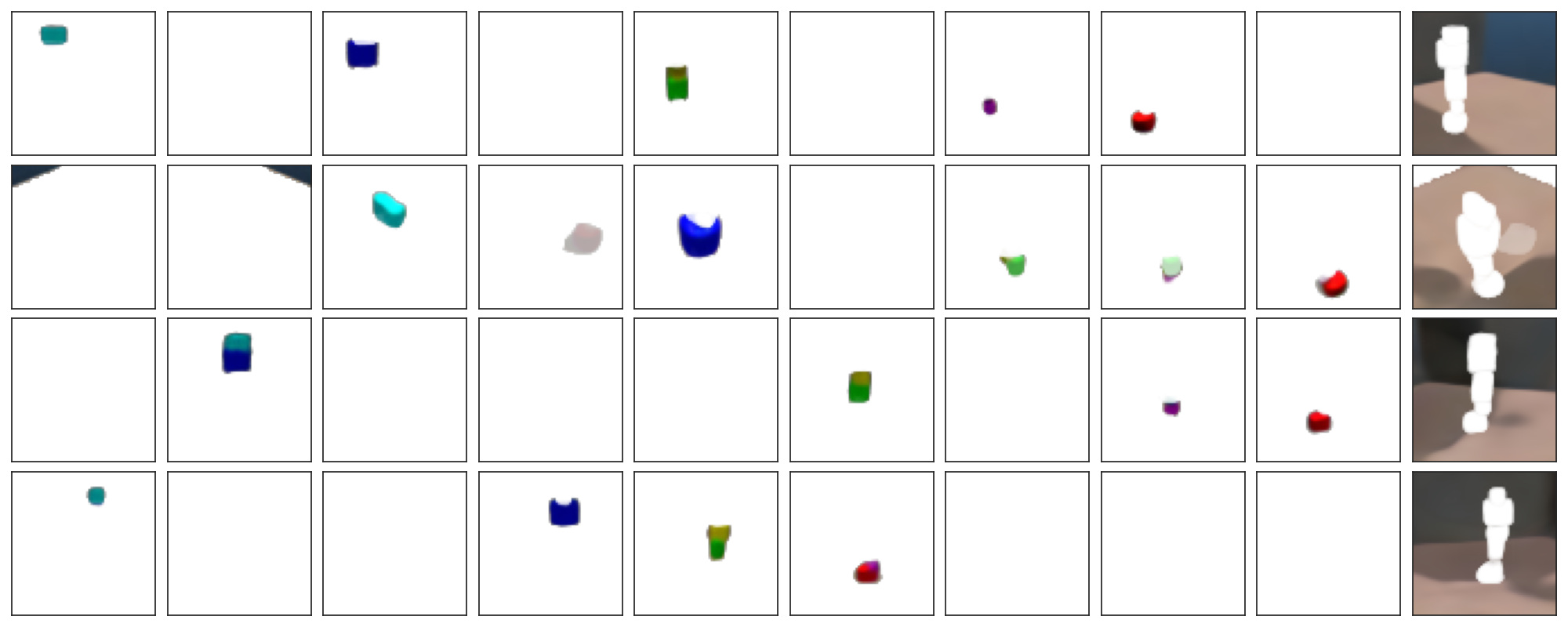}

    \vspace{0.45cm}
    \caption{\textbf{Reconstruction and segmentation} of 4 random images from the held-out test set of \textbf{Shapestacks}. Top to bottom: MONet, Slot Attention, GENESIS, SPACE. Left to right: input, reconstruction, ground-truth masks, predicted (soft) masks, slot-wise reconstructions (masked with the predicted masks). As explained in the text, for SPACE we select the 10 most salient slots using the predicted masks. For each model type, we visualize the specific model with the highest ARI score in the \textit{validation} set. The images shown here are from the \textit{test} set and were not used for model selection.}
    \label{fig:app_model_viz_shapestacks}
\end{figure}

\begin{figure}
    \centering
    \includegraphics[height=4cm]{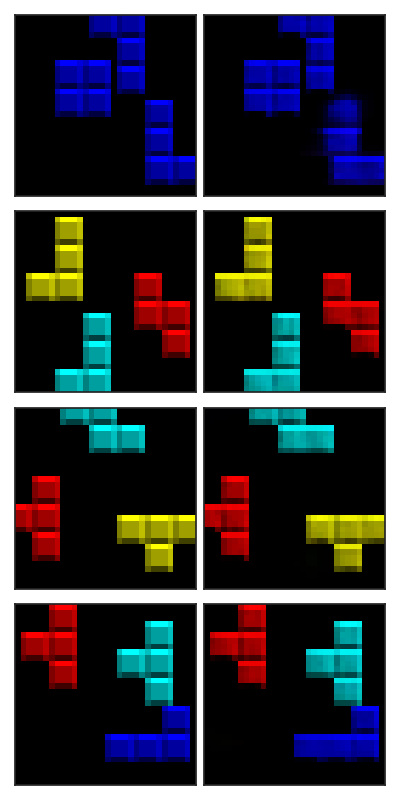}
    \hspace{0.2cm}
    \includegraphics[height=4cm]{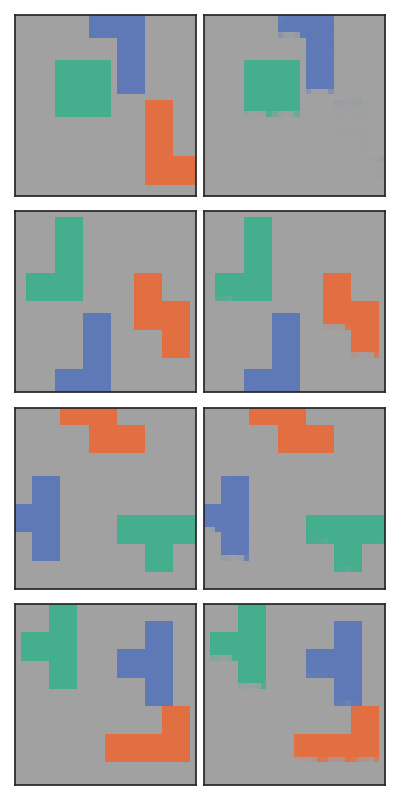}
    \hspace{0.2cm}
    \includegraphics[height=4cm]{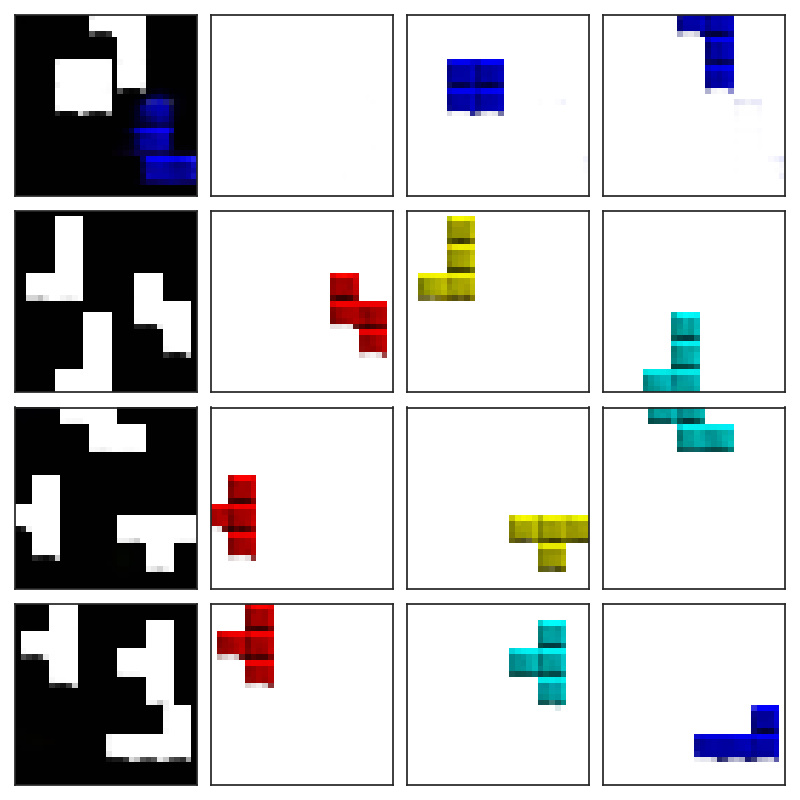}
    
    \vspace{0.45cm}
    
    \includegraphics[height=4cm]{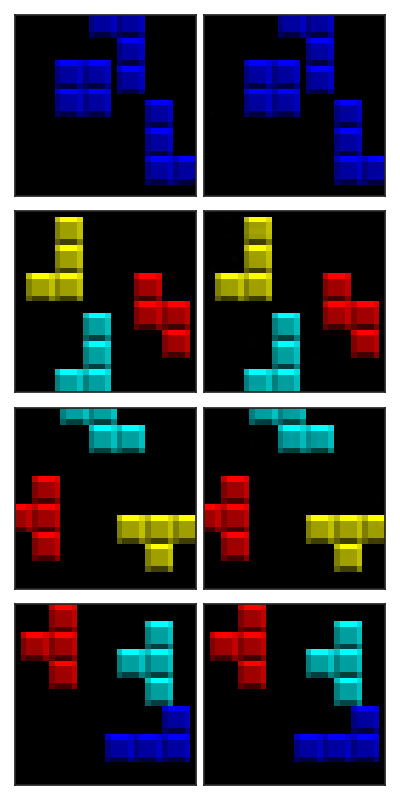}
    \hspace{0.2cm}
    \includegraphics[height=4cm]{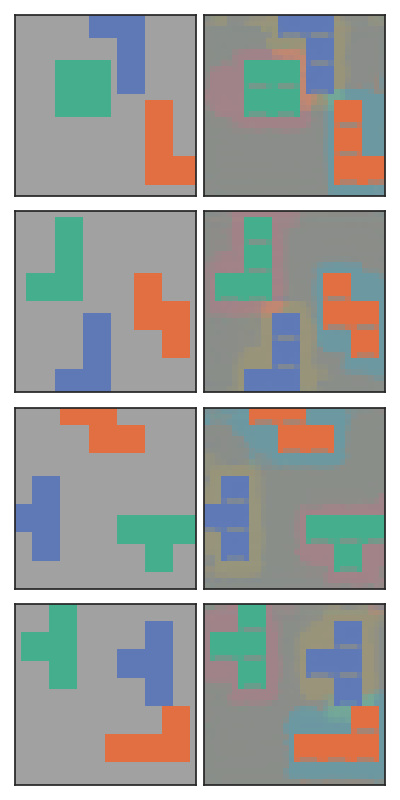}
    \hspace{0.2cm}
    \includegraphics[height=4cm]{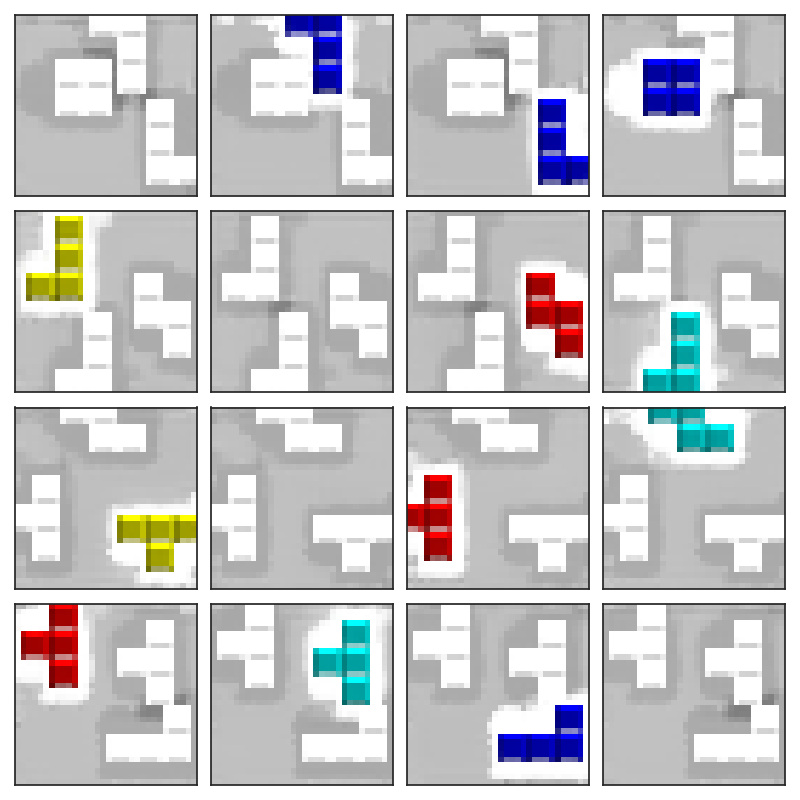}
    
    \vspace{0.45cm}
    
    \includegraphics[height=4cm]{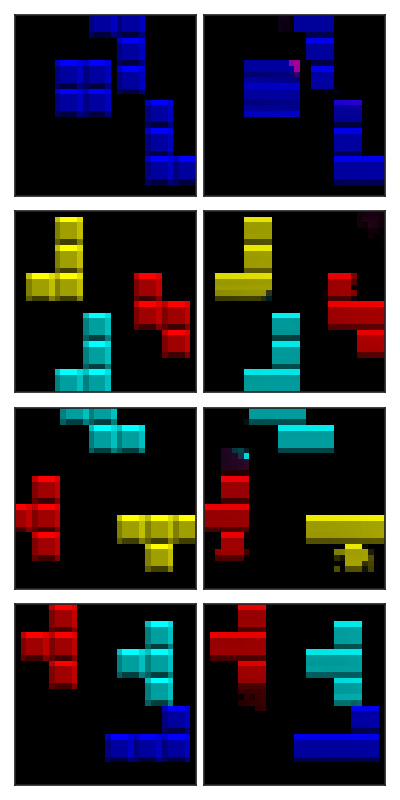}
    \hspace{0.2cm}
    \includegraphics[height=4cm]{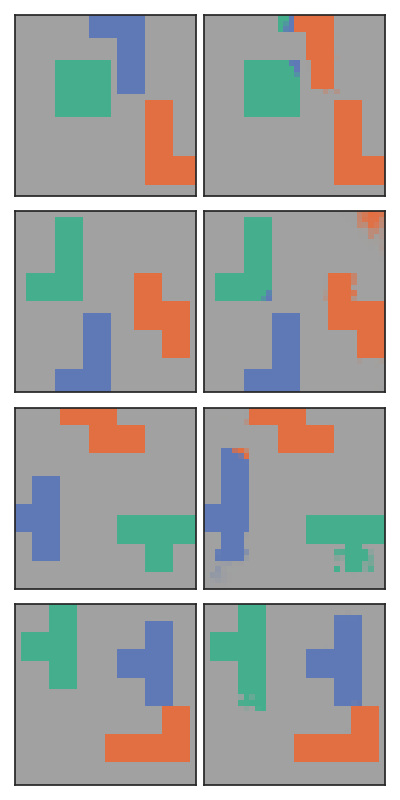}
    \hspace{0.2cm}
    \includegraphics[height=4cm]{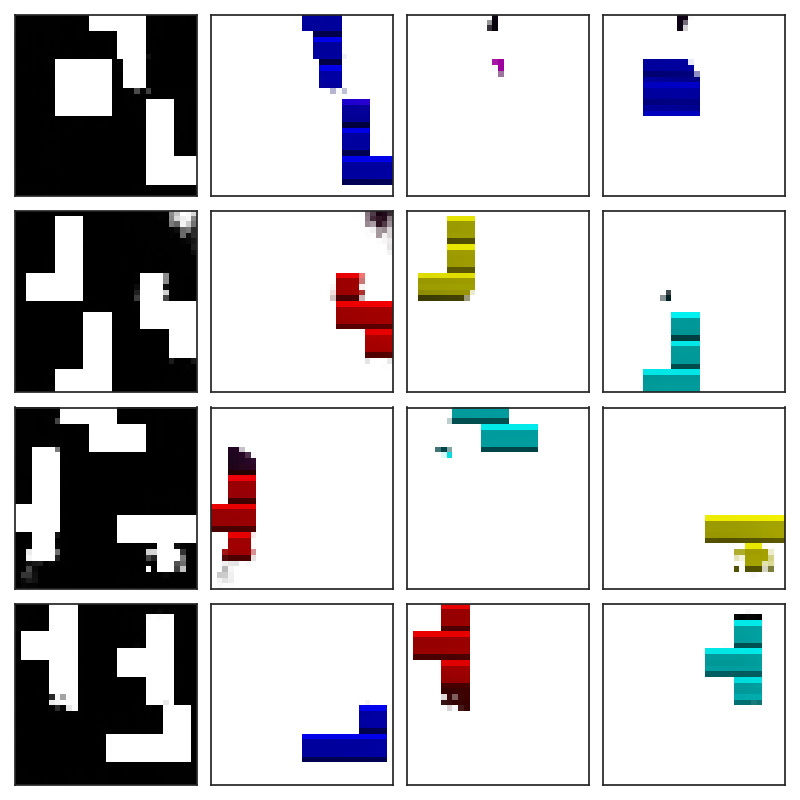}
    
    \vspace{0.45cm}

    \includegraphics[height=4cm]{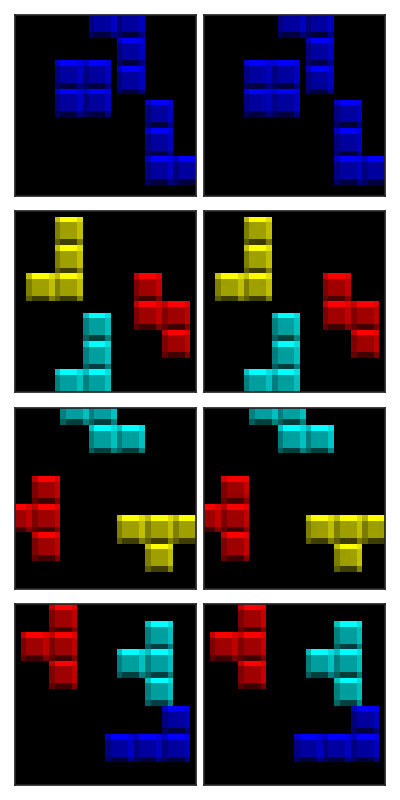}
    \hspace{0.2cm}
    \includegraphics[height=4cm]{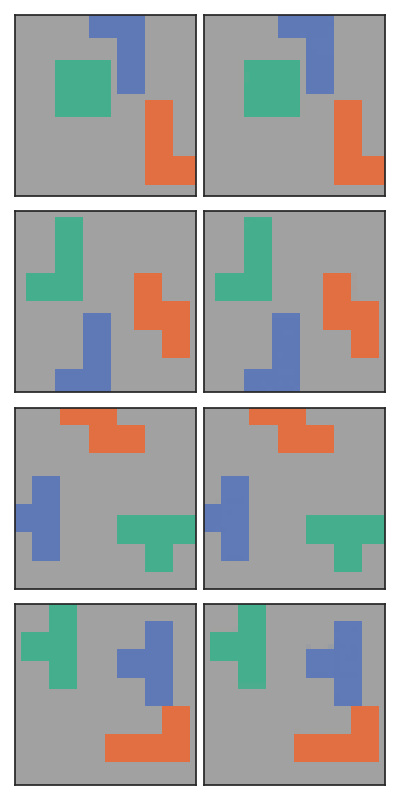}
    \hspace{0.2cm}
    \includegraphics[height=4cm]{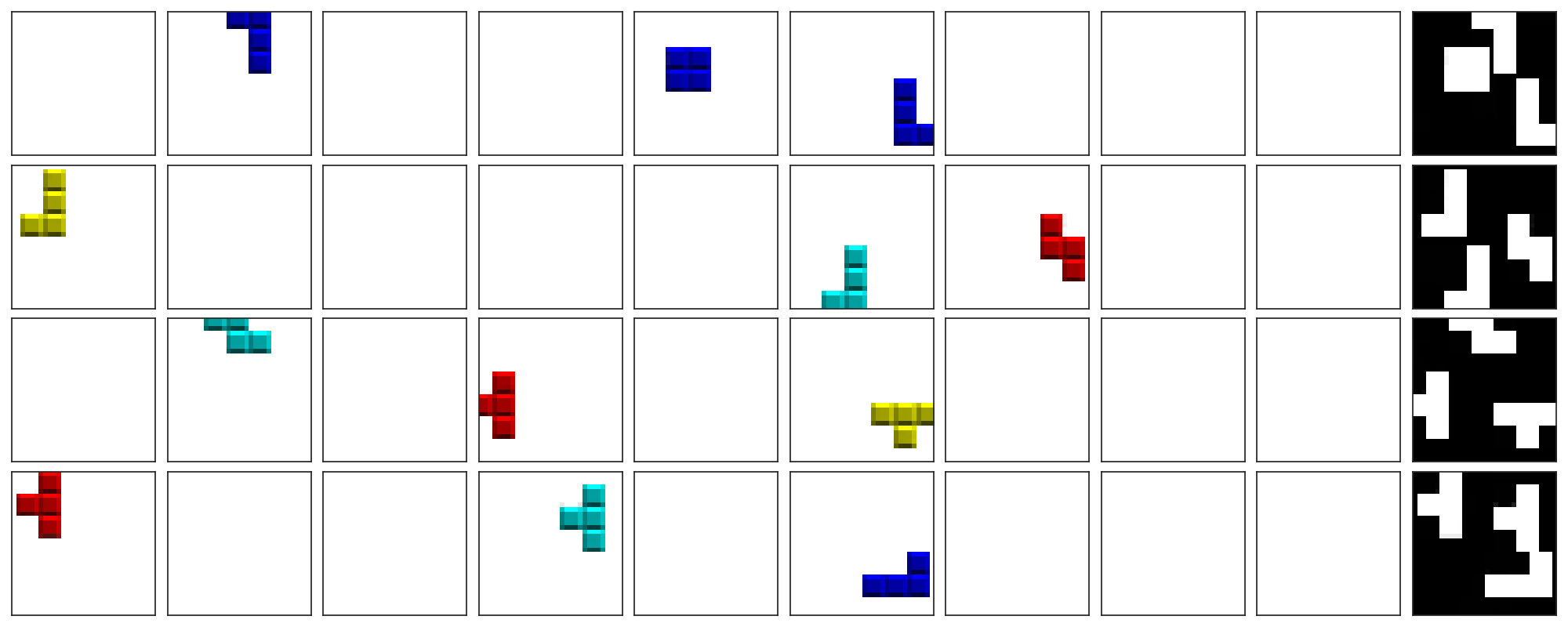}

    \vspace{0.45cm}
    \caption{\textbf{Reconstruction and segmentation} of 4 random images from the held-out test set of \textbf{Tetrominoes}. Top to bottom: MONet, Slot Attention, GENESIS, SPACE. Left to right: input, reconstruction, ground-truth masks, predicted (soft) masks, slot-wise reconstructions (masked with the predicted masks). As explained in the text, for SPACE we select the 10 most salient slots using the predicted masks. For each model type, we visualize the specific model with the highest ARI score in the \textit{validation} set. The images shown here are from the \textit{test} set and were not used for model selection.}
    \label{fig:app_model_viz_tetrominoes}
\end{figure}

\begin{figure}
    \centering
    \includegraphics[height=4cm]{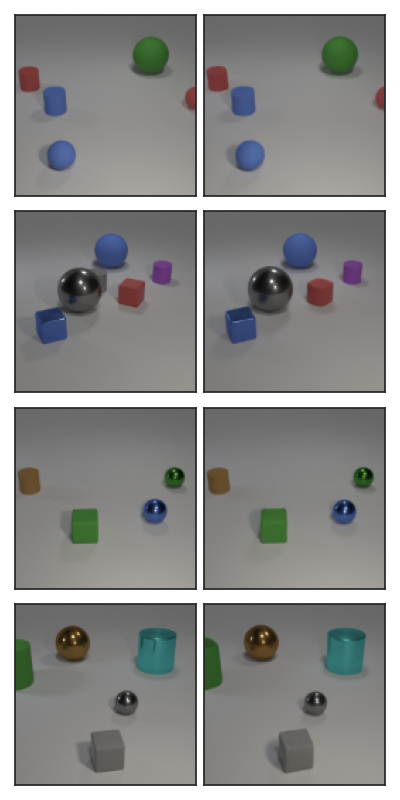}
    \hspace{0.2cm}
    \includegraphics[height=4cm]{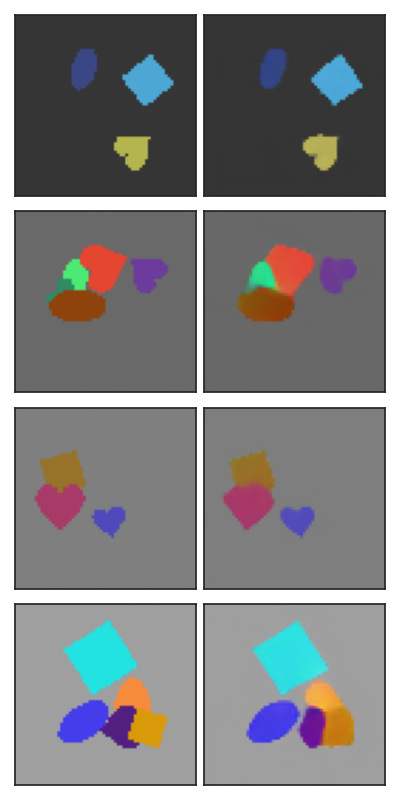}
    \hspace{0.2cm}
    \includegraphics[height=4cm]{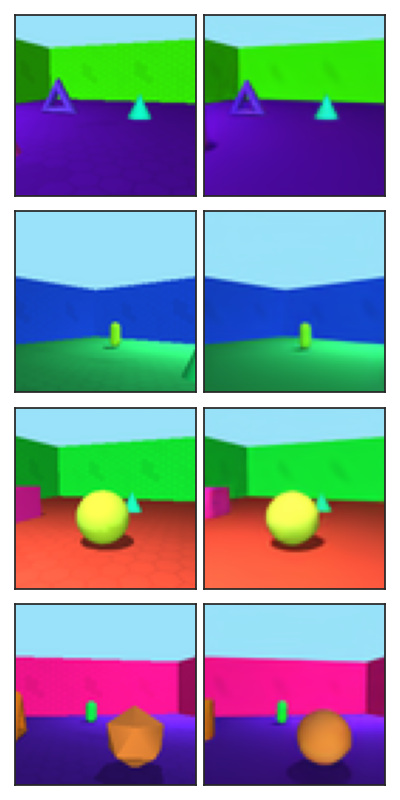}
    \hspace{0.2cm}
    \includegraphics[height=4cm]{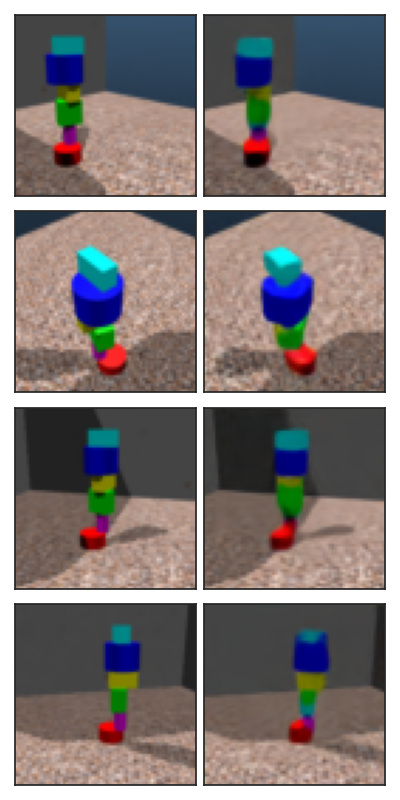}
    \hspace{0.2cm}
    \includegraphics[height=4cm]{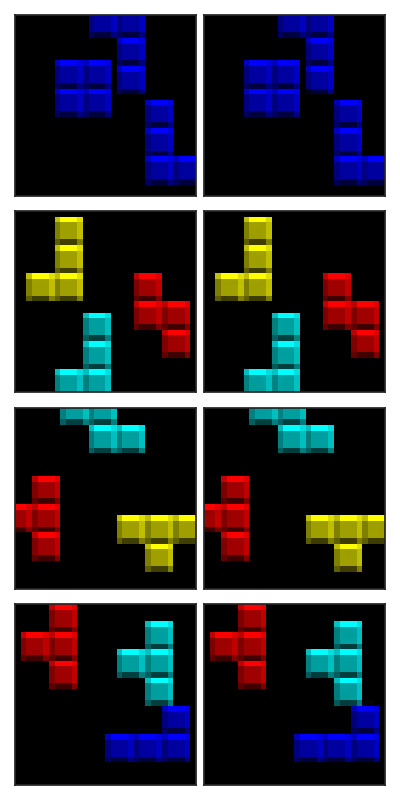}
    
    \vspace{0.45cm}
    
    \includegraphics[height=4cm]{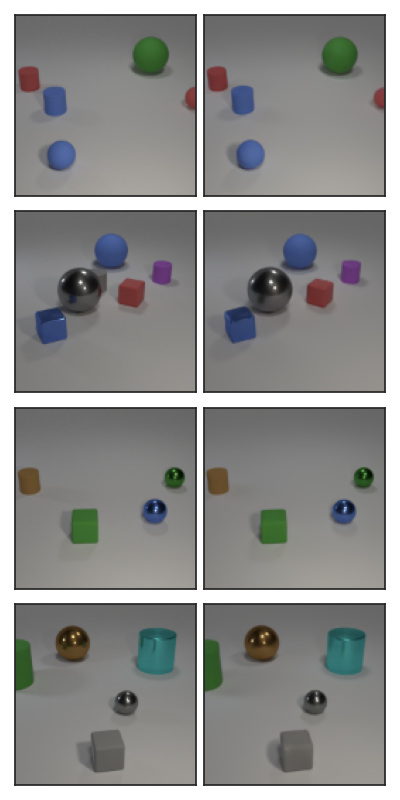}
    \hspace{0.2cm}
    \includegraphics[height=4cm]{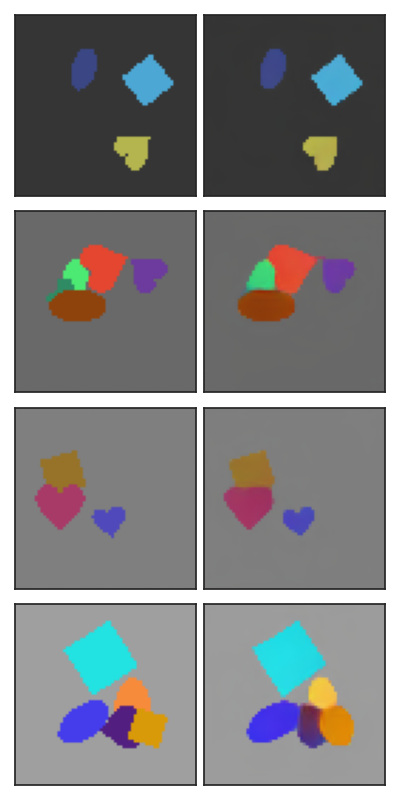}
    \hspace{0.2cm}
    \includegraphics[height=4cm]{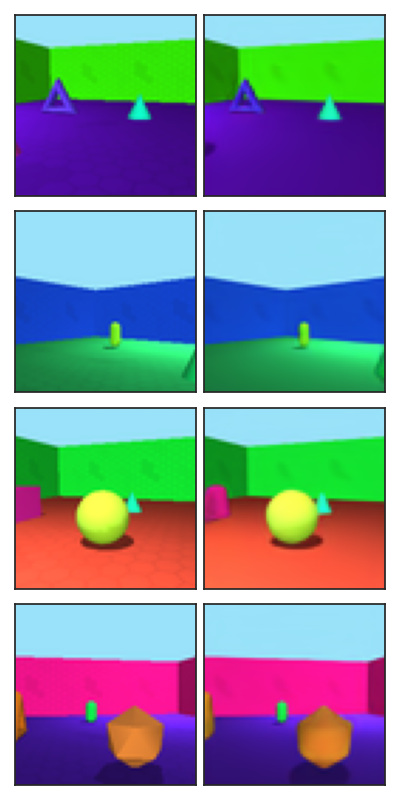}
    \hspace{0.2cm}
    \includegraphics[height=4cm]{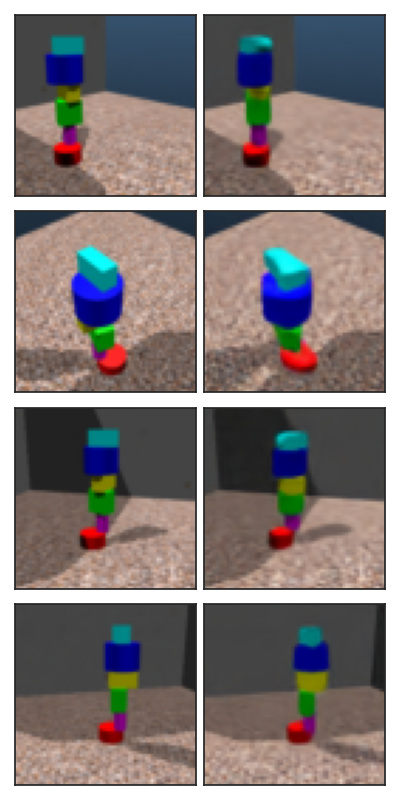}
    \hspace{0.2cm}
    \includegraphics[height=4cm]{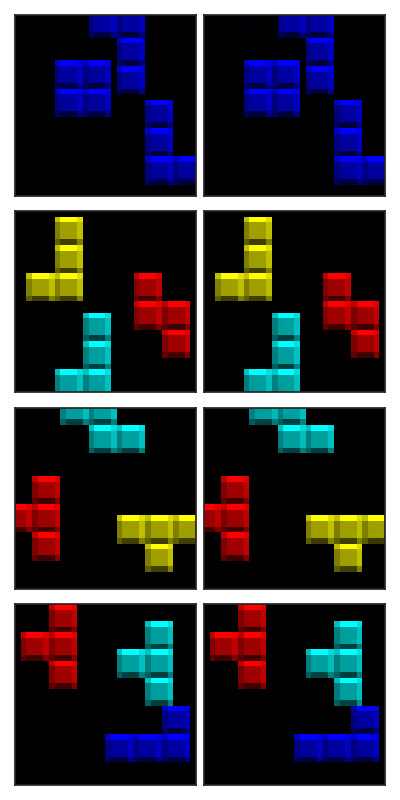}
    
    \vspace{0.45cm}
    \caption{\textbf{Input--reconstruction pairs} of 4 random images from the held-out \emph{test} set of all 5 datasets, for the VAE model with convolutional (top) and broadcast (bottom) decoder. Each VAE type was trained with 5 random seeds, and for each type we show here the model with the lowest MSE on the \textit{validation} set. The images shown here are from the \textit{test} set and were not used for model selection. For each image, we show the input on the left and the reconstruction on the right. As these are not slot-based models, segmentation masks and slot-wise reconstructions are not available.}
    \label{fig:app_model_viz_vaes}
\end{figure}

In \cref{fig:app_model_viz_clevr,fig:app_model_viz_multidsprites,fig:app_model_viz_tetrominoes,fig:app_model_viz_objectsroom,fig:app_model_viz_shapestacks} we show the reconstruction and segmentation performance of a selection of object-centric models on a random subset of held-out test images, for all 5 datasets. We select one object-centric model per type (MONet, Slot Attention, GENESIS, and SPACE) based on the ARI score on the validation set. The images we show were not used for model selection. For each model we show the following:
\begin{itemize}
    \item \emph{Input and reconstructed images}.
    \item \emph{Ground-truth and inferred segmentation maps}. Here we use a set of 8 colors and assign each object (or slot) to a color. If there are more than 8 slots, we loop over the 8 colors again (this does not happen here, except in SPACE, where it is not an issue in practice).
    Rather than taking hard masks, we treat the masks as ``soft'', such that a pixel's color is a weighted mean of the 8 colors according to the masks. This is evident in Slot Attention, which typically splits the background smoothly across slots (consistently with the qualitative results shown in \citet{locatello2020object}).
    For clarity, we match (with the Hungarian algorithm) the colors of the ground-truth and predicted masks using the cosine distance ($1$ minus the cosine similarity) between masks.
    \item \emph{Slot-wise reconstructions}. Each column corresponds to a slot in the object-centric representation of the model. Here we show the entire slot reconstruction with the inferred slot mask as alpha (transparency) channel. The overall reconstruction is the sum of these images.
    Since SPACE has in total up to 69 slots in our experiments ($K=5$ background slots, and a grid of foreground slots of size $G \times G$ with $G=8$), it is impractical to show all slots here. We choose instead to show the 10 most salient slots, selected according to the average mask value over the image. This number is sufficient as most slots are unused. When selecting slots this way, the selected slots are shown in their original order (in SPACE, the background slots are appended to the foreground slots).
\end{itemize}
For completeness, in \cref{fig:app_model_viz_vaes} we show inputs and reconstructions for one VAE baseline per type (convolutional and broadcast decoder), selected using the reconstruction MSE on the validation set.

Finally, \cref{fig:ood_visualizations_clevr,fig:ood_visualizations_multidsprites,fig:ood_visualizations_objects_room,fig:ood_visualizations_shapestacks,fig:ood_visualizations_tetrominoes} show input--reconstruction pairs for each dataset, model type, and distribution shift.
Note that the comparison is not necessarily fair, since object-centric models were chosen using the validation ARI on the training distribution, while VAEs were chosen in a similar way but using the MSE.
However, these qualitative results can still be highly informative. 
We report some examples:
\begin{itemize}
    \item Most object-centric models are relatively robust to shifts affecting a single object, as discussed in the main text based on quantitative results.
    
    \item On the other hand, they are often not robust to global shifts, especially when cropping and enlarging the scene.
    
    \item MONet achieves relatively good reconstructions even out of distribution, probably because images are segmented mostly based on color. This was suggested by \citet{papa2022inductive}, where the models are trained on objects with style transfer. However, we conjecture the behavior may be the same in our case, and that the argument should also apply to other distribution shifts, as seen by the relatively accurate reconstructions under both single-object and global distribution shifts. Note that, while reconstructions are potentially more accurate than for other models, this does not mean that MONet has segmented the object correctly.
    
    \item Although its ARI score does not decrease significantly, Slot Attention may not always handle more objects than in the training distribution, even when the number of slots in the model is increased. This is consistent with the results reported by \citet{locatello2020object}, and increasing the number of Slot Attention iterations at test time seems to be a promising approach \citep[Fig.~2]{locatello2020object}. 
    
    \item VAEs seem to be relatively good at generalizing to a greater number of objects in CLEVR. In particular, they reconstruct images with the correct number of objects, although a few details of the objects may not be inferred correctly (e.g. an object may be reconstructed with the wrong size, color, or shape). This is surprising, since VAEs do not have any inductive bias for this, and the fact that the encoder is OOD (i.e., the encoder input is OOD w.r.t. the distribution used to train the encoder itself) might lead us to expect poor generalization capabilities, as discussed by \citet{dittadi2020transfer} and \citet{traeuble2022role} in the ``OOD2'' case.
    On the other hand, some object-centric models are remarkably robust to this shift (in particular SPACE, as confirmed by the ARI in \cref{fig:figures/ood/metrics/only_ARI_ood_global/barplots_ARI}).
    
\end{itemize}



\def\OodVizVspace{\hspace{11pt}}
\def\OodVizHspace{\hspace{11pt}}
\newcommand{\OodViz}[3]{\includegraphics[height=0.17\textheight]{figures/model_viz/#1/#2/#3/input_recon}}
\newcommand{\OodVizRow}[2]{%
    \OodViz{#1}{monet}{#2}
    \OodVizHspace
    \OodViz{#1}{slot-attention}{#2}
    \OodVizHspace
    \OodViz{#1}{genesis}{#2}
    \OodVizHspace
    \OodViz{#1}{space}{#2}
    \OodVizHspace
    \OodViz{#1}{baseline_vae_mlp}{#2}
    \OodVizHspace
    \OodViz{#1}{baseline_vae_broadcast}{#2}
}

\begin{figure}
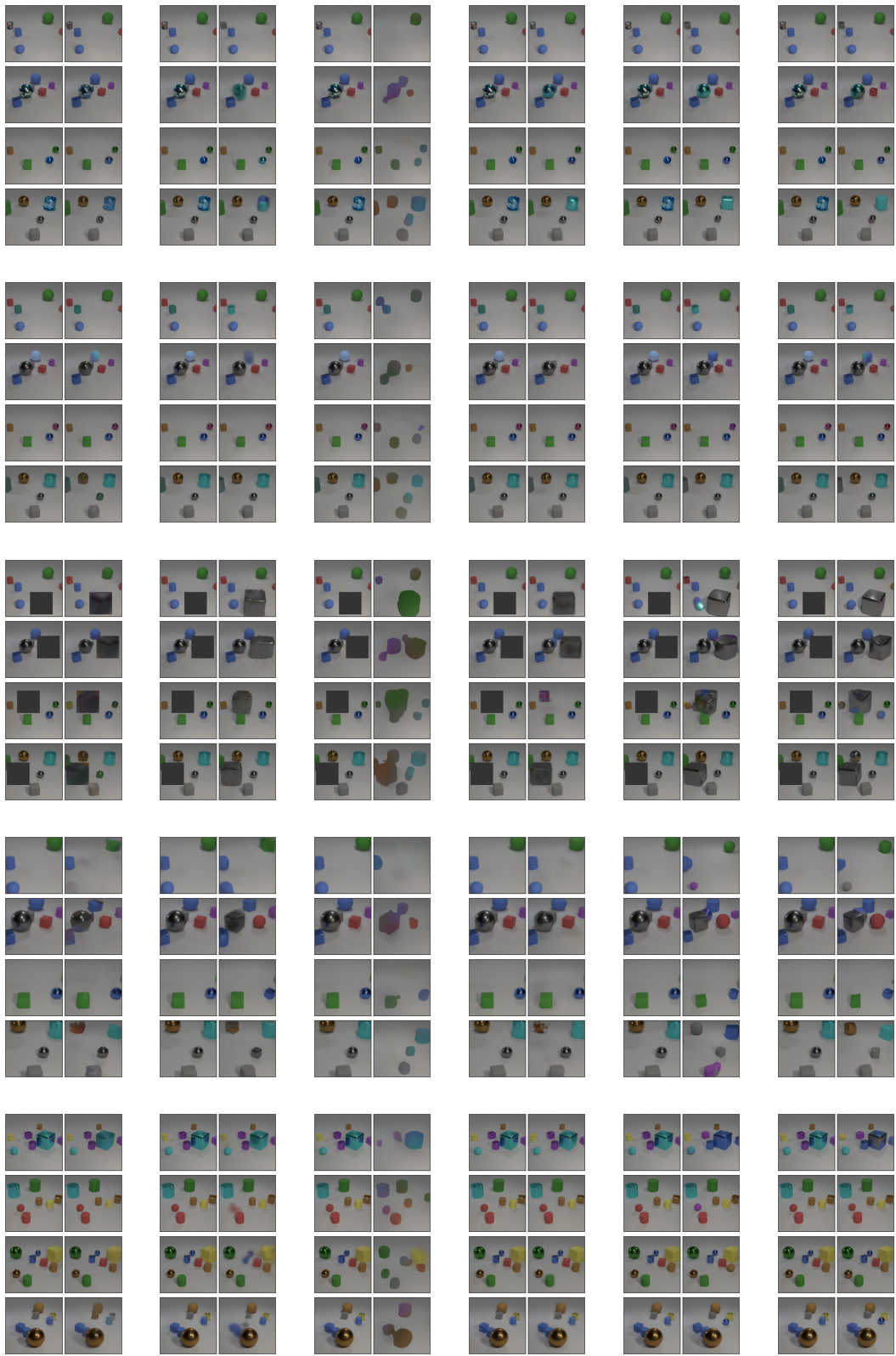

    \centering
    \OodVizRow{clevr}{object_style}
    
    \OodVizVspace
    
    \OodVizRow{clevr}{object_color}
    
    \OodVizVspace
    
    \OodVizRow{clevr}{occlusion}
    
    \OodVizVspace
    
    \OodVizRow{clevr}{crop}
    
    \OodVizVspace
    
    \OodVizRow{clevr}{num_objects}
    
    \caption{\textbf{Inputs and reconstructions for OOD images in CLEVR.} Columns from left to right: MONet, Slot Attention, GENESIS, SPACE, convolutional decoder VAE, broadcast decoder VAE. Rows from top to bottom: object style, object color, occlusion, crop, number of objects.}
    \label{fig:ood_visualizations_clevr}
\end{figure}

\begin{figure}
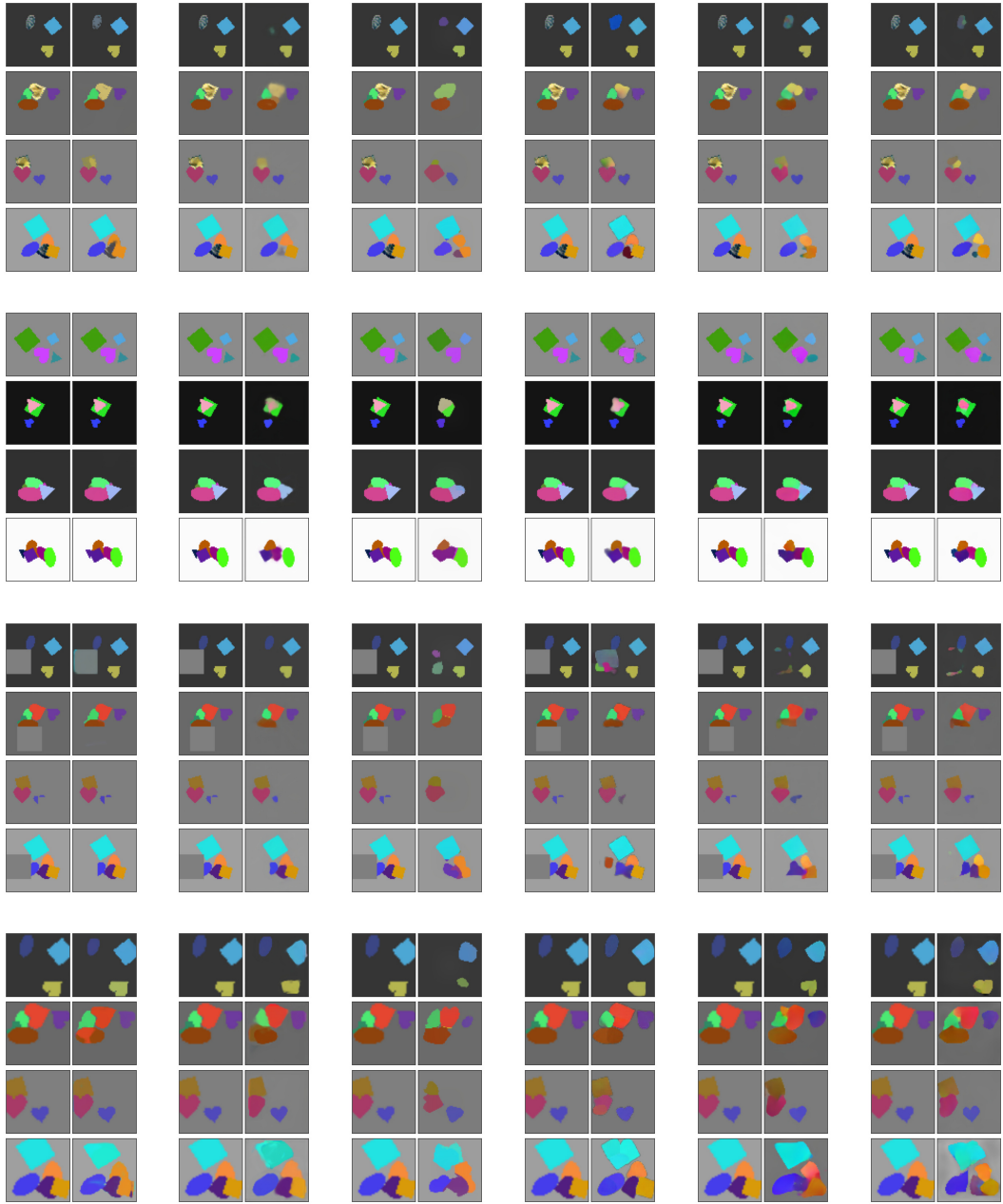

    \centering
    \OodVizRow{multidsprites}{object_style}
    
    \OodVizVspace
    
    \OodVizRow{multidsprites}{object_shape}
    
    \OodVizVspace
    
    \OodVizRow{multidsprites}{occlusion}
    
    \OodVizVspace
    
    \OodVizRow{multidsprites}{crop}
    
    \caption{\textbf{Inputs and reconstructions for OOD images in Multi-dSprites.} Columns from left to right: MONet, Slot Attention, GENESIS, SPACE, convolutional decoder VAE, broadcast decoder VAE. Rows from top to bottom: object style, object shape, occlusion, crop.}
    \label{fig:ood_visualizations_multidsprites}
\end{figure}

\begin{figure}
    \centering
    \OodVizRow{objects_room}{object_style}
    
    \OodVizVspace
    
    \OodVizRow{objects_room}{object_color}
    
    \OodVizVspace
    
    \OodVizRow{objects_room}{occlusion}
    
    \OodVizVspace
    
    \OodVizRow{objects_room}{crop}
    
    \caption{\textbf{Inputs and reconstructions for OOD images in Objects Room.} Columns from left to right: MONet, Slot Attention, GENESIS, SPACE, convolutional decoder VAE, broadcast decoder VAE. Rows from top to bottom: object style, object color, occlusion, crop.}
    \label{fig:ood_visualizations_objects_room}
\end{figure}

\begin{figure}
    \centering
    \OodVizRow{shapestacks}{object_style}
    
    \OodVizVspace
    
    \OodVizRow{shapestacks}{object_color}
    
    \OodVizVspace
    
    \OodVizRow{shapestacks}{occlusion}
    
    \OodVizVspace
    
    \OodVizRow{shapestacks}{crop}
    
    \caption{\textbf{Inputs and reconstructions for OOD images in Shapestacks.} Columns from left to right: MONet, Slot Attention, GENESIS, SPACE, convolutional decoder VAE, broadcast decoder VAE. Rows from top to bottom: object style, object color, occlusion, crop.}
    \label{fig:ood_visualizations_shapestacks}
\end{figure}

\begin{figure}
    \centering
    \OodVizRow{tetrominoes}{object_style}
    
    \OodVizVspace
    
    \OodVizRow{tetrominoes}{object_color}
    
    \OodVizVspace
    
    \OodVizRow{tetrominoes}{occlusion}
    
    \OodVizVspace
    
    \OodVizRow{tetrominoes}{crop}
    
    \caption{\textbf{Inputs and reconstructions for OOD images in Tetrominoes.} Columns from left to right: MONet, Slot Attention, GENESIS, SPACE, convolutional decoder VAE, broadcast decoder VAE. Rows from top to bottom: object style, object color, occlusion, crop.}
    \label{fig:ood_visualizations_tetrominoes}
\end{figure}

\end{document}